\documentclass[review]{elsarticle}

\usepackage{lineno,hyperref}
\usepackage{tabularx}
\usepackage{tabulary}
\usepackage{array}
\newcolumntype{P}[1]{>{\centering\arraybackslash}p{#1}}
\usepackage{multirow}
\usepackage{graphicx}
\usepackage{subcaption}
\usepackage{rotating}
\usepackage{lipsum}
\usepackage{lscape}
\usepackage{amsmath,amssymb,amsthm,mathtools}
\usepackage{lmodern}
\usepackage{titlesec}
\titleformat{\subsection}
  {\normalfont\bfseries} 
  {\thesubsection.}       
  {1em}                  
  {}                    
\usepackage[table,xcdraw]{xcolor}
\usepackage[ruled,linesnumbered]{algorithm2e}
\SetArgSty{textnormal}
\SetKwInput{KwStart}{Start}                
\SetKwInput{KwEnd}{End}   
\SetKwInput{Kwinitialization}{Initialisation}
\SetKwInput{kwdefine}{Define}
\usepackage{textcomp}
\usepackage{float}
\usepackage[margin=1in]{geometry}
\modulolinenumbers[5]
\usepackage{pdflscape}
\journal{arxiv}

\usepackage{numcompress}\biboptions{authoryear}

\begin{document}
\setlength\abovedisplayskip{0.5pt}
\setlength\belowdisplayskip{0.5pt}
\RestyleAlgo{ruled}
\SetKwComment{Comment}{/* }{ */}

\begin{frontmatter}
\title{SHS: Scorpion Hunting Strategy Swarm Algorithm}

\author[iiserb]{Abhilash Singh \corref{corr}}

\author[inr]{Seyed Muhammad Hossein Mousavi }
\author[iiserb]{Kumar Gaurav \corref{corr}}
\cortext[corr]{Corresponding authors \\ 
 Email address: sabhilash@iiserb.com (Abhilash Singh) ORCID: 0000-0001-6270-9355\\ Email address: hossein-mousavi@ieee.org (Seyed Muhammaad Hossein Mousavi) ORCID: 0000-0001-6906-2152\\
 Email address: kgaurav@iiserb.ac.in (Kumar Gaurav) ORCID: 0000-0003-1636-9622} 

\address[iiserb]{Fluvial Geomorphology and Remote Sensing Laboratory, Indian Institute of Science Education and Research Bhopal, India}
\address[inr]{PARS-AI, Tehran, Iran





}


\begin{abstract}

We introduced the Scorpion Hunting Strategy (SHS), a novel population-based, nature-inspired optimisation algorithm. This algorithm draws inspiration from the hunting strategy of scorpions, which identify, locate, and capture their prey using the alpha and beta vibration operators. These operators control the SHS algorithm's exploitation and exploration abilities. To formulate an optimisation method, we mathematically simulate these dynamic events and behaviors. We evaluate the effectiveness of the SHS algorithm by employing 20 benchmark functions (including 10 conventional and 10 CEC2020 functions), using both qualitative and quantitative analyses. Through a comparative analysis with 12 state-of-the-art meta-heuristic algorithms, we demonstrate that the proposed SHS algorithm yields exceptionally promising results. These findings are further supported by statistically significant results obtained through the Wilcoxon rank sum test. Additionally, the ranking of SHS, as determined by the average rank derived from the Friedman test, positions it at the forefront when compared to other algorithms. Going beyond theoretical validation, we showcase the practical utility of the SHS algorithm by applying it to six distinct real-world optimisation tasks. These applications illustrate the algorithm's potential in addressing complex optimisation challenges. In summary, this work not only introduces the innovative SHS algorithm but also substantiates its effectiveness and versatility through rigorous benchmarking and real-world problem-solving scenarios.

\end{abstract}


\end{frontmatter}

\section{Introduction}

A family of stochastic search strategies known as the meta-heuristic algorithm performs exceptionally well when handling multi-modal, discontinuous, and non-differentiable optimisation tasks. To solve real-world optimisation problems, the meta-heuristic algorithms have become very popular \citep{mousavi2020evolutionary,singh2021nature,dhiman2021ssc,abualigah2021group,mousavi2022introducing,abualigah2022meta,rahimi2021machine,kalayci2020mutual,ccevik2017voxel,taylan2021new,ozougur2020ensemble,savku2017optimal,baltas2022optimal,vasant2017handbook}. For instant, \citet{thanh2023hydrogen} successfully demonstrated the use of these meta-heuristic algorithms to predict the hydrogen storage by leveraging random forest algorithm from numerous input features such as pressure, temperature, different structural features, chemical activating, and solid adsorbents. \citet{zhang2023improving} exploits the capability of metaheuristic algorithms in combination with a multi-layer perceptron and radial basis function neural networks to forecast shale wettability: a crucial aspect for efficient carbon capture utilisation and storage. The inputs encompass pressure, temperature, salinity, total organic carbon, and theta zero.

It should be emphasised that varied aspects of nature serve as meta-heuristic inspiration. Algorithms such as differential evolution, genetic algorithms, and particle swarm optimisation are being widely used. Particle Swarm Optimisation (PSO), for instance, draws inspiration from the actions of flocking fish or birds, which are portrayed as potential solutions by moving around the search field. Through the stochastic behaviours of mutation, reproduction, and selection, the Genetic Algorithm (GA) imitates the evolutionary process. Numerous unique meta-heuristics have been created and used in various disciplines. Section \ref{sec:rw} of this article provides a complete literature review.

According to No Free Lunch (NFL) theorem \citep{wolpert1997no},  all meta-heuristic algorithms perform more or less similarly, given all sorts of optimisation problems. In other words, there isn't a single algorithm that is the greatest at solving all optimisation problems, which means that when prior knowledge (such as algorithmic parameters and convergence criteria) is provided to a meta-heuristic algorithm to solve a particular problem, the performances of different algorithms aren't all supposed to be equal. Finding the best algorithm for each distinct type of optimisation problem continues to be a difficulty. Due to the various natural or biological tendencies, that each meta-heuristic algorithm has been inspired by, they all have unique properties. To assess performance and determine an appropriate application space with ongoing progress, a meta-heuristic algorithm requires extensive tests from various benchmark functions and real-world applications in many domains. The aforementioned justifications encourage the development of new meta-heuristic algorithms to address various optimisation issues.

The existing literature on meta-heuristic algorithms has demonstrated significant advancements in various optimisation techniques; however, a notable gap is observed in the representation of algorithms that are rooted in hunting patterns. While a handful of existing algorithms have explored this approach, such as the hunting search algorithm inspired by the group hunting of wolves \citep{oftadeh2010novel}, deer hunting optimisation inspired by the hunting behavior of humans towards deer \citep{brammya2019deer}, and the cheetah optimiser inspired by the hunting strategies of cheetahs \citep{akbari2022cheetah}, a comprehensive exploration of hunting patterns as a basis for optimisation, especially using cannibalistic animals that do not engage in group hunting, remains largely unexplored. This underscores the need for further research and innovation in developing novel meta-heuristic algorithms that draw inspiration from the intricate hunting strategies of cannibalistic animals observed in nature. Such efforts hold the potential to lead to the discovery of more effective and adaptable optimisation techniques.

In this study, we propose a novel meta-heuristic algorithm, ``Scorpion Hunting Strategy (SHS)," for optimisation. It is a swarm-based algorithm that is inspired by the hunting strategy of scorpions, a cannibalistic arachnids, to attack their prey. We assess the efficiency and robustness of the SHS algorithm by evaluating its performance on twenty different benchmark optimisation functions (including unimodal, multimodal, hybrid, and composite). For a fair and unbiased evaluation, we compared its performance with twelve different meta-heuristic algorithms. We performed the statistical analysis to report the quantitative outcome of the SHS algorithm. Finally, we applied SHS to solve six real-world optimisation tasks in various application domains.

This manuscript consists of six sections. Section \ref{sec:rw} discusses the literature review of meta-heuristic algorithms. Section \ref{sec:shs-alg} explains the inspiration and workings of the proposed SHS algorithm, including the benchmark test functions and simulation setup. Section \ref{sec:exp_results} presents the performance of the SHS algorithm on the test benchmark functions, compares the results with different meta-heuristic algorithms, and provides statistical analysis. Lastly, Sections \ref{sec:discussion} and \ref{sec:conclusion}, explores the potential of the SHS algorithm in solving optimisation problems across various application domains and summarise the conclusions, respectively.

\begin{figure}[ht!]
    \centering
    \includegraphics[width=\textwidth]{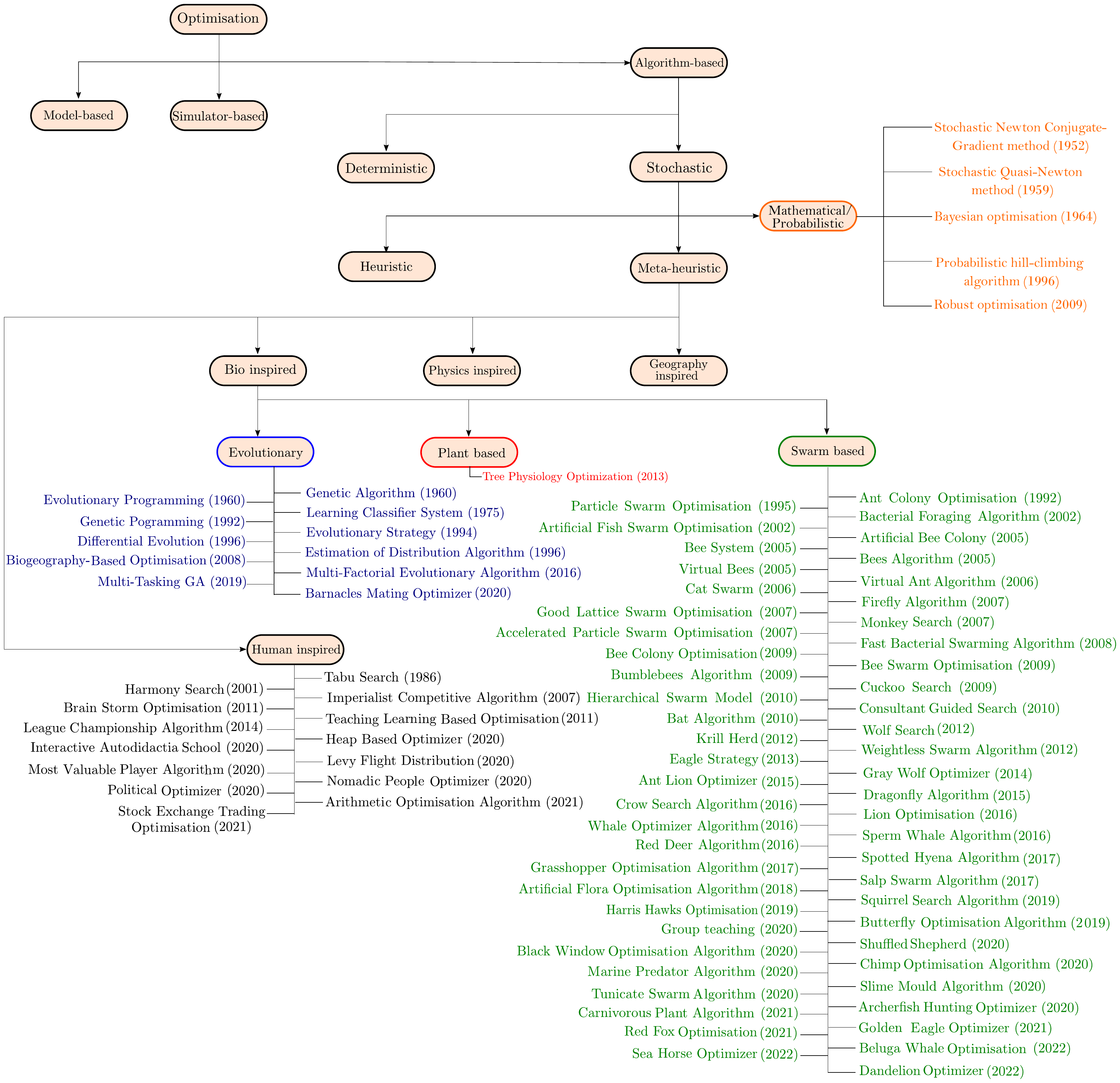}
    \caption{Taxonomy (modified after \citet{singh2021nature}) of four categories of meta-heuristic algorithms (bio-inspired, physics-inspired, geography-inspired, and human-inspired).}
    \label{fig:tax}
\end{figure}

\section{Literature review of meta-heuristic algorithms}
\label{sec:rw}
Meta-heuristic algorithms can be broadly divided into four categories: bio-inspired, physics-inspired, geography-inspired, and human-inspired (Figure \ref{fig:tax}). Bio-inspired algorithms are mainly classified into evolutionary, swarm-based, and plant-based. The optimisation algorithms based on evolutionary category are GA \citep{holland1992adaptation}, evolutionary programming \citep{yao1999evolutionary}, learning classifier system \citep{lanzi2000learning}, genetic programming \citep{koza1997genetic}, evolutionary strategy \citep{back1997handbook}, estimation of distribution \citep{muhlenbein1996recombination,zhang2008rm}, differential evolution \citep{qin2008differential}, multi-factorial evolutionary \citep{gupta2015multifactorial}, multi-tasking genetic \citep{dongrui}, and barnacles mating optimiser \citep{sulaiman2020barnacles}. GA is the most popular algorithm that has been widely used in different application domains \citep{singh2021nature}. These algorithms are based on Darwin's evolution theory of biological systems. They are controlled by three operators: reproduction, crossover, and mutation. They are used to solve both constrained and unconstrained optimisation problems by repeatedly modifying the population of an individual solution.

 The important algorithms under swarm category are ant colony optimisation \citep{blum2005ant,dorigo2005ant,socha2008ant,dorigo2010ant}, particle swarm optimisation \citep{kennedy1995particle,eberhart1995particle,shi1999empirical}, bacterial foraging \citep{passino2002biomimicry,das2009bacterial,passino2010bacterial}, artificial fish swarm optimisation \citep{li2003studies,li2004applications}, artificial bee colony \citep{karaboga2007artificial,karaboga2007powerful,karaboga2008performance,karaboga2009comparative,karaboga2011novel}, bee system \citep{lucic2001bee,luvcic2003vehicle}, bees \citep{pham2005bees,pham2006bees}, virtual bees \citep{yang2005engineering}, virtual ant \citep{yang2006application}, cat swarm \citep{chu2006cat,chu2007computational}, accelerated particle swarm optimisation \citep{yang2011accelerated}, good lattice swarm optimisation \citep{su2007good}, monkey search \citep{mucherino2007monkey}, firefly \citep{yang2009firefly,yang2010eagle,yang2010firefly}, fast bacterial swarming \citep{chu2008fast}, bee colony optimisation \citep{davidovic2016bee}, bee swarm optimisation \citep{drias2010bees,djenouri2012bees}, bumblebees \citep{comellas2009bumblebees}, cuckoo search \citep{9yang2009cuckoo,3yang2013multiobjective,4yang2014cuckoo}, hierarchical swarm model \citep{chen2010hierarchical}, consultant guided search \citep{iordache2010consultant,iordache2010consultant1,iordache2010consultant2}, bat \citep{yang2010new}, wolf search \citep{tang2012wolf}, krill herd \citep{gandomi2012krill}, weightless swarm \citep{ting2012weightless}, eagle strategy \citep{yang2012two}, gray wolf optimiser \citep{mirjalili2015effective,mirjalili2014grey}, ant lion optimiser \citep{mirjalili2015ant}, dragonfly \citep{mirjalili2016dragonfly}, crow search \citep{askarzadeh2016novel}, lion optimisation \citep{yazdani2016lion}, whale optimiser \citep{mirjalili2016whale}, sperm whale \citep{ebrahimi2016sperm}, red deer \citep{fard2016red}, grasshopper optimisation \citep{mirjalili2018grasshopper}, spotted hyena \citep{dhiman2018multi}, salp swarm \citep{mirjalili2017salp}, artificial flora optimisation \citep{cheng2018artificial}, squirrel search \citep{jain2019novel}, group teaching \citep{zhang2020group}, shuffled shepherd \citep{kaveh2020shuffled}, marine predator \citep{faramarzi2020marine}, slime mould \citep{li2020slime}, Harris hawks optimisation \citep{heidari2019harris}, tunicate swarm \citep{kaur2020tunicate}, archerfish hunting optimiser \citep{zitouni2022archerfish}, carnivorous plant \citep{ong2021carnivorous}, golden eagle optimiser \citep{mohammadi2021golden}, red fox optimisation \citep{polap2021red}, beluga whale optimisation \citep{zhong2022beluga}, sea horse optimiser \citep{zhao2022sea}, and dandelion optimiser \citep{zhao2022dandelion}. Ant colony optimisation, PSO, and artificial bee colony are the three most popular and classical swarm-based meta-heuristic algorithms. Ant colony optimisation is the first swarm intelligence-based meta-heuristic algorithm which is based on the foraging behaviour of ants in a colony. It is widely used to solve real-world optimisation problems, specifically discrete optimisation problems. PSO is based on the behaviour of moving creatures or particles in a group, like a school of fish or a flock of birds. It assumes that each individual creature gets benefit from the experience of all other creatures. The artificial bee colony algorithm is inspired by the foraging behaviour of bees.  Numerous real-world optimisation problems have been tackled using both PSO and artificial bee colony, both individually and in conjunction with other swarm algorithms. Recently developed algorithms, such as grey wolf optimiser, whale optimisation, Harris hawks optimisation, and marine predator have attracted the attention of the swarm intelligence community. The Grey wolf optimiser is inspired by the hunting mechanism and leadership hierarchy of grey wolves in nature. The whale optimisation algorithm is inspired by the foraging behaviour of whales consisting of three operators; prey, encircling prey, and bubble-net feeding maneuver. Harris hawks optimisation algorithm mimics the behaviour of the collaborative hunting strategy of Harris hawks. The marine predator algorithm simulates the biological interaction between marine predators and prey. All these algorithms have good convergence for various optimisation problems. Tree physiology optimisation \citep{halim2013nonlinear} comes under plant-based bio-inspired algorithms.

\begin{table}[t!]
\centering
\caption{Applications of meta-heuristic algorithms.}
 \resizebox{\textwidth}{!}{ 
\begin{tabular}{ccc}
\hline
Reference & Problem domain & \begin{tabular}[c]{@{}c@{}}Algorithms used\end{tabular} \\ \hline

\citet{patel2023simulation} & \begin{tabular}[c]{@{}c@{}}Aquifer parameter estimation \end{tabular}  & \begin{tabular}[c]{@{}c@{}}Differential evolution\\PSO\end{tabular}\\ \hline

\citet{singh2023leveraging} & \begin{tabular}[c]{@{}c@{}}Predicting malaria cases \end{tabular}  & \begin{tabular}[c]{@{}c@{}}PSO\end{tabular}\\ \hline

\citet{rahimi2023yield} & \begin{tabular}[c]{@{}c@{}}Biofuel yields prediction \end{tabular}  & \begin{tabular}[c]{@{}c@{}}GA\end{tabular}\\ \hline

\citet{mousavi2022fatty} & \begin{tabular}[c]{@{}c@{}}Fatty liver level recognition \\ (Image segmentation)   \end{tabular}  & \begin{tabular}[c]{@{}c@{}} PSO  \end{tabular}\\ \hline

\citet{mousavi2022weevil} & \begin{tabular}[c]{@{}c@{}}Optimal inventory control\\ Bin packing problem   \end{tabular}  & \begin{tabular}[c]{@{}c@{}} Weevil damage\\ optimisation  \end{tabular}\\ \hline

\citet{arabameri2022flood} & \begin{tabular}[c]{@{}c@{}}Flood susceptibility mapping  \end{tabular}  & \begin{tabular}[c]{@{}c@{}} PSO and GA  \end{tabular}\\ \hline

\citet{singh2021nature} & \begin{tabular}[c]{@{}c@{}}Network coverage in\\ Wireless Sensor Networks (WSNs)  \end{tabular}  & \begin{tabular}[c]{@{}c@{}} Binary ant colony \\ Improved GA \\ Lion optimisation \end{tabular}\\ \hline

\citet{kotiyal2021ecs} & \begin{tabular}[c]{@{}c@{}}Node localisation in WSNs  \end{tabular}  & \begin{tabular}[c]{@{}c@{}}Enhance cuckoo search \end{tabular}\\ \hline

\citet{rahimi2021machine} & \begin{tabular}[c]{@{}c@{}}Hydrogen adsorption of activated carbons  \end{tabular}  & \begin{tabular}[c]{@{}c@{}}GA\end{tabular}\\ \hline

\citet{yerlikaya2020new} & \begin{tabular}[c]{@{}c@{}}Classification in presence\\ of outliers \end{tabular}  & \begin{tabular}[c]{@{}c@{}}Conic quadratic optimisation\end{tabular}\\ \hline

\citet{singh2019mathematical} & \begin{tabular}[c]{@{}c@{}}Minimising redundant\\ sensing by sensors \end{tabular}  & Binary ant colony\\ \hline

\citet{kropat2016multi} & \begin{tabular}[c]{@{}c@{}}For shortest path planning in dynamically\\ changing and uncertain environments \end{tabular}  & Slime mold-based optimisation\\ \hline

\citet{kuter2015inversion} & \begin{tabular}[c]{@{}c@{}}Atmospheric correction \\ of satellite images \end{tabular}  & \begin{tabular}[c]{@{}c@{}}Conic quadratic optimisation\end{tabular}\\ \hline

\citet{ozmen2014precipitation} & \begin{tabular}[c]{@{}c@{}}Precipitation modeling \end{tabular}  & \begin{tabular}[c]{@{}c@{}}Conic quadratic optimisation\end{tabular}\\ \hline

\citet{meyer2013intercepting} & \begin{tabular}[c]{@{}c@{}}For intercepting a mobile target \\ in urban areas \end{tabular}  & \begin{tabular}[c]{@{}c@{}}PSO and\\ Ant colony optimisation\end{tabular}\\ \hline
\end{tabular}
}
\label{tab:metaapplications}
\end{table}

\begin{figure}[ht!]
    \centering
    \includegraphics[width=\textwidth]{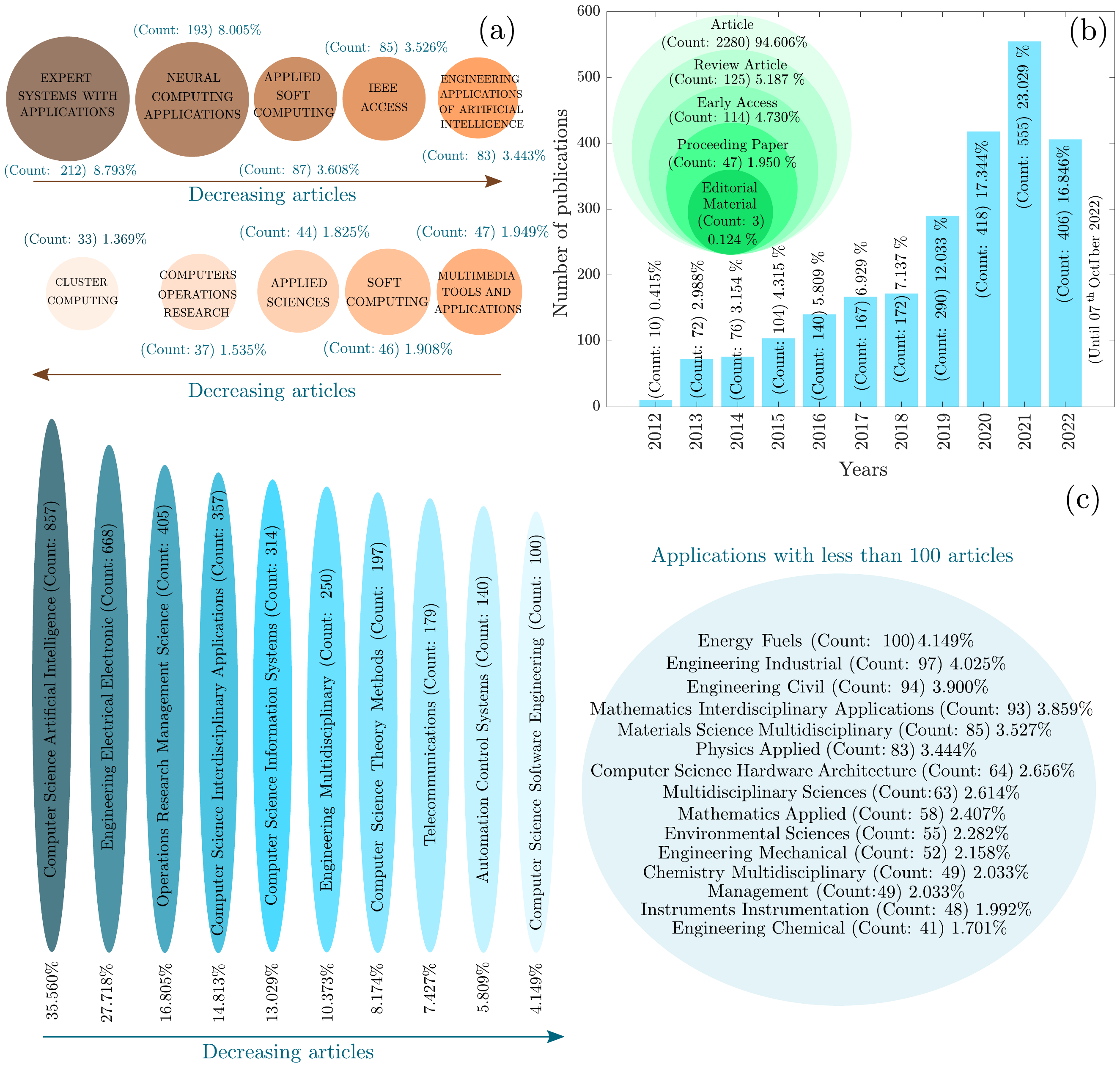}
    \caption{A brief review of the research contributions in Web Of Science (WoS) concerning application domain of meta-heuristic algorithms published in last ten years (from 2012-2022). (a) distribution of the top ten journals that publishes application of meta-heuristic algorithms, (b) the number of research publications each year and the type of research contribution, (c) the left block shows the top ten application domains, and the right block shows application domains having less than 100 articles.}
    \label{fig:review}
\end{figure}

Tabu search \citep{glover1986future}, harmony search \citep{geem2001new}, imperialist competitive algorithm \citep{atashpaz2007imperialist}, brain storm optimisation \citep{shi2015optimization}, teaching learning based optimisation \citep{rao2011teaching}, league championship algorithm \citep{kashan2014league}, heap based optimiser \citep{askari2020heap}, interactive autodidactic school \citep{jahangiri2020interactive}, l{\'e}vy flight distribution \citep{houssein2020levy}, most valuable player algorithm \citep{bouchekara2020most}, nomadic people optimiser\citep{salih2020new}, political optimiser \citep{askari2020political}, arithmetic optimisation \citep{abualigah2021arithmetic}, and stock exchange trading optimisation \citep{emami2022stock} comes under the human-inspired category. Among these, teaching learning-based optimisation has emerged as the most popular algorithm. It is based on simulating the influence of a teacher on the output of learners in a teaching-learning process. It is widely used to solve constrained and unconstrained optimisation problems, including some combinatorial optimisation. It is capable of handling the combinatorial optimisation problems \citep{baykasouglu2014testing}.

The aforementioned algorithms have gained significant traction in addressing intricate real-world optimisation challenges. This stems from their ability to efficiently explore extensive solution spaces, manage complex and nonlinear problem landscapes, adapt to dynamic environments, and yield near-optimal solutions without requiring complete problem-specific knowledge. Their versatility in handling complex optimisation challenges spans diverse fields, encompassing engineering, finance, healthcare, and logistics. We have tabulated a selection of notable optimisation problems successfully tackled using meta-heuristic and robust optimisation algorithms in the last ten years (Table \ref{tab:metaapplications}). To examine the application domains closely associated with meta-heuristic algorithms, we performed a bibliometric analysis of the literature published in the last ten years (i.e., from 2012-2022) with keywords meta-heuristic and applications. We considered 2410 research publications that are indexed in Web Of Science (WoS). We found that the Expert System with Applications, Neural Computing Applications, Applied Soft Computing, IEEE Access, and Engineering Application of Artificial Intelligence are the primary venues for publishing meta-heuristic-based application work (Figure \ref{fig:review}a). The majority of the publications were original research articles (94.6\%) with 5.2\% review articles, 4.7\% early access articles, 2\% conference proceedings, and 0.1\% editorial materials (Figure \ref{fig:review}b).  We observed an exponential increase in the number of publications over the years, with approximately 60\% publications in the last three years. This highlights the importance of meta-heuristic algorithms in solving application-based problems (Figure \ref{fig:review}b). Finally, we plotted all the application domains in which meta-heuristic algorithms play a significant role, along with the corresponding share of publications (Figure \ref{fig:review}c). The major research domains are computer science (59.6\%), engineering (47.3\%), operations research management science (16.8\%), mathematics (7.5 \%), telecommunications (7.5 \%), automation control systems (5.8\%), and physics (4.7\%).

\section{Scorpion hunting strategy (SHS)}
\label{sec:shs-alg}

\begin{figure}
    \centering
    \includegraphics[width=\textwidth]{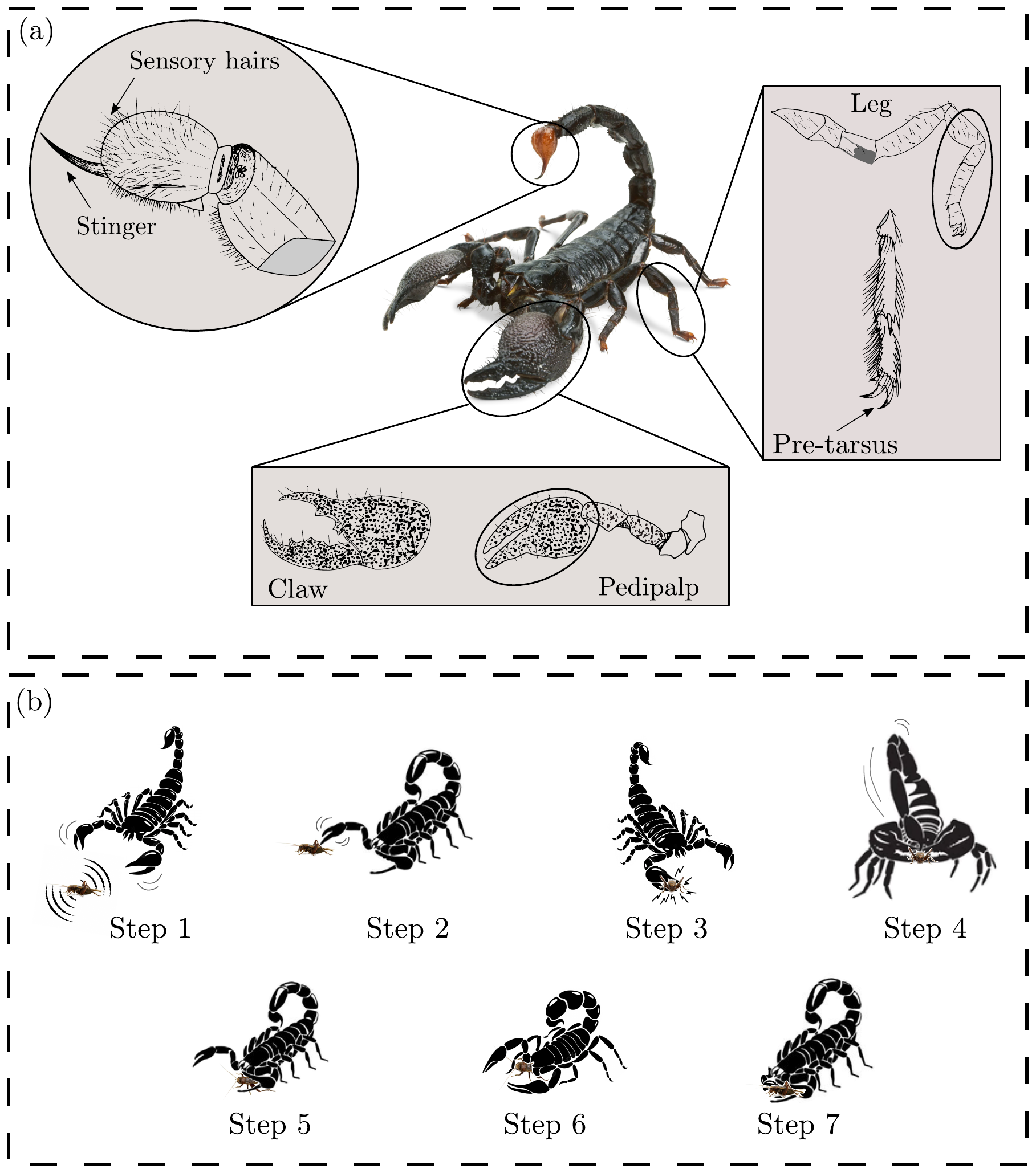}
    \caption{(a) Illustration of distinct hunting body parts of the scorpion, including the sting (tail), claw (pedipalp), and pre-tarsus (leg) (modified after \citep{polis1990biology}), (b) Sequential depiction of the scorpion's hunting strategy (Step 1: body orientation, Step 2: forwarding claw to capture prey, Step 3: capture prey, Step 4: sting action, Step 5: biting prey, Step 6: transport to the desired location, and Step 7: ingestion process).}
    \label{fig:scorpion_hunting_part_steps_merged}
\end{figure}

The Scorpion Hunting Strategy (SHS) algorithm is inspired by the hunting strategy of scorpions. Unlike other creatures that use their vision to locate prey, scorpions use a special neurosensory system that efficiently integrates the sophisticated senses to produce well-coordinated movements in locating and attacking the prey (such as spiders, crickets, pill bugs, and scorpions). They make use of the slit sensilla organ (present in the basitarsal leg part) to sense the vibration created by the prey to precisely locate their position. According to Brownell’s experiments (conducted with sand-dune scorpions), prey that is located at a distance of 15 cm can be grabbed by a single move, whereas a prey located at a distance of 30 cm requires a series of orientation movements \citep{brownell1979detection,brownell1979prey}. The vibration starts becoming stronger as the scorpion moves toward the prey \citep{brownell2001vibration}. They are capable of hunting by precisely locating the prey position that is situated up to 50 cm away \citep{brownell1977compressional}. However, for the larger distance, they can only precisely locate the direction of the prey \citep{brownell1979orientation}. Once the scorpion locates the prey, they generally use its sting, claw, and pre-tarsus to neutralise the prey (Figure \ref{fig:scorpion_hunting_part_steps_merged}a).

The hunting and ingestion process of a scorpion can be categorised into seven steps (Figure \ref{fig:scorpion_hunting_part_steps_merged}b). The first step is to locate the prey by sensing the vibration and adjusting body orientation by aligning themselves with the prey. In the second and third steps, the scorpion put forward its claws to grab and capture the prey. Depending upon the prey's size or resisting power, the scorpion can sting the prey once or twice in the fourth step. The scorpion crushes the prey with its claw and starts the ingestion process by eating the head of the prey. For smooth and uninterrupted ingestion, the scorpions sometimes carry their prey to a safer place near their burrow \citep{polis1990biology,stockmann2010scorpions}.

\begin{figure}[ht]
    \centering
    \includegraphics[width=0.8\textwidth]{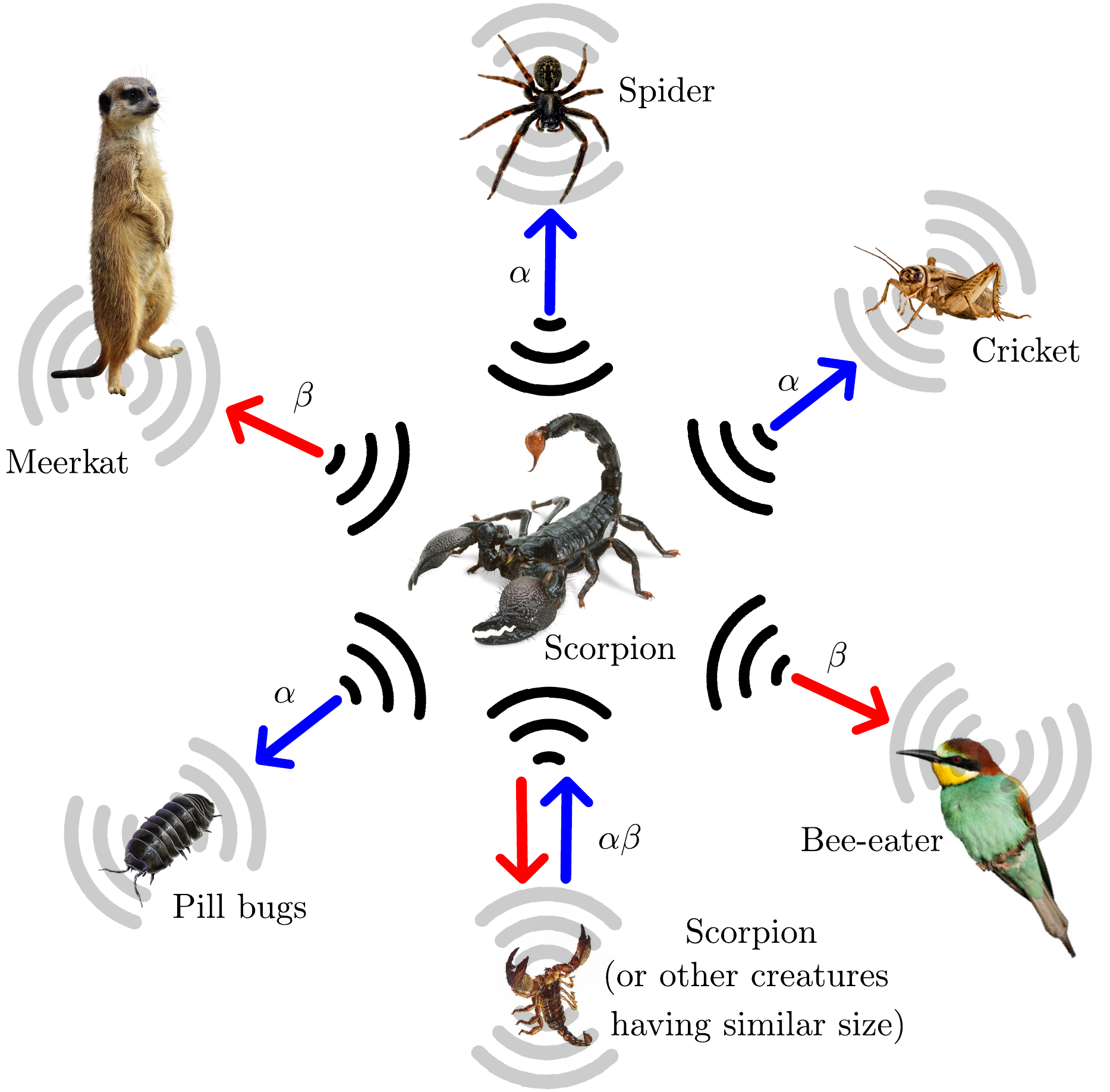}
    \caption{Illustration of the working of $\alpha$ and $\beta$ operator in response to prey or hunter, respectively.}
    \label{fig:shs_interaction}
\end{figure}
\subsection{Algorithm design}
Similar to any population-based meta-heuristic, the SHS algorithm also starts with the initialisation of the population. We initialise the population as $x_i$, where $i$ ranges from 1 to $n$, with $n$ representing the total population, a combination of scorpions and other animals (prey or hunter). Each individual in the population is regarded as a search agent (or candidate solution) whose position matrix is given by
\begin{equation}
  x_i =  \begin{bmatrix}
x_{i(1),1} & x_{i(1),2} & \hdots & x_{i(1),dim}\\
x_{i(2),1} & x_{i(2),2} & \hdots & x_{i(2),dim}\\
\vdots & \vdots & \vdots & \vdots\\
x_{i(n),1} & x_{i(n),2} & \hdots & x_{i(n),dim}
\end{bmatrix},
\end{equation}
where $dim$ is the dimension of $x_i$ variable. Initially, the position for each candidate solution is randomly generated using a continuous random uniform distribution (\textit{unifrnd}) having a lower end and upper end, denoted by $LE$ and $UE$, respectively:
\begin{equation}
    x_i^j= unifrnd \times (UE^j - LE^j) + LE^j,
\end{equation}
where $x_i^j$ represent the position of the $i^{th}$ candidate in the $j^{th}$ dimension. $j$ ranges from 1 to $dim$. $LE^j$ and $UE^j$ represent the lower end and upper end of the $j^{th}$ candidate of the optimised problem. The corresponding cost (or fitness) value of the candidate solution is given by
\begin{equation}
F_x  = \begin{bmatrix}
    f(x_{1,1}, x_{1,2}, \hdots, x_{1,dim})\\f(x_{2,1}, x_{2,2}, \hdots, x_{2,dim})\\ \vdots \\f(x_{n,1}, x_{n,2}, \hdots, x_{n,dim})
    \end{bmatrix},
\end{equation}
where $f(\cdot)$ is the cost function of the optimisation problem. Considering the minimisation optimisation task, the candidate with minimum value is considered as the best solution, denoted by $x_{best}$. The $x_{best}$ is calculated by sorting and ranking the population using Equation \ref{eq:xbest}:
\begin{equation}
x_{best} = \text{argmin}(F_x).
    \label{eq:xbest}
\end{equation}

A vibration signal defines the relation between two population members: if a scorpion member goes toward a prey member, it is called $\alpha$ vibration, and if a scorpion goes toward a hunter member, it is called $\beta$ vibration. If a scorpion member goes toward another scorpion, then it is the mutation of $\alpha\beta$ vibration (Figure \ref{fig:shs_interaction}). The cost function decides the chances of whether the scorpion meets either with prey or hunter. However, the probability of a scorpion meeting another scorpion (i.e., during $\alpha\beta$ mutation) is 20 \% ($\mu$ = $\alpha\beta$ = 0.2). In such scenarios, the scorpion having high claw ($\omega$) and sting ($\psi$) power will dominate. After any interaction among the scorpions, the signal vibration is updated by a damping ratio of $\mu damp$ = 0.98. The population members will continue to move toward each other upon detecting any vibration. However, the algorithm will work until the termination condition is reached (i.e., maximum iterations). After every iteration, $\alpha$ and $\beta$ are incremented by one unit (Algorithm 1).

\begin{algorithm}[ht]
\DontPrintSemicolon
\caption{SHS algorithm pseudo-code.}\label{alg:shs}
\Kwinitialization{Generating population of scorpions and preys ($x_i$ = 1, 2, ..., $n$)}
\kwdefine{Vibration {$\alpha$} in {$x_i$}} 
\kwdefine{ Vibration {$\beta$} in {$x_i$}} 
\kwdefine{ Claw Power coefficient of {$\omega$} in $x_i$ \Comment*[r]{random range in [0.1 to 0.3]}}
\kwdefine{ Sting Power coefficient of $\psi$ in $x_i$ \Comment*[r]{ random range in [0.1 to 0.3]}}
\kwdefine{ $\alpha \beta$ Mutation Rate of $\mu$ = 0.2}
\While{iteration $\leq$ \text{iteration}$_{\text{max}}$}{
   \For{$i = 1$ to $n$} {
   \For{$j = 1$ to $n$} {
        
    \While{$\alpha$ $>$ 0 or $\beta$ $>$ 0}{
     scorpion i goes toward prey j \\
     $\alpha$ = $\alpha$ +1\\
     $\beta$  = $\beta$ +1\\
    }
     Apply $\alpha \beta$ mutation of $\mu$ \\
 \text{Evaluate new solutions by cost function and update Vibration}\\
    }}
  \text{Sort and rank population and find the current global best}\\
  \text{Generating new generation}
}
\end{algorithm}

\begin{figure}[ht]
    \centering
    \includegraphics[width=0.8\textwidth]{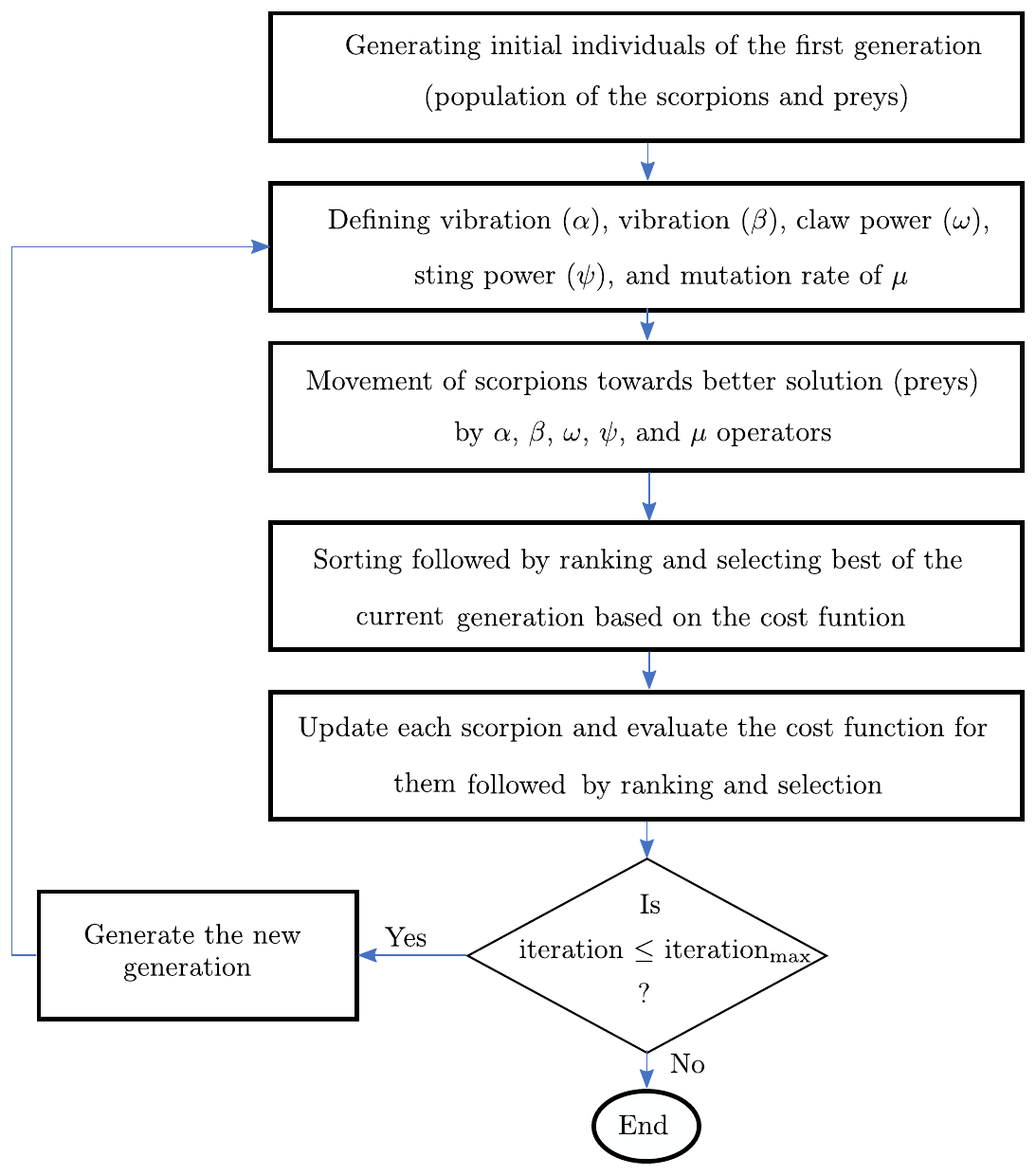}
    \caption{Flowchart of the SHS algorithm.}
    \label{fig:shs_flowchart}
\end{figure}

The movement of the population toward each other is calculated by the signal or vibration absorption coefficient according to Equation \ref{eq:move}:
\begin{equation}
    M = \psi \times e^{- \omega \times r_{ij}},
    \label{eq:move}
\end{equation}
where $r_{ij}$ represents the distance between two members of the population. $M$ represents the vibration absorption coefficient from one point to another (i.e.,  from the scorpion’s location to the prey’s location). A lower value of $M$ indicates that the distance between the two members is higher because the signal vibration is weaker. The value of $M$ will increase as the two members approach each other upon detecting the vibration. The updated position ($x_{i,new}$) of the scorpion is calculated according to Equation \ref{eq:newposition}:
\begin{equation}
    x_{i,new} = x_i + M (x_j - x_i) + \mu \times e_{div},
    \label{eq:newposition}
\end{equation}
where $x_i$ is the previous location, $\mu$ is the $\alpha \beta$ mutation, and $e_{div}$ is the diversity factor to incorporate the diversity along the movement path. This attribute makes the algorithm more dynamic and less linear. It is a uniform random vector given by Equation \ref{eq:7revised}:
\begin{equation}
    e_{div} = \delta \times unifrnd \times (-1, +1),
    \label{eq:7revised}
\end{equation}
where $\delta$ is the uniform mutation range calculated as 0.05 $\times$ ($UE$ - $LE$). At the end, the $\alpha \beta$ mutation is updated according to Equation \ref{eq:damping}:
\begin{equation}
    \mu = \mu \times \mu damp.
    \label{eq:damping}
\end{equation}

The best solution is obtained by sorting and ranking the cost function corresponding to the new position, according to Equation \ref{eq:best_final}:
\begin{equation}
    x_{best} = \text{argmin}(f(x_{i,new})).
    \label{eq:best_final}
\end{equation}

The pseudo-code of the proposed SHS algorithm is illustrated in Algorithm 1, and a detailed flowchart is shown in Figure \ref{fig:shs_flowchart}.


\subsection{Computational complexity}
Computational complexity is an important metric to measure the efficiency of any algorithm. The complexity of any algorithm depends on various factors, such as the total number of members in the population, the number of iterations, and the sorting and ranking of the individuals. The time complexity of the proposed SHS algorithm is discussed below:
\begin{enumerate}
    \item SHS takes $\mathcal{O}(n \times dim)$ time for the initialisation of population members, where $n$ represents the total population and $dim$ represents the dimension of the variable.
    \item It takes $\mathcal{O}(n)$ to compute fitness value of the population. 
    \item It takes $\mathcal{O}(\text{iteration}_{\text{max}} \times n)$ time to compute the fitness (or cost) value of each member, where $\text{iteration}_{\text{max}}$ is the maximum iteration.
    \item It takes $\mathcal{O}(n\log{}n)$ time to perform the sorting and ranking operation.
\end{enumerate}

The total time complexity of the proposed SHS algorithm is approximately $\mathcal{O}(n \times (1+dim+\text{iteration}_{\text{max}}+\log{}n))$. The space complexity of the SHS is $\mathcal{O}(n \times dim)$ because the process of initialisation of the population is considered to occupy the maximum space.

\subsection{Benchmark test functions and algorithms}
\label{sec:benchmark_sim}
To test the performance of the proposed SHS algorithm, we have used ten benchmark functions, which include unimodal (Zakharov, Booth, DeJong, Beale, Powell, and Trid) and multi-modal (Ackley, Rastrigin, Michalewicz, and Levy) functions \citep{ezugwu2022prairie,agushaka2023gazelle,kadkhoda2023novel,mohamed2020evaluating,kumar2023chaotic,yuan2023coronavirus,kumar2020test}.
Unimodal functions consist of a single global optimum. They are suitable for testing convergence accuracy and speed. The multi-modal functions consist of many local optima and are helpful in exploring the local optimum avoidance capability of an algorithm. Table \ref{tab:benchmarkfunction} summarises the mathematical formulation and key features of all the benchmark functions. These functions are grouped into different classes based on the similarities in their shapes and physical properties (i.e., many local minima, bowl-shaped, plate-shaped, valley-shaped, and ridge-shaped).

\begin{table}[t]
\caption{Details of the benchmark experiment optimisation test functions. $dim$ represents the dimension.}
 \resizebox{\textwidth}{!}{
\begin{tabular}{ccccc}
\hline
\textbf{Functions} & \textbf{Features} & \textbf{Equation} & \textbf{Global minima} & \textbf{Search domain} \\ \hline
\textbf{Ackley} & \begin{tabular}[c]{@{}c@{}} Multi-modal,\\ Many local minima \end{tabular} & \begin{tabular}[c]{@{}c@{}} $f(x)= -20 e^{\bigl(-0.02\sqrt{dim^{-1}\sum_{i=1}^{dim}}x_i^2\bigl)}$\\$-e^{\bigl(dim^{-1}\sum_{i=1}^{dim} cos(2\pi x_i)\bigl)}+20$  \end{tabular}     & $x^*= (0, ..., 0), f(x^*)=0$ & -35 $\leq$ $x_i$ $\leq$ 35 \\ \hline
\textbf{Rastrigin} & \begin{tabular}[c]{@{}c@{}} Multi-modal,\\ Many local minima \end{tabular}& $f(x)=10 dim + \sum_i^{dim} \bigl[x_i^2-10 cos(2\pi x_i)\bigl] $ & $x^*= (0, ..., 0), f(x^*)=0$ & -5.12 $\leq$ $x_i$ $\leq$ 5.12 \\ \hline
\textbf{Zakharov} & \begin{tabular}[c]{@{}c@{}} Unimodal,\\ Plate-shaped \end{tabular} & \begin{tabular}[c]{@{}c@{}} $f(x) = \sum_{i=1}^{dim} x_i^2 + \biggl(\frac{1}{2}\sum_{i=1}^{dim} ix_i\biggl)^2$ \\$+ \biggl(\frac{1}{2}\sum_{i=1}^{dim} ix_i\biggl)^4$ \end{tabular} & $x^*= (0, ..., 0), f(x^*)=0$ & -5 $\leq$ $x_i$ $\leq$ 10 \\ \hline
\textbf{Booth} & \begin{tabular}[c]{@{}c@{}} Unimodal,\\ Plate-shaped \end{tabular} & $f(x) = (x_1+x_2-7)^2+(2x_1+x_2-5)^2$ &  $x^*= (1, 3), f(x^*)=0$  & -10 $\leq$ $x_i$ $\leq$ 10 \\ \hline
\textbf{DeJong} & \begin{tabular}[c]{@{}c@{}} Unimodal,\\ Bowl-shaped \end{tabular} & $f(x) = \sum_{i=1}^{dim} x_i^2$ & $x^*= (0, ..., 0), f(x^*)=0$ & -10 $\leq$ $x_i$ $\leq$ 10 \\ \hline
\textbf{Beale} & \begin{tabular}[c]{@{}c@{}} Unimodal,\\ Plate-shaped \end{tabular} & \begin{tabular}[c]{@{}c@{}} $f(x)=(1.5-x_1-x_1 x_2 )^2+(2.25-x_1-x_1 x_2^2 )^2$\\$+(2.625-x_1-x_1 x_2^3 )^2$ \end{tabular} & $x^*= (3,0.5), f(x^*)=0$ & -4.5 $\leq$ $x_i$ $\leq$ 4.5 \\ \hline
\textbf{Powell} & \begin{tabular}[c]{@{}c@{}} Unimodal,\\ Valley-shaped \end{tabular} & \begin{tabular}[c]{@{}c@{}}$f(x) = \sum_{i=1}^{dim/4} \bigl(x_{4i-3}+10x_{4i-2} \bigl)^2+5\bigl(x_{4i-1}+x_{4i} \bigl)^2$\\$+\bigl(x_{4i-2}+x_{4i-1} \bigl)^4$ $+10\bigl(x_{4i-3}+x_{4i} \bigl)^4$\end{tabular}   & \begin{tabular}[c]{@{}c@{}} $x^*=(3,-1,0,1, ..., 3,-1,0,1),$\\$ f(x^*)=0$ \end{tabular}& -4 $\leq$ $x_i$ $\leq$ 5 \\ \hline
\textbf{Michalewicz} & \begin{tabular}[c]{@{}c@{}} Multi-modal,\\ Steep ridges \end{tabular} & $f(x) = - \sum_{i=1}^{dim} sin(x_i) \biggl(sin\biggl(\frac{ix_i^2}{\pi}\biggl)\biggl)^{2m} $ ($m=10$) & \begin{tabular}[c]{@{}c@{}} $x^*=(2.203,1.571),$\\ $f(x^*) = -1.801 $ ($n$ = 2) \end{tabular}& 0 $\leq$ $x_i$ $\leq$ $\pi$  \\ \hline
\textbf{Trid} &\begin{tabular}[c]{@{}c@{}} Unimodal,\\ Bowl-shaped \end{tabular} & $f(x)=\sum_{i=1}^{dim} (x_i-1)^2 - \sum_{i=2}^D x_ix_{i-1}$ & $f(x^*) = -50$ & -36 $\leq$ $x_i$ $\leq$ 36 \\ \hline
\textbf{Levy} &\begin{tabular}[c]{@{}c@{}} Multi-modal,\\Many local minima \end{tabular}& \begin{tabular}[c]{@{}c@{}}$f(x)= sin^2(\pi w_1 + \sum_{i}^{{dim}-1} (w_i-1)^2) [1+10sin^2(\pi w_i+1)]$ \\ $+(w_{dim}-1)^2[1+sin^2(2\pi w_{dim})]$ where $w_i = 1+\frac{x_i-1}{4}$ \end{tabular}  & $x^*= (1, ..., 1), f(x^*)=0$ & -10 $\leq$ $x_i$ $\leq$ 10 \\ \hline
\end{tabular}}
\label{tab:benchmarkfunction}
\end{table}

\begin{figure}[t!]
    \centering
    \includegraphics[width=\textwidth]{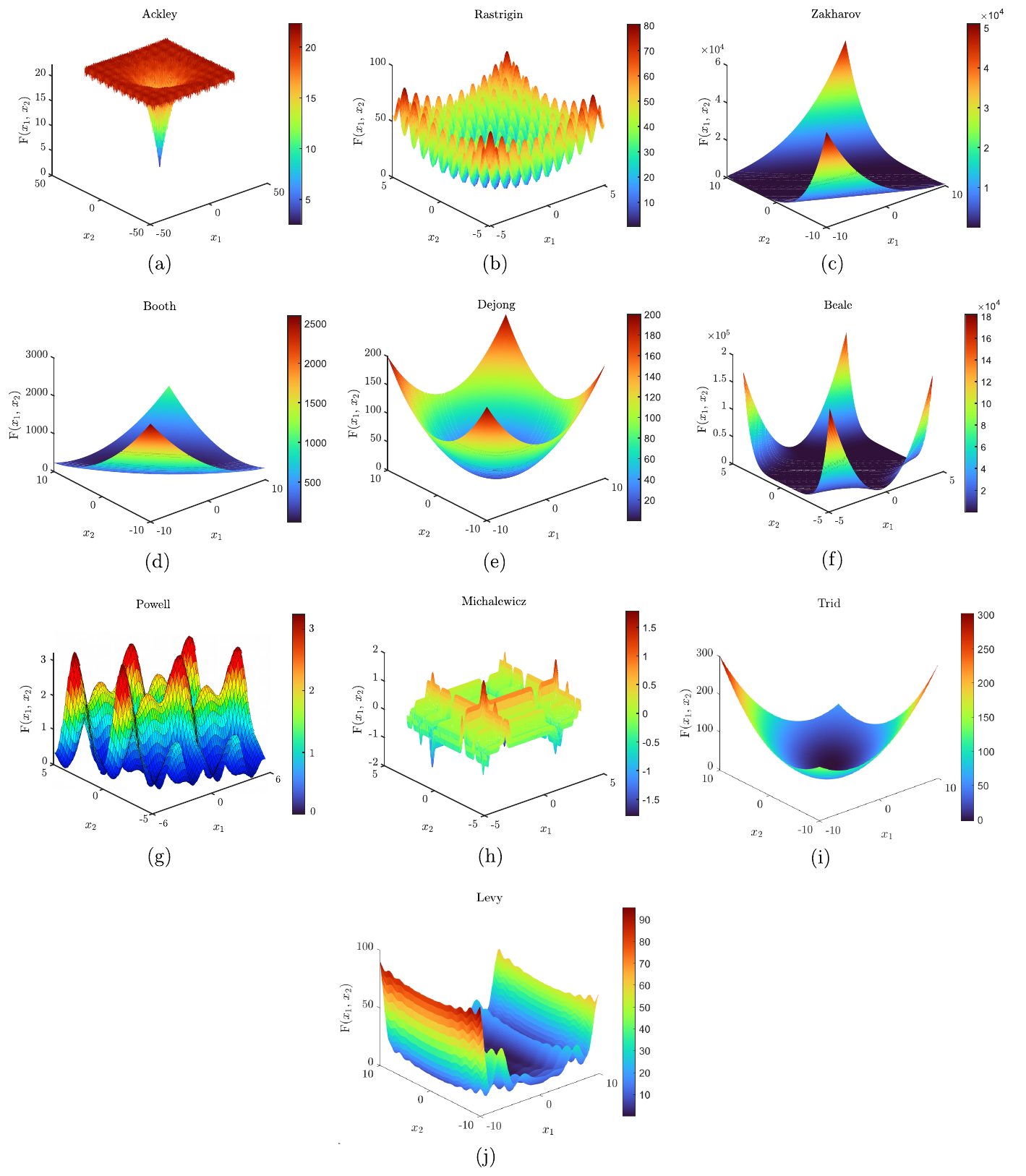}
    \caption{Illustration of the benchmark experiment optimisation functions for evaluation in three dimensions.}
    \label{fig:function_final}
\end{figure}

Ackley, also known as Ackley's path, is one of the most widely used benchmark functions \citep{ackley2012connectionist}. It is a non-convex function that is characterised by a large hole in the middle with a nearly flat outer section (Figure \ref{fig:function_final}a). It is differentiable, scalable, and non-separable. This function evaluates the risk of an algorithm getting stuck in one of its several local minima. Rastrigin function is also a non-convex function \citep{rastrigin1974systems}. It is differentiable, scalable, and separable. It is a highly multi-modal (non-linear) function that is characterised by a large number of local minima whose locations are regularly distributed (Figure \ref{fig:function_final}b). The vast search space in synergy with a large number of local minima makes it difficult to obtain the minimum of this function. Zakharov function is a convex unimodal function (plate-shaped) which is differentiable and non-separable \citep{muthiah2016galactic,jamil2013literature}. This function has only the global minimia (Figure \ref{fig:function_final}c). Booth is a convex unimodal function (plate-shaped) that is characterised by many local minima (Figure \ref{fig:function_final}d). It is differentiable, non-separable, and non-scalable. DeJong is a highly convex unimodal (bowl-shaped), also known as Sphere function \citep{schumer1968adaptive}. It does not have local minima except the global one (Figure \ref{fig:function_final}e). It is differentiable, scalable, and separable. Beale function is a non-convex multi-modal function (plate-shaped) having sharp peaks at all the corners (Figure \ref{fig:function_final}f). It is differentiable, non-separable, and non-scalable. Powell is one of the classical functions, which is a convex unimodal function (valley-shaped) (Figure \ref{fig:function_final}g). It is also known as Powell's singular function. It is scalable, differentiable, and non-separable \citep{powell1962iterative}. Michalewicz function is a ridge-shaped multi-modal function that is characterised by $dim$! (factorial of the dimension) number of local minima (Figure \ref{fig:function_final}h). The term $m$ decides the steepness of the ridges (Table \ref{tab:benchmarkfunction}). A high value of $m$ corresponds to a more difficult search (10 is the recommended value). It is differentiable, separable, and scalable. Trid is an unimodal convex function (bowl-shaped) that does not have local minima except the global one (Figure \ref{fig:function_final}i). It is differentiable, non-separable, and non-scalable. Levy is a non-convex multi-modal function with a large number of local minima (Figure \ref{fig:function_final}j). It is differentiable, non-separable, and scalable.

\begin{table}[ht!]
\label{tab:simulation_par}
\caption{Simulation parameters for all the algorithms.}
 \resizebox{\textwidth}{!}{
\begin{tabular}{|c|ccccccccccccc|}
\hline
\textbf{PARAMETERS} & \multicolumn{1}{c|}{\textbf{GA}} & \multicolumn{1}{c|}{\textbf{PSO}} & \multicolumn{1}{c|}{\textbf{FA}} & \multicolumn{1}{c|}{\textbf{BBO}} & \multicolumn{1}{c|}{\textbf{ABC}} & \multicolumn{1}{c|}{\textbf{SHS}} & \multicolumn{1}{c|}{\textbf{TLBO}} & \multicolumn{1}{c|}{\textbf{ACO}} & \multicolumn{1}{c|}{\textbf{HS}} & \multicolumn{1}{c|}{\textbf{SA}} & \multicolumn{1}{c|}{\textbf{DE}} & \multicolumn{1}{c|}{\textbf{BA}} & \textbf{IWO} \\ \hline
\textbf{Decision Variables (DV)} & \multicolumn{1}{c|}{20} & \multicolumn{1}{c|}{20} & \multicolumn{1}{c|}{20} & \multicolumn{1}{c|}{20} & \multicolumn{1}{c|}{20} & \multicolumn{1}{c|}{20} & \multicolumn{1}{c|}{20} & \multicolumn{1}{c|}{20} & \multicolumn{1}{c|}{20} & \multicolumn{1}{c|}{20} & \multicolumn{1}{c|}{20} & \multicolumn{1}{c|}{20} & 20 \\ \hline
\textbf{Decision Variables Size} & \multicolumn{1}{c|}{{[}1, 20{]}} & \multicolumn{1}{c|}{{[}1, 20{]}} & \multicolumn{1}{c|}{{[}1, 20{]}} & \multicolumn{1}{c|}{{[}1, 20{]}} & \multicolumn{1}{c|}{{[}1, 20{]}} & \multicolumn{1}{c|}{{[}1, 20{]}} & \multicolumn{1}{c|}{{[}1, 20{]}} & \multicolumn{1}{c|}{{[}1, 20{]}} & \multicolumn{1}{c|}{{[}1, 20{]}} & \multicolumn{1}{c|}{{[}1, 20{]}} & \multicolumn{1}{c|}{{[}1, 20{]}} & \multicolumn{1}{c|}{{[}1, 20{]}} & {[}1, 20{]} \\ \hline
\textbf{Lower Bound   of Variables (L)} & \multicolumn{1}{c|}{-10} & \multicolumn{1}{c|}{-10} & \multicolumn{1}{c|}{-10} & \multicolumn{1}{c|}{-10} & \multicolumn{1}{c|}{-10} & \multicolumn{1}{c|}{-10} & \multicolumn{1}{c|}{-10} & \multicolumn{1}{c|}{-10} & \multicolumn{1}{c|}{-10} & \multicolumn{1}{c|}{-10} & \multicolumn{1}{c|}{-10} & \multicolumn{1}{c|}{-10} & -10 \\ \hline
\textbf{Upper Bound of Variables (U)} & \multicolumn{1}{c|}{10} & \multicolumn{1}{c|}{10} & \multicolumn{1}{c|}{10} & \multicolumn{1}{c|}{10} & \multicolumn{1}{c|}{10} & \multicolumn{1}{c|}{10} & \multicolumn{1}{c|}{10} & \multicolumn{1}{c|}{10} & \multicolumn{1}{c|}{10} & \multicolumn{1}{c|}{10} & \multicolumn{1}{c|}{10} & \multicolumn{1}{c|}{10} & 10 \\ \hline
\textbf{Iterations} & \multicolumn{1}{c|}{300} & \multicolumn{1}{c|}{300} & \multicolumn{1}{c|}{300} & \multicolumn{1}{c|}{300} & \multicolumn{1}{c|}{300} & \multicolumn{1}{c|}{300} & \multicolumn{1}{c|}{300} & \multicolumn{1}{c|}{300} & \multicolumn{1}{c|}{300} & \multicolumn{1}{c|}{300} & \multicolumn{1}{c|}{300} & \multicolumn{1}{c|}{300} & 300 \\ \hline
\textbf{Population Size (P)} & \multicolumn{1}{c|}{25} & \multicolumn{1}{c|}{25} & \multicolumn{1}{c|}{25} & \multicolumn{1}{c|}{25} & \multicolumn{1}{c|}{25} & \multicolumn{1}{c|}{25} & \multicolumn{1}{c|}{25} & \multicolumn{1}{c|}{25} & \multicolumn{1}{c|}{25} & \multicolumn{1}{c|}{25} & \multicolumn{1}{c|}{25} & \multicolumn{1}{c|}{25} & 25 \\ \hline
\textbf{Crossover Percentage (PC)} & \multicolumn{1}{c|}{0.7} & \multicolumn{1}{c|}{-} & \multicolumn{1}{c|}{-} & \multicolumn{1}{c|}{-} & \multicolumn{1}{c|}{-} & \multicolumn{1}{c|}{-} & \multicolumn{1}{c|}{-} & \multicolumn{1}{c|}{-} & \multicolumn{1}{c|}{-} & \multicolumn{1}{c|}{-} & \multicolumn{1}{c|}{-} & \multicolumn{1}{c|}{-} & - \\ \hline
\textbf{Number of Offspring’s   (Parents)} & \multicolumn{1}{c|}{2$\times$(PC$\times$P/2)} & \multicolumn{1}{c|}{-} & \multicolumn{1}{c|}{-} & \multicolumn{1}{c|}{-} & \multicolumn{1}{c|}{-} & \multicolumn{1}{c|}{-} & \multicolumn{1}{c|}{-} & \multicolumn{1}{c|}{-} & \multicolumn{1}{c|}{-} & \multicolumn{1}{c|}{-} & \multicolumn{1}{c|}{-} & \multicolumn{1}{c|}{-} & - \\ \hline
\textbf{Mutation Percentage (MP)} & \multicolumn{1}{c|}{0.3} & \multicolumn{1}{c|}{-} & \multicolumn{1}{c|}{-} & \multicolumn{1}{c|}{-} & \multicolumn{1}{c|}{-} & \multicolumn{1}{c|}{-} & \multicolumn{1}{c|}{-} & \multicolumn{1}{c|}{-} & \multicolumn{1}{c|}{-} & \multicolumn{1}{c|}{-} & \multicolumn{1}{c|}{-} & \multicolumn{1}{c|}{-} & - \\ \hline
\textbf{Number of Mutants} & \multicolumn{1}{c|}{MP$\times$P} & \multicolumn{1}{c|}{-} & \multicolumn{1}{c|}{-} & \multicolumn{1}{c|}{-} & \multicolumn{1}{c|}{-} & \multicolumn{1}{c|}{-} & \multicolumn{1}{c|}{-} & \multicolumn{1}{c|}{-} & \multicolumn{1}{c|}{-} & \multicolumn{1}{c|}{-} & \multicolumn{1}{c|}{-} & \multicolumn{1}{c|}{-} & - \\ \hline
\textbf{Mutation Rate} & \multicolumn{1}{c|}{0.2} & \multicolumn{1}{c|}{0.2} & \multicolumn{1}{c|}{0.2} & \multicolumn{1}{c|}{0.2} & \multicolumn{1}{c|}{0.2} & \multicolumn{1}{c|}{0.2} & \multicolumn{1}{c|}{0.2} & \multicolumn{1}{c|}{0.2} & \multicolumn{1}{c|}{0.2} & \multicolumn{1}{c|}{0.2} & \multicolumn{1}{c|}{0.2} & \multicolumn{1}{c|}{0.2} & 0.2 \\ \hline
\textbf{Inertia Weight} & \multicolumn{1}{c|}{-} & \multicolumn{1}{c|}{1} & \multicolumn{1}{c|}{-} & \multicolumn{1}{c|}{-} & \multicolumn{1}{c|}{-} & \multicolumn{1}{c|}{-} & \multicolumn{1}{c|}{-} & \multicolumn{1}{c|}{-} & \multicolumn{1}{c|}{-} & \multicolumn{1}{c|}{-} & \multicolumn{1}{c|}{-} & \multicolumn{1}{c|}{-} & - \\ \hline
\textbf{Inertia Weight Damping Ratio} & \multicolumn{1}{c|}{-} & \multicolumn{1}{c|}{0.99} & \multicolumn{1}{c|}{-} & \multicolumn{1}{c|}{-} & \multicolumn{1}{c|}{-} & \multicolumn{1}{c|}{-} & \multicolumn{1}{c|}{-} & \multicolumn{1}{c|}{-} & \multicolumn{1}{c|}{-} & \multicolumn{1}{c|}{-} & \multicolumn{1}{c|}{-} & \multicolumn{1}{c|}{-} & - \\ \hline
\textbf{Personal Learning Coefficient} & \multicolumn{1}{c|}{-} & \multicolumn{1}{c|}{1.5} & \multicolumn{1}{c|}{-} & \multicolumn{1}{c|}{-} & \multicolumn{1}{c|}{-} & \multicolumn{1}{c|}{-} & \multicolumn{1}{c|}{-} & \multicolumn{1}{c|}{-} & \multicolumn{1}{c|}{-} & \multicolumn{1}{c|}{-} & \multicolumn{1}{c|}{-} & \multicolumn{1}{c|}{-} & - \\ \hline
\textbf{Global Learning Coefficient} & \multicolumn{1}{c|}{-} & \multicolumn{1}{c|}{2} & \multicolumn{1}{c|}{-} & \multicolumn{1}{c|}{-} & \multicolumn{1}{c|}{-} & \multicolumn{1}{c|}{-} & \multicolumn{1}{c|}{-} & \multicolumn{1}{c|}{-} & \multicolumn{1}{c|}{-} & \multicolumn{1}{c|}{-} & \multicolumn{1}{c|}{-} & \multicolumn{1}{c|}{-} & - \\ \hline
\textbf{Light Absorption Coefficient} & \multicolumn{1}{c|}{-} & \multicolumn{1}{c|}{-} & \multicolumn{1}{c|}{1} & \multicolumn{1}{c|}{-} & \multicolumn{1}{c|}{-} & \multicolumn{1}{c|}{-} & \multicolumn{1}{c|}{-} & \multicolumn{1}{c|}{-} & \multicolumn{1}{c|}{-} & \multicolumn{1}{c|}{-} & \multicolumn{1}{c|}{-} & \multicolumn{1}{c|}{-} & - \\ \hline
\textbf{Attraction Coefficient} & \multicolumn{1}{c|}{-} & \multicolumn{1}{c|}{-} & \multicolumn{1}{c|}{2} & \multicolumn{1}{c|}{-} & \multicolumn{1}{c|}{-} & \multicolumn{1}{c|}{-} & \multicolumn{1}{c|}{-} & \multicolumn{1}{c|}{-} & \multicolumn{1}{c|}{-} & \multicolumn{1}{c|}{-} & \multicolumn{1}{c|}{-} & \multicolumn{1}{c|}{-} & - \\ \hline
\textbf{Mutation Damping Ratio} & \multicolumn{1}{c|}{-} & \multicolumn{1}{c|}{-} & \multicolumn{1}{c|}{0.98} & \multicolumn{1}{c|}{-} & \multicolumn{1}{c|}{-} & \multicolumn{1}{c|}{0.98} & \multicolumn{1}{c|}{-} & \multicolumn{1}{c|}{-} & \multicolumn{1}{c|}{-} & \multicolumn{1}{c|}{-} & \multicolumn{1}{c|}{-} & \multicolumn{1}{c|}{-} & - \\ \hline
\textbf{Keep Rate (KR)} & \multicolumn{1}{c|}{-} & \multicolumn{1}{c|}{-} & \multicolumn{1}{c|}{-} & \multicolumn{1}{c|}{0.2} & \multicolumn{1}{c|}{-} & \multicolumn{1}{c|}{-} & \multicolumn{1}{c|}{-} & \multicolumn{1}{c|}{-} & \multicolumn{1}{c|}{-} & \multicolumn{1}{c|}{-} & \multicolumn{1}{c|}{-} & \multicolumn{1}{c|}{-} & - \\ \hline
\textbf{Kept Habitats} & \multicolumn{1}{c|}{-} & \multicolumn{1}{c|}{-} & \multicolumn{1}{c|}{-} & \multicolumn{1}{c|}{KR$\times$P} & \multicolumn{1}{c|}{-} & \multicolumn{1}{c|}{-} & \multicolumn{1}{c|}{-} & \multicolumn{1}{c|}{-} & \multicolumn{1}{c|}{-} & \multicolumn{1}{c|}{-} & \multicolumn{1}{c|}{-} & \multicolumn{1}{c|}{-} & - \\ \hline
\textbf{New Habitats} & \multicolumn{1}{c|}{-} & \multicolumn{1}{c|}{-} & \multicolumn{1}{c|}{-} & \multicolumn{1}{c|}{P-KR} & \multicolumn{1}{c|}{-} & \multicolumn{1}{c|}{-} & \multicolumn{1}{c|}{-} & \multicolumn{1}{c|}{-} & \multicolumn{1}{c|}{-} & \multicolumn{1}{c|}{-} & \multicolumn{1}{c|}{-} & \multicolumn{1}{c|}{-} & - \\ \hline
\textbf{Emmigration Rates (ER)} & \multicolumn{1}{c|}{-} & \multicolumn{1}{c|}{-} & \multicolumn{1}{c|}{-} & \multicolumn{1}{c|}{0.2} & \multicolumn{1}{c|}{-} & \multicolumn{1}{c|}{-} & \multicolumn{1}{c|}{-} & \multicolumn{1}{c|}{-} & \multicolumn{1}{c|}{-} & \multicolumn{1}{c|}{-} & \multicolumn{1}{c|}{-} & \multicolumn{1}{c|}{-} & - \\ \hline
\textbf{Immigration Rates} & \multicolumn{1}{c|}{-} & \multicolumn{1}{c|}{-} & \multicolumn{1}{c|}{-} & \multicolumn{1}{c|}{1-(ER)} & \multicolumn{1}{c|}{-} & \multicolumn{1}{c|}{-} & \multicolumn{1}{c|}{-} & \multicolumn{1}{c|}{-} & \multicolumn{1}{c|}{-} & \multicolumn{1}{c|}{-} & \multicolumn{1}{c|}{-} & \multicolumn{1}{c|}{-} & - \\ \hline
\textbf{Onlooker Bees} & \multicolumn{1}{c|}{-} & \multicolumn{1}{c|}{-} & \multicolumn{1}{c|}{-} & \multicolumn{1}{c|}{-} & \multicolumn{1}{c|}{P} & \multicolumn{1}{c|}{-} & \multicolumn{1}{c|}{-} & \multicolumn{1}{c|}{-} & \multicolumn{1}{c|}{-} & \multicolumn{1}{c|}{-} & \multicolumn{1}{c|}{-} & \multicolumn{1}{c|}{-} & - \\ \hline
\textbf{Abandonment Limit Parameter} & \multicolumn{1}{c|}{-} & \multicolumn{1}{c|}{-} & \multicolumn{1}{c|}{-} & \multicolumn{1}{c|}{-} & \multicolumn{1}{c|}{0.6$\times$DV$\times$P} & \multicolumn{1}{c|}{-} & \multicolumn{1}{c|}{-} & \multicolumn{1}{c|}{-} & \multicolumn{1}{c|}{-} & \multicolumn{1}{c|}{-} & \multicolumn{1}{c|}{-} & \multicolumn{1}{c|}{-} & - \\ \hline
\textbf{Acceleration Coefficient Upper} & \multicolumn{1}{c|}{-} & \multicolumn{1}{c|}{-} & \multicolumn{1}{c|}{-} & \multicolumn{1}{c|}{-} & \multicolumn{1}{c|}{1} & \multicolumn{1}{c|}{-} & \multicolumn{1}{c|}{-} & \multicolumn{1}{c|}{-} & \multicolumn{1}{c|}{-} & \multicolumn{1}{c|}{-} & \multicolumn{1}{c|}{-} & \multicolumn{1}{c|}{-} & - \\ \hline
\textbf{Claw Power coefficient} & \multicolumn{1}{c|}{-} & \multicolumn{1}{c|}{-} & \multicolumn{1}{c|}{-} & \multicolumn{1}{c|}{-} & \multicolumn{1}{c|}{-} & \multicolumn{1}{c|}{{[}1 3{]}} & \multicolumn{1}{c|}{-} & \multicolumn{1}{c|}{-} & \multicolumn{1}{c|}{-} & \multicolumn{1}{c|}{-} & \multicolumn{1}{c|}{-} & \multicolumn{1}{c|}{-} & - \\ \hline
\textbf{Sting Power coefficient} & \multicolumn{1}{c|}{-} & \multicolumn{1}{c|}{-} & \multicolumn{1}{c|}{-} & \multicolumn{1}{c|}{-} & \multicolumn{1}{c|}{-} & \multicolumn{1}{c|}{{[}1 3{]}} & \multicolumn{1}{c|}{-} & \multicolumn{1}{c|}{-} & \multicolumn{1}{c|}{-} & \multicolumn{1}{c|}{-} & \multicolumn{1}{c|}{-} & \multicolumn{1}{c|}{-} & - \\ \hline
\textbf{Phromone Exponential Weight} & \multicolumn{1}{c|}{-} & \multicolumn{1}{c|}{-} & \multicolumn{1}{c|}{-} & \multicolumn{1}{c|}{-} & \multicolumn{1}{c|}{-} & \multicolumn{1}{c|}{-} & \multicolumn{1}{c|}{-} & \multicolumn{1}{c|}{1} & \multicolumn{1}{c|}{-} & \multicolumn{1}{c|}{-} & \multicolumn{1}{c|}{-} & \multicolumn{1}{c|}{-} & - \\ \hline
\textbf{Heuristic Exponential Weight} & \multicolumn{1}{c|}{-} & \multicolumn{1}{c|}{-} & \multicolumn{1}{c|}{-} & \multicolumn{1}{c|}{-} & \multicolumn{1}{c|}{-} & \multicolumn{1}{c|}{-} & \multicolumn{1}{c|}{-} & \multicolumn{1}{c|}{1} & \multicolumn{1}{c|}{-} & \multicolumn{1}{c|}{-} & \multicolumn{1}{c|}{-} & \multicolumn{1}{c|}{-} & - \\ \hline
\textbf{Evaporation Rate} & \multicolumn{1}{c|}{-} & \multicolumn{1}{c|}{-} & \multicolumn{1}{c|}{-} & \multicolumn{1}{c|}{-} & \multicolumn{1}{c|}{-} & \multicolumn{1}{c|}{-} & \multicolumn{1}{c|}{-} & \multicolumn{1}{c|}{0.5} & \multicolumn{1}{c|}{-} & \multicolumn{1}{c|}{-} & \multicolumn{1}{c|}{-} & \multicolumn{1}{c|}{-} & - \\ \hline
\textbf{Number of New Harmonies} & \multicolumn{1}{c|}{-} & \multicolumn{1}{c|}{-} & \multicolumn{1}{c|}{-} & \multicolumn{1}{c|}{-} & \multicolumn{1}{c|}{-} & \multicolumn{1}{c|}{-} & \multicolumn{1}{c|}{-} & \multicolumn{1}{c|}{-} & \multicolumn{1}{c|}{5} & \multicolumn{1}{c|}{-} & \multicolumn{1}{c|}{-} & \multicolumn{1}{c|}{-} & - \\ \hline
\textbf{Harmony Memory Consideration Rate} & \multicolumn{1}{c|}{-} & \multicolumn{1}{c|}{-} & \multicolumn{1}{c|}{-} & \multicolumn{1}{c|}{-} & \multicolumn{1}{c|}{-} & \multicolumn{1}{c|}{-} & \multicolumn{1}{c|}{-} & \multicolumn{1}{c|}{-} & \multicolumn{1}{c|}{0.9} & \multicolumn{1}{c|}{-} & \multicolumn{1}{c|}{-} & \multicolumn{1}{c|}{-} & - \\ \hline
\textbf{Pitch Adjustment Rate} & \multicolumn{1}{c|}{-} & \multicolumn{1}{c|}{-} & \multicolumn{1}{c|}{-} & \multicolumn{1}{c|}{-} & \multicolumn{1}{c|}{-} & \multicolumn{1}{c|}{-} & \multicolumn{1}{c|}{-} & \multicolumn{1}{c|}{-} & \multicolumn{1}{c|}{0.1} & \multicolumn{1}{c|}{-} & \multicolumn{1}{c|}{-} & \multicolumn{1}{c|}{-} & - \\ \hline
\textbf{Fret Width} & \multicolumn{1}{c|}{-} & \multicolumn{1}{c|}{-} & \multicolumn{1}{c|}{-} & \multicolumn{1}{c|}{-} & \multicolumn{1}{c|}{-} & \multicolumn{1}{c|}{-} & \multicolumn{1}{c|}{-} & \multicolumn{1}{c|}{-} & \multicolumn{1}{c|}{0.02$\times$(U-L)} & \multicolumn{1}{c|}{-} & \multicolumn{1}{c|}{-} & \multicolumn{1}{c|}{-} & - \\ \hline
\textbf{Fret Width Damp Ratio} & \multicolumn{1}{c|}{-} & \multicolumn{1}{c|}{-} & \multicolumn{1}{c|}{-} & \multicolumn{1}{c|}{-} & \multicolumn{1}{c|}{-} & \multicolumn{1}{c|}{-} & \multicolumn{1}{c|}{-} & \multicolumn{1}{c|}{-} & \multicolumn{1}{c|}{0.995} & \multicolumn{1}{c|}{-} & \multicolumn{1}{c|}{-} & \multicolumn{1}{c|}{-} & - \\ \hline
\textbf{Maximum Number   of Sub-iterations} & \multicolumn{1}{c|}{-} & \multicolumn{1}{c|}{-} & \multicolumn{1}{c|}{-} & \multicolumn{1}{c|}{-} & \multicolumn{1}{c|}{-} & \multicolumn{1}{c|}{-} & \multicolumn{1}{c|}{-} & \multicolumn{1}{c|}{-} & \multicolumn{1}{c|}{-} & \multicolumn{1}{c|}{10} & \multicolumn{1}{c|}{-} & \multicolumn{1}{c|}{-} & - \\ \hline
\textbf{Initial Temperature} & \multicolumn{1}{c|}{-} & \multicolumn{1}{c|}{-} & \multicolumn{1}{c|}{-} & \multicolumn{1}{c|}{-} & \multicolumn{1}{c|}{-} & \multicolumn{1}{c|}{-} & \multicolumn{1}{c|}{-} & \multicolumn{1}{c|}{-} & \multicolumn{1}{c|}{-} & \multicolumn{1}{c|}{5} & \multicolumn{1}{c|}{-} & \multicolumn{1}{c|}{-} & - \\ \hline
\textbf{Temperature Reduction   Rate} & \multicolumn{1}{c|}{-} & \multicolumn{1}{c|}{-} & \multicolumn{1}{c|}{-} & \multicolumn{1}{c|}{-} & \multicolumn{1}{c|}{-} & \multicolumn{1}{c|}{-} & \multicolumn{1}{c|}{-} & \multicolumn{1}{c|}{-} & \multicolumn{1}{c|}{-} & \multicolumn{1}{c|}{0.99} & \multicolumn{1}{c|}{-} & \multicolumn{1}{c|}{-} & - \\ \hline
\textbf{Lower Bound of Scaling Factor} & \multicolumn{1}{c|}{-} & \multicolumn{1}{c|}{-} & \multicolumn{1}{c|}{-} & \multicolumn{1}{c|}{-} & \multicolumn{1}{c|}{-} & \multicolumn{1}{c|}{-} & \multicolumn{1}{c|}{-} & \multicolumn{1}{c|}{-} & \multicolumn{1}{c|}{-} & \multicolumn{1}{c|}{-} & \multicolumn{1}{c|}{0.2} & \multicolumn{1}{c|}{-} & - \\ \hline
\textbf{Upper Bound   of Scaling Factor} & \multicolumn{1}{c|}{-} & \multicolumn{1}{c|}{-} & \multicolumn{1}{c|}{-} & \multicolumn{1}{c|}{-} & \multicolumn{1}{c|}{-} & \multicolumn{1}{c|}{-} & \multicolumn{1}{c|}{-} & \multicolumn{1}{c|}{-} & \multicolumn{1}{c|}{-} & \multicolumn{1}{c|}{-} & \multicolumn{1}{c|}{0.8} & \multicolumn{1}{c|}{-} & - \\ \hline
\textbf{Crossover Probability} & \multicolumn{1}{c|}{-} & \multicolumn{1}{c|}{-} & \multicolumn{1}{c|}{-} & \multicolumn{1}{c|}{-} & \multicolumn{1}{c|}{-} & \multicolumn{1}{c|}{-} & \multicolumn{1}{c|}{-} & \multicolumn{1}{c|}{-} & \multicolumn{1}{c|}{-} & \multicolumn{1}{c|}{-} & \multicolumn{1}{c|}{0.2} & \multicolumn{1}{c|}{-} & - \\ \hline
\textbf{No of   Selected Sites (NS)} & \multicolumn{1}{c|}{-} & \multicolumn{1}{c|}{-} & \multicolumn{1}{c|}{-} & \multicolumn{1}{c|}{-} & \multicolumn{1}{c|}{-} & \multicolumn{1}{c|}{-} & \multicolumn{1}{c|}{-} & \multicolumn{1}{c|}{-} & \multicolumn{1}{c|}{-} & \multicolumn{1}{c|}{-} & \multicolumn{1}{c|}{-} & \multicolumn{1}{c|}{0.5$\times$P} & - \\ \hline
\textbf{No of Selected Elite Sites} & \multicolumn{1}{c|}{-} & \multicolumn{1}{c|}{-} & \multicolumn{1}{c|}{-} & \multicolumn{1}{c|}{-} & \multicolumn{1}{c|}{-} & \multicolumn{1}{c|}{-} & \multicolumn{1}{c|}{-} & \multicolumn{1}{c|}{-} & \multicolumn{1}{c|}{-} & \multicolumn{1}{c|}{-} & \multicolumn{1}{c|}{-} & \multicolumn{1}{c|}{0.4$\times$NS} & - \\ \hline
\textbf{No of   Recruited Bees (Selected) (RS)} & \multicolumn{1}{c|}{-} & \multicolumn{1}{c|}{-} & \multicolumn{1}{c|}{-} & \multicolumn{1}{c|}{-} & \multicolumn{1}{c|}{-} & \multicolumn{1}{c|}{-} & \multicolumn{1}{c|}{-} & \multicolumn{1}{c|}{-} & \multicolumn{1}{c|}{-} & \multicolumn{1}{c|}{-} & \multicolumn{1}{c|}{-} & \multicolumn{1}{c|}{0.5$\times$P} & - \\ \hline
\textbf{No of Recruited Bees (Elite)} & \multicolumn{1}{c|}{-} & \multicolumn{1}{c|}{-} & \multicolumn{1}{c|}{-} & \multicolumn{1}{c|}{-} & \multicolumn{1}{c|}{-} & \multicolumn{1}{c|}{-} & \multicolumn{1}{c|}{-} & \multicolumn{1}{c|}{-} & \multicolumn{1}{c|}{-} & \multicolumn{1}{c|}{-} & \multicolumn{1}{c|}{-} & \multicolumn{1}{c|}{2$\times$RS} & - \\ \hline
\textbf{Neighborhood   Radius} & \multicolumn{1}{c|}{-} & \multicolumn{1}{c|}{-} & \multicolumn{1}{c|}{-} & \multicolumn{1}{c|}{-} & \multicolumn{1}{c|}{-} & \multicolumn{1}{c|}{-} & \multicolumn{1}{c|}{-} & \multicolumn{1}{c|}{-} & \multicolumn{1}{c|}{-} & \multicolumn{1}{c|}{-} & \multicolumn{1}{c|}{-} & \multicolumn{1}{c|}{0.1$\times$(U-L)} & - \\ \hline
\textbf{Neighborhood Radius Damp Rate} & \multicolumn{1}{c|}{-} & \multicolumn{1}{c|}{-} & \multicolumn{1}{c|}{-} & \multicolumn{1}{c|}{-} & \multicolumn{1}{c|}{-} & \multicolumn{1}{c|}{-} & \multicolumn{1}{c|}{-} & \multicolumn{1}{c|}{-} & \multicolumn{1}{c|}{-} & \multicolumn{1}{c|}{-} & \multicolumn{1}{c|}{-} & \multicolumn{1}{c|}{0.95} & - \\ \hline
\textbf{Minimum   Number of Seeds} & \multicolumn{1}{c|}{-} & \multicolumn{1}{c|}{-} & \multicolumn{1}{c|}{-} & \multicolumn{1}{c|}{-} & \multicolumn{1}{c|}{-} & \multicolumn{1}{c|}{-} & \multicolumn{1}{c|}{-} & \multicolumn{1}{c|}{-} & \multicolumn{1}{c|}{-} & \multicolumn{1}{c|}{-} & \multicolumn{1}{c|}{-} & \multicolumn{1}{c|}{-} & 0 \\ \hline
\textbf{Maximum Number of Seeds} & \multicolumn{1}{c|}{-} & \multicolumn{1}{c|}{-} & \multicolumn{1}{c|}{-} & \multicolumn{1}{c|}{-} & \multicolumn{1}{c|}{-} & \multicolumn{1}{c|}{-} & \multicolumn{1}{c|}{-} & \multicolumn{1}{c|}{-} & \multicolumn{1}{c|}{-} & \multicolumn{1}{c|}{-} & \multicolumn{1}{c|}{-} & \multicolumn{1}{c|}{-} & 5 \\ \hline
\textbf{Variance   Reduction Exponent} & \multicolumn{1}{c|}{-} & \multicolumn{1}{c|}{-} & \multicolumn{1}{c|}{-} & \multicolumn{1}{c|}{-} & \multicolumn{1}{c|}{-} & \multicolumn{1}{c|}{-} & \multicolumn{1}{c|}{-} & \multicolumn{1}{c|}{-} & \multicolumn{1}{c|}{-} & \multicolumn{1}{c|}{-} & \multicolumn{1}{c|}{-} & \multicolumn{1}{c|}{-} & 2 \\ \hline
\textbf{Initial Value of Standard Deviation} & \multicolumn{1}{c|}{-} & \multicolumn{1}{c|}{-} & \multicolumn{1}{c|}{-} & \multicolumn{1}{c|}{-} & \multicolumn{1}{c|}{-} & \multicolumn{1}{c|}{-} & \multicolumn{1}{c|}{-} & \multicolumn{1}{c|}{-} & \multicolumn{1}{c|}{-} & \multicolumn{1}{c|}{-} & \multicolumn{1}{c|}{-} & \multicolumn{1}{c|}{-} & 0.5 \\ \hline
\textbf{Final Value   of Standard Deviation} & \multicolumn{1}{c|}{-} & \multicolumn{1}{c|}{-} & \multicolumn{1}{c|}{-} & \multicolumn{1}{c|}{-} & \multicolumn{1}{c|}{-} & \multicolumn{1}{c|}{-} & \multicolumn{1}{c|}{-} & \multicolumn{1}{c|}{-} & \multicolumn{1}{c|}{-} & \multicolumn{1}{c|}{-} & \multicolumn{1}{c|}{-} & \multicolumn{1}{c|}{-} & 0.001 \\ \hline
\end{tabular}}
\label{Tab:simulation_para}
\end{table}

\begin{figure}[ht!]
    \centering
    \includegraphics[width=\textwidth]{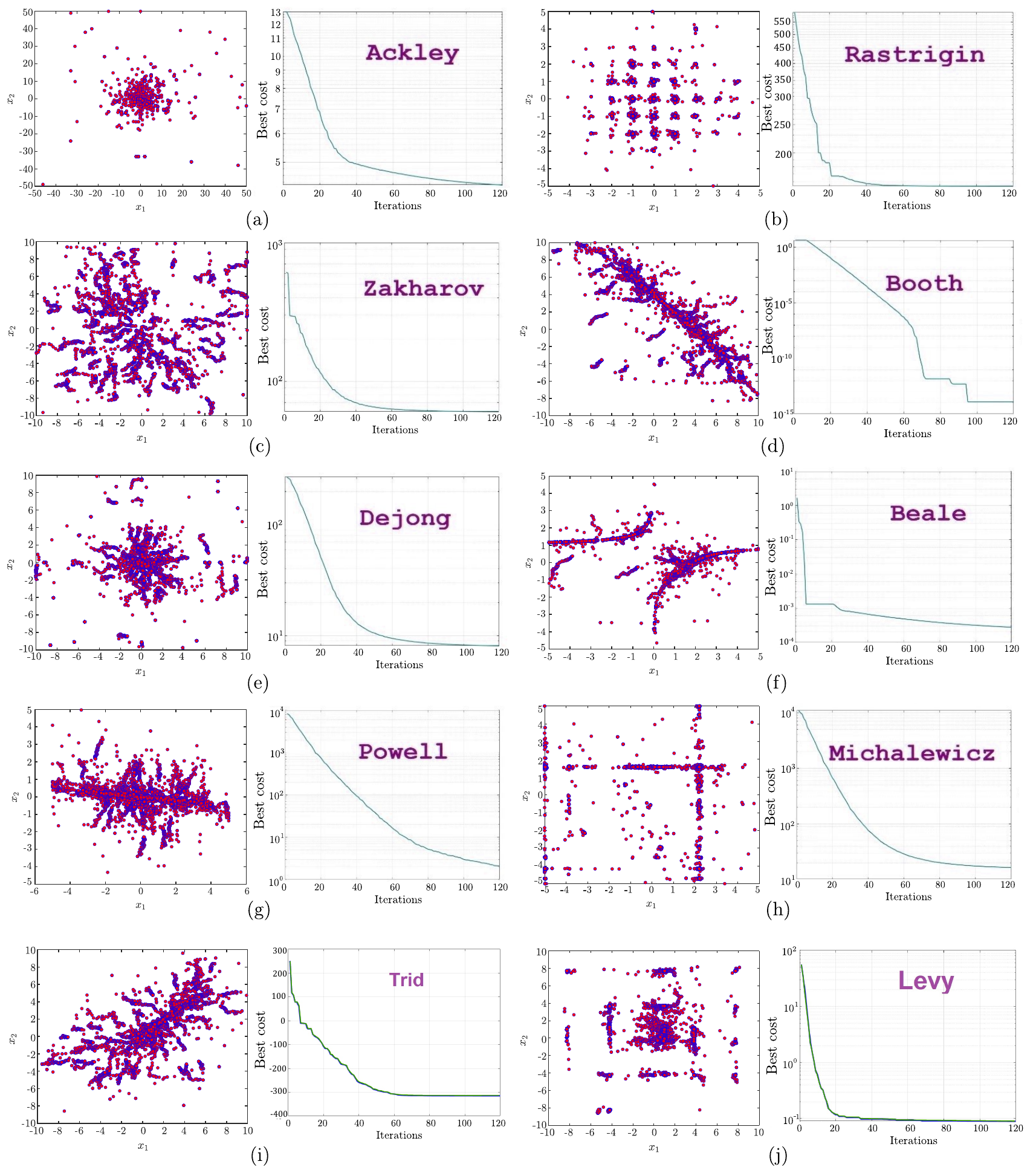}
    \caption{Illustrating the search history (i.e., the position of the scorpions) and the corresponding best cost curve for all the experiment optimisation functions.}
    \label{fig:scorpion_position}
\end{figure}

We compared the performance of the SHS with twelve meta-heuristic algorithms, namely GA \citep{holland1992adaptation}, PSO \citep{kennedy1995particle,eberhart1995particle,shi1999empirical}, FA \citep{yang2009firefly,yang2010eagle,yang2010firefly}, Biogeography-based Optimisation (BBO) \citep{simon2008biogeography}, ABC \citep{karaboga2007powerful}, TLBO \citep{rao2011teaching}, ACO \citep{dorigo2006ant}, HS \citep{geem2001new}, Simulated Annealing (SA) \citep{kirkpatrick1983optimization}, DE \citep{storn1997differential}, BA \citep{pham2006bees}, and Invasive Weed Optimisation (IWO) \citep{mehrabian2006novel}. These meta-heuristic algorithms have been utilised extensively to solve optimisation tasks. Table \ref{Tab:simulation_para} reports the simulation parameters for all these algorithms. The population size (P) and maximum iteration for all twelve algorithms are set as 25 and 300, respectively, for all the functions. We conducted 30 independent runs for all the algorithms in MATLAB 2022a on Windows 10 (CPU @ 3.31GHz, 64 GB, 10 Cores) to test their robustness.


\section{Results}
\label{sec:exp_results}

\subsection{Qualitative analysis}
We assessed the behavioral optimisation ability of the proposed SHS algorithm by using unimodal and multi-modal functions. Figure \ref{fig:function_final} illustrates the landscape of different benchmark functions. We plot the search history of the search agents and the convergence curve of the best solution of each benchmark function (Figure \ref{fig:scorpion_position}). The search history usually illustrates the exploration ability of the scorpions in the entire search space, and the convergence curve is used to interpret the general tendency of the population.

The location distribution of all the search agents during the iterative process may be seen in the search history, with the maximum agents located near the global optima. This indicates the proposed SHS algorithm can successfully identify the optimal solution. Figure \ref{fig:scorpion_position} shows the search agents congregate near the global optimum for unimodal functions. In multi-modal tasks, the majority of search agents are dispersed throughout the whole search space near various local optima. The presence of a linear pattern in the search history for multi-modal functions indicates the ability of SHS to avoid local optima. The SHS algorithm should be locally re-exploited for unimodal functions in order to increase the accuracy. The multi-modal functions search space illustrates the trade-off situation between various local optima.

The average best cost function value for all scorpions in a given iteration is represented by the average fitness value. The fitness levels tend to decrease as a result of the ongoing update of scorpions in each iteration. We observe rapid changes during the initial iterations. In unimodal functions, the curve starts to fall quickly right away. Following the sharp decline, variability in the curve's tangent slope approaches stability. The SHS algorithm quickly approaches the desired value and subsequently improves the precision of exploitation. Multi-modal functions have greater curve rates; the curve's amplitude decreases as iteration increases. This suggests global exploration in the early stages and local exploitation in the latter stages of SHS. The curves in multi-model functions decrease gradually or step-by-step, indicating the avoidance of local optima and gradual searching for global optima.

\begin{figure}[ht!]
    \centering
    \includegraphics[width=\textwidth]{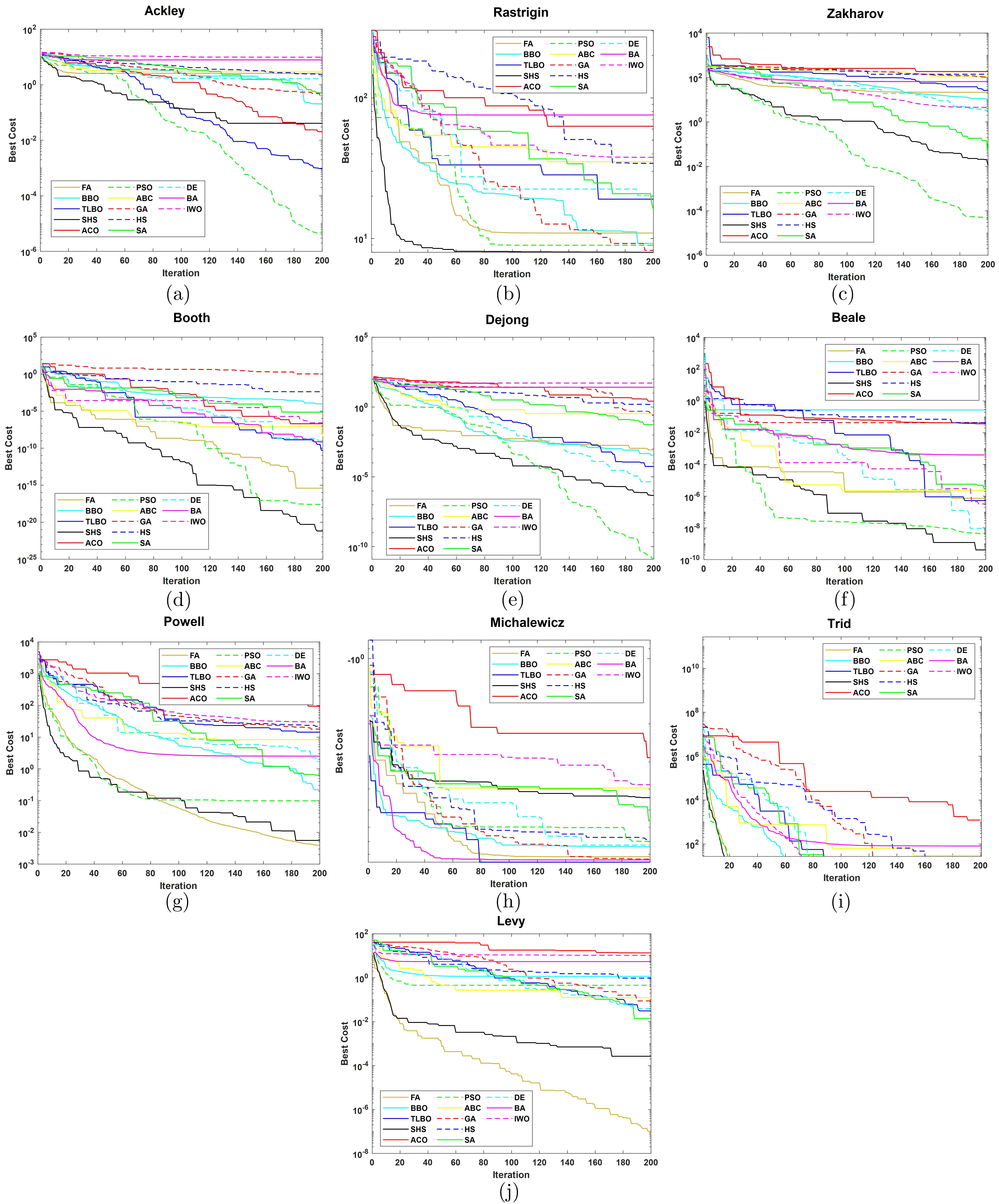}
    \caption{Comparison of SHS algorithm with benchmark algorithms.}
    \label{fig:comparison}
\end{figure}

\subsection{Quantitative analysis}
\begin{table}[ht!]
\caption{Comparison of the best cost value of all the benchmark algorithms after 300 iterations. }
 \resizebox{\textwidth}{!}{
\begin{tabular}{ccccccccccccccc}
\hline
\rowcolor[HTML]{EFEFEF} 
\textbf{Function} & \textbf{Value} & \textbf{GA} & \textbf{PSO} & \textbf{FA} & \textbf{BBO} & \textbf{ABC} & \textbf{SHS} & \textbf{TLBO} & \textbf{ACO} & \textbf{HS} & \textbf{SA} & \textbf{DE} & \textbf{BA} & \textbf{IWO} \\ \hline
 & Best & 0.1014 & 0.0019 & 0.0119 & 0.0108 & 0.0746 & \textbf{0.001} & 0.0292 & 0.2577 & 2.4076 & 0.0405 & 0.042 & 2.8143 & 0.0036 \\ \cline{2-15} 
 & Median & 0.135 & 0.0111 & 0.1479 & 0.2256 & 0.3667 & 0.0109 & 0.5527 & 0.3391 & 3.7901 & 0.2208 & 0.447 & 3.661 & 0.04 \\ \cline{2-15} 
 & Worst & 0.1591 & 0.012 & 0.2541 & 0.3601 & 0.4479 & 0.0112 & 0.714 & 0.4599 & 5.9947 & 0.3245 & 0.6634 & 4.0512 & 0.0818 \\ \cline{2-15} 
\multirow{-4}{*}{\textbf{Ackley}} & Avg & 0.1318 & 0.0083 & 0.138 & 0.1988 & 0.2964 & \textbf{0.0077} & 0.432 & 0.3522 & 4.0641 & 0.1953 & 0.3841 & 3.5088 & 0.0418 \\ \hline
\rowcolor[HTML]{EFEFEF} 
\textbf{} &  &  &  &  &  &  &  &  &  &  &  &  &  &  \\ \hline
 & Best & \textbf{4.9723} & 10.728 & 19.9103 & 17.8559 & 37.7878 & 10.9447 & 15.1785 & 37.2308 & 17.0712 & 18.1546 & 13.2091 & 58.7023 & 68.6557 \\ \cline{2-15} 
 & Median & 5.12 & 16.9713 & 21.2109 & 21.671 & 55.5198 & 12.29 & 17 & 39.0077 & 21.9713 & 19.808 & 14.777 & 63.9007 & 77.1966 \\ \cline{2-15} 
 & Worst & 7.9902 & 20.6566 & 35.0075 & 29.6 & 84.0071 & 15.012 & 21.99 & 47.0909 & 28.8078 & 23.2008 & 16.4251 & 66.7755 & 78.9102 \\ \cline{2-15} 
\multirow{-4}{*}{\textbf{Rastrigin}} & Avg & \textbf{6.0275} & 16.1186 & 25.3762 & 23.0423 & 59.1049 & 12.7489 & 18.0562 & 41.1098 & 22.6168 & 20.3878 & 14.8037 & 63.1262 & 74.9208 \\ \hline
\rowcolor[HTML]{EFEFEF} 
\textbf{} &  &  &  &  &  &  &  &  &  &  &  &  &  &  \\ \hline
 & Best & 94.7443 & 0.0792 & \textbf{0.001} & 0.264 & 54.0885 & 0.0047 & 0.0013 & 164.7525 & 81.5905 & 0.0012 & 0.0026 & 47.7029 & 0.0011 \\ \cline{2-15} 
 & Median & 99.1101 & 1.2008 & 0.0096 & 0.6684 & 63.313 & 0.0096 & 0.0399 & 190.0007 & 88.7 & 0.0062 & 0.0117 & 69.0102 & 0.0187 \\ \cline{2-15} 
 & Worst & 121.5001 & 3.6097 & 0.0202 & 0.9866 & 97.87 & 0.0164 & 0.092 & 213.97 & 96.1108 & 0.0088 & 0.0369 & 71.0019 & 0.0221 \\ \cline{2-15} 
\multirow{-4}{*}{\textbf{Zakharov}} & Avg & 105.1182 & 1.6299 & 0.0103 & 0.6397 & 71.7572 & \textbf{0.0102} & 0.0444 & 189.5744 & 88.8004 & 0.0054 & 0.0171 & 62.5717 & 0.014 \\ \hline
\rowcolor[HTML]{EFEFEF} 
\textbf{} &  &  &  &  &  &  &  &  &  &  &  &  &  &  \\ \hline
 & Best & 0.1424 & \textbf{1.97E-29} & 2.83E-15 & 8.02E-06 & 3.19E-15 & 8.80E-24 & 5.70E-10 & 1.82E-19 & 0.0058 & 9.02E-08 & 1.49E-07 & 2.37E-16 & 6.97E-08 \\ \cline{2-15} 
 & Median & 0.2755 & 2.75E-29 & 3.96E-15 & 1.04E-05 & 4.14E-15 & 1.23E-23 & 7.61E-10 & 2.86E-19 & 0.0228 & 1.26E-07 & 2.08E-07 & 3.31E-16 & 9.75E-08 \\ \cline{2-15} 
 & Worst & 0.6501 & 3.15E-29 & 4.52E-15 & 1.36E-05 & 4.78E-15 & 1.40E-23 & 9.12E-10 & 2.91E-19 & 0.0745 & 1.71E-07 & 2.38E-07 & 3.79E-16 & 1.11E-07 \\ \cline{2-15} 
\multirow{-4}{*}{\textbf{Booth}} & Avg & 0.356 & \textbf{2.62E-29} & 3.77E-15 & 1.06E-05 & 4.03E-15 & 1.17E-23 & 7.47E-10 & 2.53E-19 & 0.0344 & 1.29E-07 & 1.98E-07 & 3.15E-16 & 9.27E-08 \\ \hline
\rowcolor[HTML]{EFEFEF} 
\textbf{} &  &  &  &  &  &  &  &  &  &  &  &  &  &  \\ \hline
 & Best & 0.0035 & 0.000002 & 0.0002 & 0.053 & 0.0003 & \textbf{0.000001} & 0.0014 & 0.0145 & 1.0656 & 0.0005 & 0.0009 & 6.692 & 26.2144 \\ \cline{2-15} 
 & Median & 0.0197 & 0.000033 & 0.0019 & 0.0966 & 0.0009 & 0.000014 & 0.0614 & 0.0376 & 2.997 & 0.0037 & 0.0011 & 8.1907 & 38.4491 \\ \cline{2-15} 
 & Worst & 0.0227 & 0.000096 & 0.0027 & 0.123 & 0.0011 & 0.000021 & 0.0958 & 0.0882 & 3.9098 & 0.0044 & 0.0073 & 10.4409 & 69.0018 \\ \cline{2-15} 
\multirow{-4}{*}{\textbf{DeJong}} & Avg & 0.0153 & 0.000044 & 0.0016 & 0.0909 & 0.0008 & \textbf{0.000012} & 0.0529 & 0.0468 & 2.6575 & 0.0029 & 0.0031 & 8.4412 & 44.5551 \\ \hline
\rowcolor[HTML]{EFEFEF} 
\textbf{} &  &  &  &  &  &  &  &  &  &  &  &  &  &  \\ \hline
 & Best & 0.0014 & 1.67E-08 & 9.16E-08 & 0.000003 & 6.45E-11 & \textbf{3.53E-13} & 1.59E-07 & 1.33E-10 & 0.0027 & 2.11E-08 & 4.34E-07 & 0.000005 & 1.95E-09 \\ \cline{2-15} 
 & Median & 0.0444 & 2.33E-08 & 1.28E-07 & 0.000019 & 9.03E-11 & 4.94E-13 & 2.22E-07 & 1.86E-10 & 0.0391 & 2.74E-08 & 5.64E-07 & 0.000022 & 2.53E-09 \\ \cline{2-15} 
 & Worst & 0.0667 & 2.67E-08 & 1.46E-07 & 0.000181 & 1.03E-10 & 5.64E-13 & 2.54E-07 & 2.12E-10 & 0.0449 & 3.79E-08 & 7.81E-07 & 0.000198 & 3.51E-09 \\ \cline{2-15} 
\multirow{-4}{*}{\textbf{Beale}} & Avg & 0.0375 & 2.22E-08 & 1.21E-07 & 0.000068 & 8.59E-11 & \textbf{4.70E-13} & 2.11E-07 & 1.77E-10 & 0.0289 & 2.88E-08 & 5.93E-07 & 0.000075 & 2.66E-09 \\ \hline
\rowcolor[HTML]{EFEFEF} 
\textbf{} &  &  &  &  &  &  &  &  &  &  &  &  &  &  \\ \hline
 & Best & 15.5555 & 0.0771 & 0.0025 & 1.2768 & 0.229 & \textbf{0.002} & 0.0053 & 4.1154 & 49.7533 & 0.0049 & 0.008 & 0.5378 & 0.0066 \\ \cline{2-15} 
 & Median & 17.666 & 0.1177 & 0.0228 & 2.9917 & 0.6607 & 0.0045 & 0.0061 & 5.9 & 72.8109 & 0.0111 & 0.017 & 0.991 & 0.0488 \\ \cline{2-15} 
 & Worst & 20.367 & 0.1494 & 0.0338 & 3.88 & 0.8794 & 0.0196 & 0.0099 & 9.6607 & 91.5515 & 0.0117 & 0.0214 & 1.3902 & 0.1199 \\ \cline{2-15} 
\multirow{-4}{*}{\textbf{Powell}} & Avg & 17.8628 & 0.1147 & 0.0197 & 2.7162 & 0.5897 & 0.0087 & \textbf{0.0071} & 6.5587 & 71.3719 & 0.0092 & 0.0155 & 0.973 & 0.0584 \\ \hline
\rowcolor[HTML]{EFEFEF} 
\textbf{} &  &  &  &  &  &  &  &  &  &  &  &  &  &  \\ \hline
 & Best & -8.9968 & -5.4824 & -7.4916 & -7.138 & -4.7891 & -8.683 & -7.3695 & -2.6708 & -8.8002 & -6.6745 & -7.3803 & \textbf{-9.0557} & -4.0588 \\ \cline{2-15} 
 & Median & -6.9713 & -3.667 & -4.991 & -5 & -3.9019 & -5.0097 & -6.1408 & -1.1222 & -7.7009 & -4.9131 & -6.0009 & -8.0066 & -2.0014 \\ \cline{2-15} 
 & Worst & -4.3207 & -2.946 & -3.1299 & -4.909 & -1.148 & -6.902 & -5.3141 & 0.1892 & -6.2424 & -2.4528 & -4.4449 & -5.3212 & -1.8009 \\ \cline{2-15} 
\multirow{-4}{*}{\textbf{Michalewicz}} & Avg & -6.7629 & -4.0318 & -5.2042 & -5.6823 & -3.2797 & -6.8649 & -6.2748 & -1.2013 & \textbf{-7.5812} & -4.6801 & -5.942 & -7.4612 & -2.6204 \\ \hline
\rowcolor[HTML]{EFEFEF} 
\textbf{} &  &  &  &  &  &  &  &  &  &  &  &  &  &  \\ \hline
 & Best & -24.1436 & -28.95 & -31.9753 & -28.8239 & 49.0756 & \textbf{-32.0596} & -31.8571 & -30.7573 & -37.0128 & -31.8311 & -31.9412 & -32.0592 & -30.4138 \\ \cline{2-15} 
 & Median & -20.222 & -21.6574 & -29.11 & -24.6633 & 78.0409 & -31.1993 & -30.1474 & -18.9607 & -34.0006 & -29.9912 & -27.0008 & -26.0017 & -26.9998 \\ \cline{2-15} 
 & Worst & -18.9971 & -19.12 & -15.771 & -21.0007 & 88.9412 & -27.0809 & -26.4404 & -15.0097 & -28.0045 & -18.9991 & -19.94 & -21.6604 & -22.8217 \\ \cline{2-15} 
\multirow{-4}{*}{\textbf{Trid}} & Avg & -21.1209 & -23.2425 & -25.6188 & -24.8293 & 72.0192 & \textbf{-30.1133} & -29.4816 & -21.5759 & -33.006 & -26.9405 & -26.294 & -26.5738 & -26.7451 \\ \hline
\rowcolor[HTML]{EFEFEF} 
\textbf{} &  &  &  &  &  &  &  &  &  &  &  &  &  &  \\ \hline
 & Best & 0.0035 & 0.0595 & 0.00012 & 0.0273 & 0.1725 & \textbf{0.00011} & 0.00029 & 0.1132 & 0.0711 & 0.00026 & 0.0007 & 0.00023 & 13.7537 \\ \cline{2-15} 
 & Median & 0.011 & 0.199 & 0.00228 & 0.4807 & 0.6399 & 0.00081 & 0.0022 & 0.3961 & 0.094 & 0.00136 & 0.0029 & 0.00312 & 19.2408 \\ \cline{2-15} 
 & Worst & 0.0147 & 0.2278 & 0.00642 & 0.639 & 0.8209 & 0.00096 & 0.00409 & 0.5543 & 0.1553 & 0.00449 & 0.0039 & 0.0049 & 55.91 \\ \cline{2-15} 
\multirow{-4}{*}{\textbf{Levy}} & Avg & 0.0097 & 0.1621 & 0.00294 & 0.3823 & 0.5444 & \textbf{0.00063} & 0.00219 & 0.3545 & 0.1068 & 0.00204 & 0.0025 & 0.00275 & 29.6348 \\ \hline
\end{tabular}}
\label{tab:stat_all}
\end{table}

Table \ref{tab:stat_all} reports the statistical result (best, median, worst, and average) of SHS and other benchmark algorithms. We conducted 30 independent runs (300 iterations) of the benchmark algorithms for all the functions, thus evolving 9000 (300 $\times$ 30) times on the benchmark functions. The leading values in the case of best and average are made bold. Out of ten benchmark functions, SHS achieves first rank (1/13) either in best or average in seven functions (Ackley, Zakharov, DeJong, Beale, Powell, Trid, and Levy). In Rastrigin, SHS obtained second rank (2/13) in average after GA and third rank (3/13) in best after GA and PSO. In Booth, SHS ranks second (2/13) in both best and average after PSO. In Michalewicz, SHS ranks fourth (4/13) in best after BA, GA, \& HS and ranks second (2/13) in average. We observed similar results for 100 and 1000 iterations (Tables \ref{tab:100iterations} \& \ref{tab:1000iterations}). We selected 300 iterations as the focal point because this value strikes a well-balanced tradeoff between lower and higher iteration counts. Through a meticulous examination of the data, we found that 300 iterations
offer optimal convergence without straying into excessive computation times or unnecessary fine-tuning.

The convergence analysis is very useful in explaining the algorithm's local exploitation and global exploration. Figure \ref{fig:comparison} shows the convergence curve of all the thirteen optimisation algorithms, including SHS, for all the benchmark functions. Overall, SHO shows a better optimisation performance with the best (or second best) results in eight out of ten (8/10) benchmark functions (i.e., Ackley, Rastrigin, Booth, DeJong, Beale, Powell, Trid, and Levy). The SHS demonstrated its local exploitation capabilities for unimodal functions, resulting in fast convergence and high accuracy. This correlates well with the adaptive parameters of SHS, such as $\alpha$ and $\beta$, that become stronger as the scorpion moves towards the prey, resulting in SHS convergence faster towards the optimum than other algorithms. For multi-modal, SHS demonstrated its global optimisation potential. For the function, Ackley, SHS, and PSO show outstanding optimisation accuracy. The performance of SHS over the Rastrigin function clearly illustrates the most obvious case of exploration in which the SHS algorithms first jumped out of local optima and then converged accurately near global optima. Although BE, HS, and GA perform better than SHS for the Michalewicz function, the step-by-step convergence of SHS clearly demonstrates its superb exploration capability. Similar to the case of Ackley, Levy also shows outstanding optimisation accuracy. Overall, the convergence analysis confirms the effectiveness of the SHS algorithm.

\begin{figure}[t!]
    \centering
    \includegraphics[width=\textwidth]{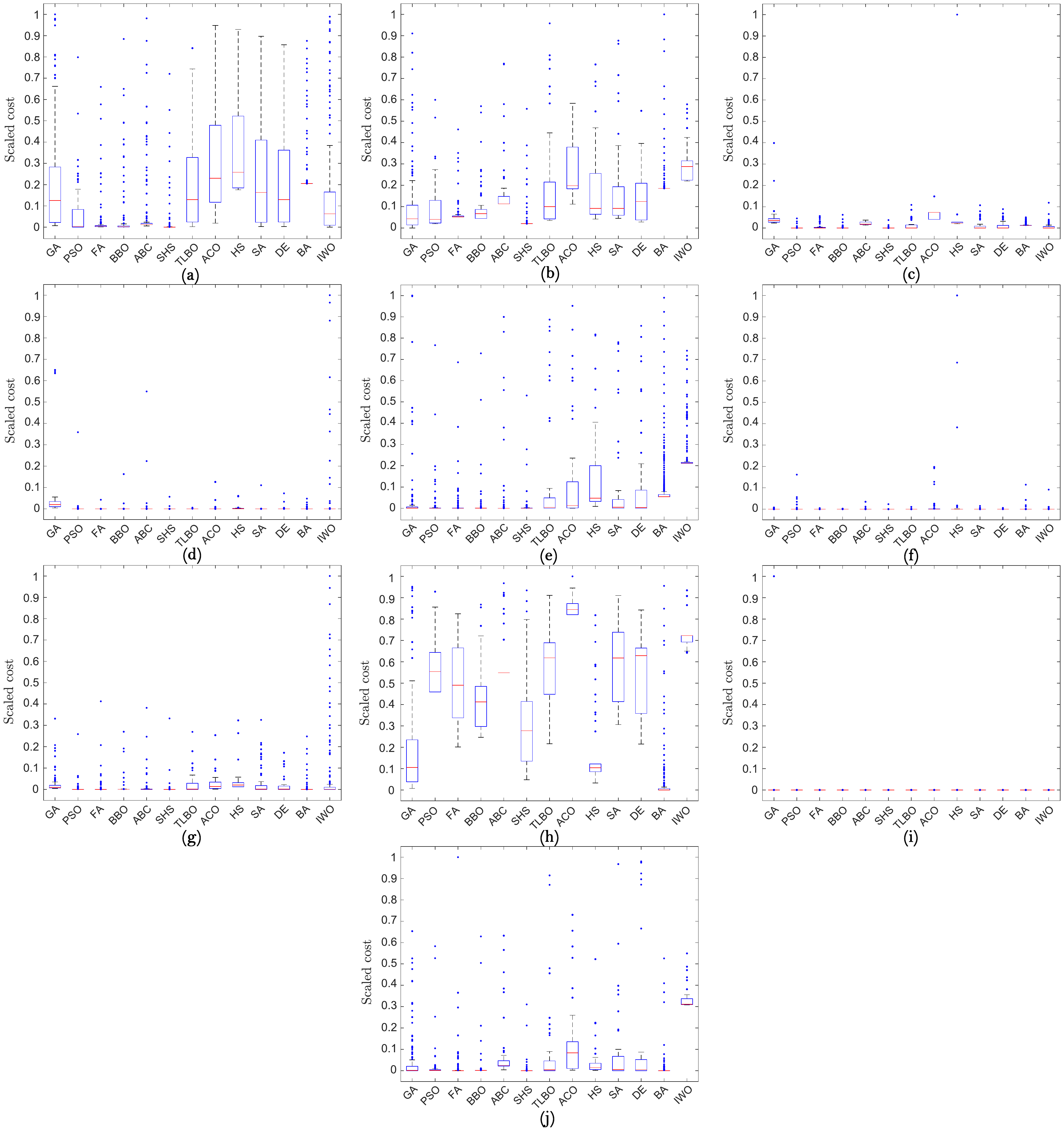}
    \caption{Dispersion analysis of the best cost for all the benchmark algorithms.}
    \label{fig:dispersion}
\end{figure}

In addition to the convergence analysis, we perform dispersion analysis \citep{stefanello2015biased}. We first scale the best cost values of all the algorithms because each instance might have very different values. The scaling process involves a simple transformation with the most substantial cost among all the examined algorithms normalised to 1, while the smallest cost is adjusted to 0. We plotted the spread of the normalised costs for each distinct algorithm (Figure \ref{fig:dispersion}). The diagram employs box plots to indicate key statistical points such as the minimum value, lower quartile, median, upper quartile, and maximum value associated with each algorithm. Notably, distant data points are depicted as blue dots, signifying outliers. Our analysis revealed that the SHS method consistently outperforms other algorithms across nine functions: Ackley, Rastrigin, Zakharov, Booth, DeJong, Beale, Powell, Trid, and Levy. However, in Michalewicz function, SHS ranks fourth, following BA, HS, and GA.

\subsection{Statistical analysis}
\begin{figure}[ht!]
    \centering
    \includegraphics[width=\textwidth]{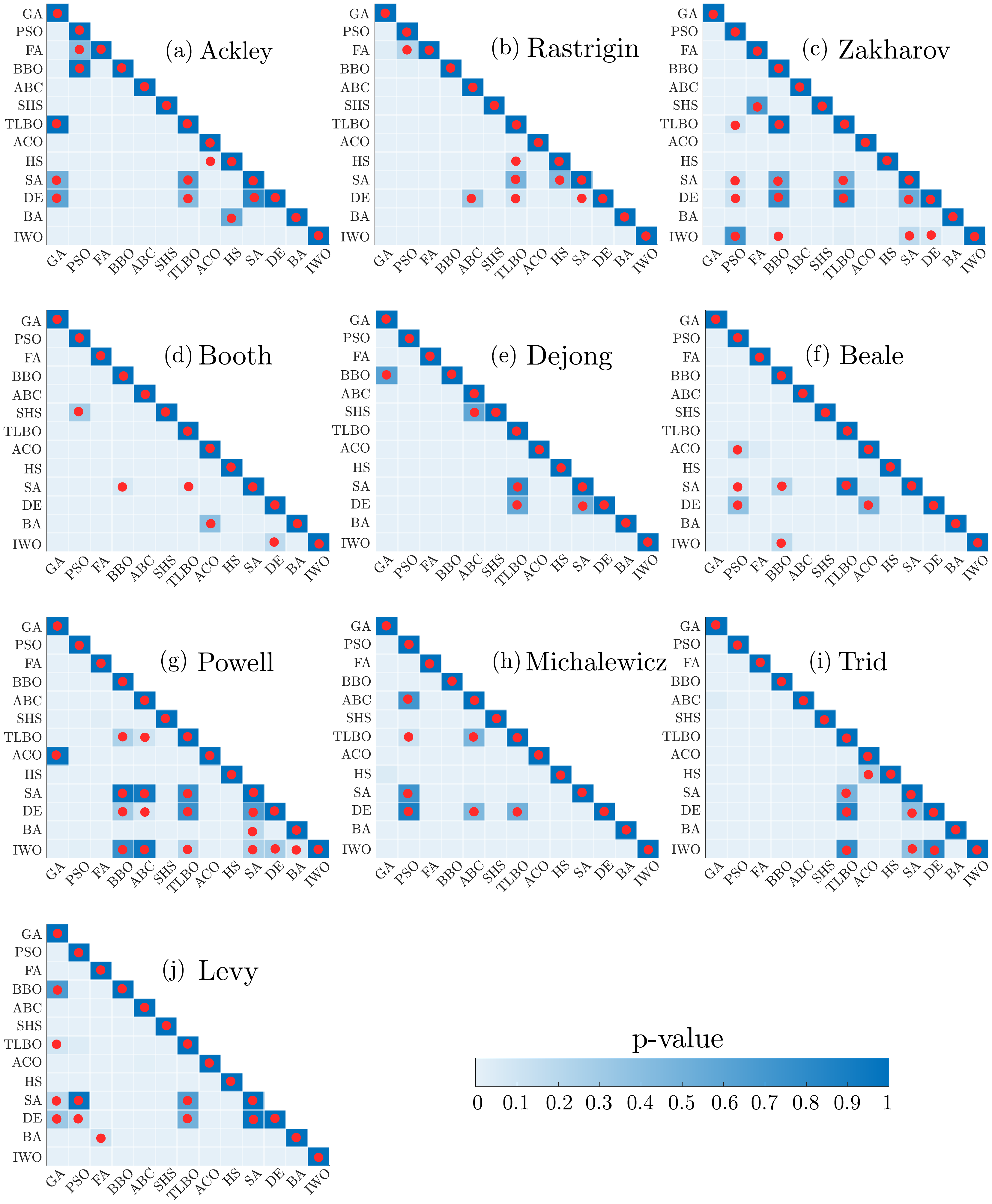}
    \caption{Wilcoxon test of all the meta-heuristic algorithms for all the benchmark functions. The heatmap (a)-(j) represents the p-values for all the algorithms with each other. The cells with p $\ge$ 0.5 between the two algorithms have been marked with a red-filled circle.}
    \label{fig:Wilcoxon}
\end{figure}

\begin{figure}
    \centering
    \includegraphics[width=\textwidth]{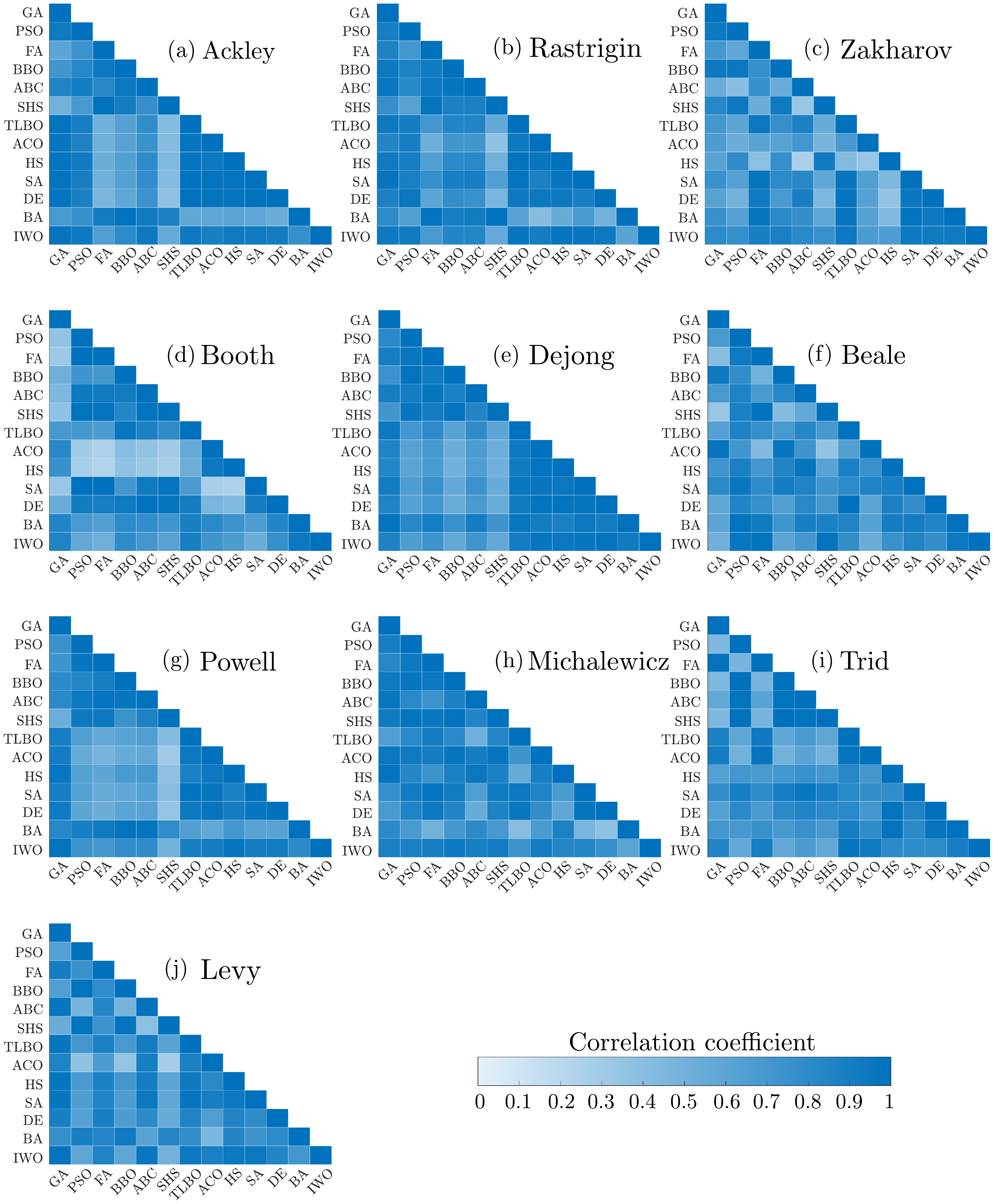}
    \caption{The heatmap (a)-(j) represents the correlation coefficient between the best costs of all the algorithms.}
    \label{fig:correlation_matrix}
\end{figure}

We perform statistical analysis to interpret better and avoid any accidental results. We apply the Wilcoxon rank sum test on all the benchmark functions at 5\% significance level \citep{wilcoxon1992individual,zimmerman1993relative}. It is a nonparametric test that checks the null hypothesis, whether the compared data belongs to the same distribution with an equal median. It is worth mentioning that the Wilcoxon rank sum test is equivalent to the Mann-Whitney U test. We computed the pairwise p-values between SHS and all other twelve algorithms. The two algorithms indicate a significant difference if the p-value is less than 0.05 (Figure \ref{fig:Wilcoxon}). We highlighted the cell with a p-value greater than 0.05 with a filled red circle. We observed the performance of SHS is statistically different from the twelve benchmark algorithms for Ackley (Figure \ref{fig:Wilcoxon}a), Rastrigin (Figure \ref{fig:Wilcoxon}b), Zakharov with an exception from FA (Figure \ref{fig:Wilcoxon}c), Booth with an exception from PSO (Figure \ref{fig:Wilcoxon}d), DeJong with an exception from ABC (Figure \ref{fig:Wilcoxon}e), Beale (Figure \ref{fig:Wilcoxon}f), Powell (Figure \ref{fig:Wilcoxon}g), Michalewicz (Figure \ref{fig:Wilcoxon}h), Trid (Figure \ref{fig:Wilcoxon}i), and Levy (Figure \ref{fig:Wilcoxon}j). The presence of a single non-significant case in only three benchmark functions, that too with different algorithms, confirms that the performance of the SHS is better in the statistical sense than the twelve other benchmark algorithms. On the other hand, we found that the performance of TLBO is non-significant when compared with SA and DE in most of the benchmark functions. Along with the p-value, we also reported the correlation coefficient (R) in Figure \ref{fig:correlation_matrix}. Upon analysing the data, we have observed that the SHS algorithm demonstrates a relatively low R-value when compared to other algorithms across most benchmark optimisation functions, with the exception of the Michalewicz function. This observation suggests that SHS follows a distinctive convergence behavior.

For the Ackley function, SHS displays a high R-value (i.e., R $>$ 0.90) solely with the FA and the BBO algorithm (2 out of 12 cases). Similarly, when considering the Rastrigin function, SHS shows a high R-value only in relation to FA and BA (2 out of 12 cases). In the context of the Zakharov function, SHS exhibits a high R-value exclusively with PSO, BBO, and HS (3 out of 12 cases). Moving on to the Booth function, SHS demonstrates a high R-value with PSO, FA, ABC, SA, and DE (5 out of 12 cases). For the DeJong function, SHS displays a high R-value solely with PSO, FA, and BBO (3 out of 12 cases). In the case of the Beale function, SHS has a high R-value only with PSO and IWO (2 out of 12 cases). When analysing the Powell function, SHS exhibits a high R-value with only PSO and FA (2 out of 12 cases). 
Interestingly, for the Michalewicz function, nearly all the algorithms exhibit a high R-value with each other (7 out of 12 cases). Concerning the Trid function, SHS shows a high R-value with PSO, BBO, and ABC (3 out of 12 cases). Finally, when considering the Levy function, SHS displays a high R-value exclusively with PSO and BBO (2 out of 12 cases).

Apart from the Wilcoxon rank sum test, we also performed the Friedman rank test \citep{derrac2011practical}. This is also a non-parametric test that ranks according to the significant differences. We applied it to compare the average performance of the twelve algorithms for all the ten benchmark functions. The results of the Friedman rank test are provided in Table \ref{tab:friedman}. We found that the SHS algorithm ranked first in nine out of the ten benchmark functions by outperforming twelve algorithms. SHS ranked third in the case of the DeJong function, where PSO ranked first and ABC ranked second. Thus, the overall results of the Friedman rank test confirm the stability and efficacy of the proposed SHS algorithm.

\begin{table}[ht!]
\caption{Friedman rank-based test for algorithms on different benchmark functions. The values represent the average rank for each algorithm for up to 300 iterations. The best values are marked in red for each benchmark function.}
 \resizebox{\textwidth}{!}{
\begin{tabular}{cccccccccccccc}
\hline
{\color[HTML]{000000} \begin{tabular}[c]{@{}c@{}}Functions/\\ Algorithms\end{tabular}} & GA & PSO & FA & BBO & ABC & SHS & TLBO & ACO & HS & SA & DE & BA & IWO \\ \hline
Ackley & 8.5 & 3.0 & 3.4 & 2.7 & 5.7 & {\color[HTML]{FF0000} \textbf{1.2}} & 8.5 & 11.7 & 11.7 & 9.2 & 9.2 & 10.0 & 6.2 \\ \hline
Rastrigin & 3.2 & 3.9 & 4.3 & 4.5 & 8.1 & {\color[HTML]{FF0000} \textbf{1.4}} & 7.9 & 11.7 & 8.9 & 8.2 & 7.1 & 9.8 & 12.1 \\ \hline
Zakharov & 12.0 & 4.4 & 2.5 & 5.5 & 9.7 & {\color[HTML]{FF0000} \textbf{2.1}} & 6.8 & 13.0 & 10.5 & 5.7 & 6.4 & 8.3 & 4.2 \\ \hline
Booth & 12.9 & 3.0 & 3.8 & 7.4 & 3.3 & {\color[HTML]{FF0000} \textbf{1.6}} & 7.9 & 6.4 & 11.9 & 8.3 & 9.1 & 5.6 & 9.7 \\ \hline
DeJong & 6.2 & {\color[HTML]{FF0000} \textbf{1.9}} & 3.4 & 5.9 & 2.6 & 2.8 & 7.6 & 10.3 & 11.3 & 7.5 & 8.2 & 10.7 & 12.6 \\ \hline
Beale & 11.0 & 6.6 & 3.7 & 7.7 & 1.7 & {\color[HTML]{FF0000} \textbf{1.6}} & 7.2 & 7.0 & 12.7 & 7.1 & 5.4 & 10.5 & 8.9 \\ \hline
Powell & 10.7 & 5.0 & 1.9 & 7.2 & 6.6 & {\color[HTML]{FF0000} \textbf{1.4}} & 7.6 & 11.5 & 12.2 & 7.6 & 7.4 & 5.5 & 6.4 \\ \hline
Michalewicz & 3.0 & 8.9 & 6.7 & 5.1 & 8.9 & 4.0 & 8.0 & 12.9 & 2.6 & 9.5 & 8.3 & {\color[HTML]{FF0000} \textbf{1.1}} & 12.0 \\ \hline
Trid & 12.6 & 5.4 & 3.8 & 4.7 & 11.5 & {\color[HTML]{FF0000} \textbf{1.7}} & 7.8 & 9.9 & 9.0 & 7.1 & 7.7 & 2.0 & 7.7 \\ \hline
Levy & 6.1 & 6.5 & 3.2 & 5.2 & 11.0 & {\color[HTML]{FF0000} \textbf{1.3}} & 7.6 & 11.8 & 8.9 & 7.4 & 7.3 & 2.3 & 12.6 \\ \hline
\end{tabular}}
\label{tab:friedman}
\end{table}

\subsection{Performance of SHS on CEC2020 functions}
We also evaluate the performance of SHS over modern benchmark functions to check its local extremum avoidance capability. For this, we considered the latest CEC2020 (IEEE 2020 Congress on Evolutionary Computation) benchmark functions \citep{biswas2020large,ezugwu2022prairie,agushaka2023gazelle,kadkhoda2023novel,mohamed2020evaluating,kumar2023chaotic,yuan2023coronavirus,kumar2020test}. The CEC2020 consists of ten benchmark functions, including unimodal, basic, hybrid, and composite functions (Table \ref{tab:cec2020}). They are developed by leveraging translation, rotation, and a combination of conventional benchmark functions. These functions are more challenging to optimise since they are quite dynamic and complicated with many local optimum values. Figure \ref{fig:cec2020funcplot} shows the landscape and details of all the ten CEC2020 benchmark functions. We plot the search agents' search history and the best solution's convergence curve in the left and right panels for each CEC2020 benchmark function (Figure \ref{fig:cec2020searchplot}). We observed that the search agents are aggregated near the global optimum (Figure \ref{fig:cec2020searchplot}a). The search agents are widely distributed throughout the search space while reaching the global optimum (Figures \ref{fig:cec2020searchplot}b, c, \&d). We observed similar behaviour for all the hybrid functions (Figures \ref{fig:cec2020searchplot}e, f, \&g). The presence of a linear search pattern in the composite functions demonstrates the local optimum avoidance ability of the SHS (Figures \ref{fig:cec2020searchplot}h, i, \&j). We found that out of ten CEC2020 benchmark functions, SHS achieves first rank (1/13) either in best or average in six functions (F1, F2, F3, F4, F8, and F10). In the case of F5, SHS obtained fifth rank (5/13) after HS, BBO, BA, and FA. In the case of F6, SHS obtained second rank (2/13) after BA. The SHS ranks third (3/13) in the case of F7 after HS and BA. Lastly, in the case of F9, SHS ranks seventh (7/13) after FA, BA, TLBO, DE, HS, and ACO. Figure \ref{fig:cec2020convergence} demonstrate the convergence curves of all the algorithm over CEC2020 benchmark functions. We observed in all the functions, SHS reaches the global optimum.

\section{Discussion}
\label{sec:discussion}
Most real-world problems are constraint optimisation, which requires a constraint handling approach enabled meta-heuristic \citep{mirjalili2014grey}. Constraint handling techniques such as penalty function, $\epsilon-$constrained method, stochastic ranking, gradient-based repair, decoders, feasibility rules, special operators, and objective function separation techniques can perform this task \citep{liu2019handling,cantu2021constraint,coello2022constraint}. Out of these techniques, the penalty function technique is the most frequently used for handling constraints because of its ease of use and simplicity \citep{deb2000efficient}. It converts the constrained optimisation problem to an unconstrained one by penalising the objective function for violating the constraints. 

\subsection{Clustering using SHS}
\begin{figure}[!]
    \centering
    \includegraphics[width=\textwidth]{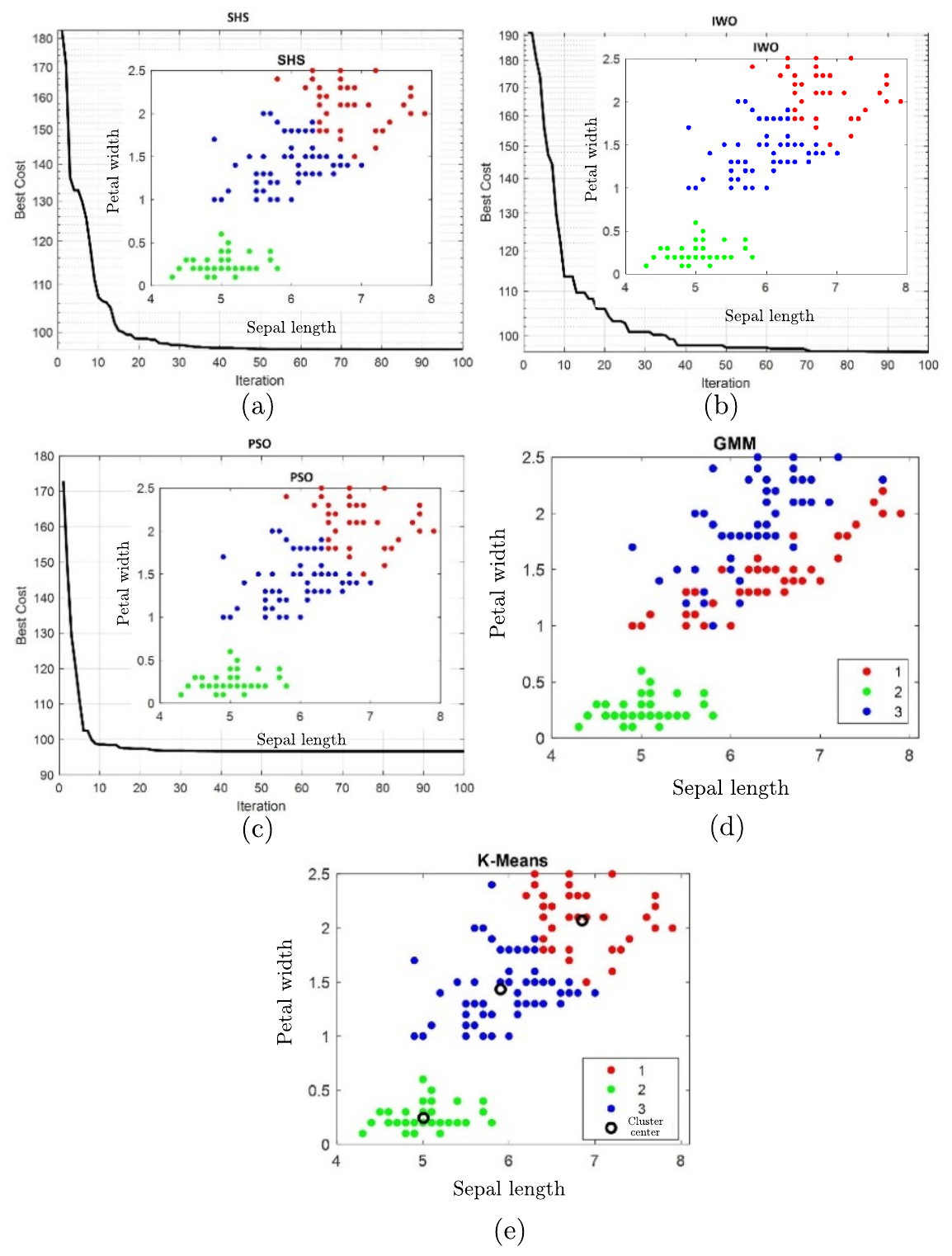}
    \caption{Clustering on iris data by (a) SHS algorithm, (b) IWO algorithm, (c) PSO algorithm, (d) Gaussian Mixture Model (GMM) algorithm, and (e) K-means algorithm.}
    \label{fig:shs_clustering}
\end{figure}

We can easily turn a clustering problem into an optimisation problem that can be solved by using a nature-inspired algorithm. A single-objective clustering problem ($\psi$,$f$) can be done by determining the optimal cluster $C^{\#}$:
\begin{equation}
   f(C^{\#}) = \text{min}\{f(C)\mid C  \in  \psi \},
\end{equation}
where $\psi = \{C^1, C^2, ..., C^n\}$ represent the set of all feasible clusters and $C = \{c_1, c_2, ..., c_k\}$, where $k$ is the number of clusters for the given dataset (\textbf{X}). $f(.)$ is the between-group scatter criterion function given by

\begin{equation}
f = \sum_{c_k, c_r \in C} d_e( \overline{c_k}, \overline{c_r}),
\end{equation}
where $d_e$ is the Euclidean distance between the centroid of the two clusters $c_k$ and $c_r$. We have considered the widely used Iris dataset \citep{fisher1936use} for evaluating the clustering capability of the SHS algorithm. We used petal width and sepal length as the two features for clustering the data into three different classes: setosa (class 1), versicolor (class 2), and virginica (class 3). We found that the SHS algorithm efficiently partitioned the entire data into three clusters (Figure \ref{fig:shs_clustering}). Further, we compared the SHS results with two popular evolutionary (i.e., IWO and PSO) and two conventional clustering algorithms (i.e., GMM and K-means). We observed a comparable performance by all these algorithms.

\subsection{Regression using SHS}
To solve the regression-based machine learning problems using the SHS algorithm, we coupled the SHS algorithm with the fuzzy inference system (FIS) of type Sugeno using FCM clustering (fuzzy cluster = 2). In this hybrid model, all the input variables have a Gaussian membership function for each fuzzy cluster, and the output variable has a linear membership function. We have one rule for each fuzzy cluster.

We evaluate the performance of the SHS regression algorithm using publicly available datasets, namely AutoML-ID \citep{singh2022automl}, LT-FS-ID \citep{singh2022lt}, FF-ANN-ID \citep{singh2022deep}, and ALE \citep{singh2020machine}. We first trained the SHS regression algorithm and tested its performance on unseen datasets using R, RMSE, and bias as the performance metrics (Table \ref{tab:SHSregression}). We found that the SHS regression algorithm performs well on all four datasets. In addition, we compared its performance with the primary algorithms on the same datasets. We found that in the case of LT-FS-ID and FF-ANN-ID datasets, the SHS algorithm outperforms the results of the primary algorithm.
\begin{table}[ht!]
\caption{Comparison of the performance of the SHS regression algorithm over four publicly available datasets.}
\label{tab:SHSregression}
\resizebox{\textwidth}{!}{ 
\begin{tabular}{|c|cccccccccc|}
\hline
\multirow{3}{*}{\textbf{\begin{tabular}[c]{@{}c@{}}Performance\\ metrics\end{tabular}}} & \multicolumn{10}{c|}{\textbf{Datasets}} \\ \cline{2-11} 
 & \multicolumn{2}{c|}{\textbf{AutoML-ID datasets}} & \multicolumn{2}{c|}{\textbf{LT-FS-ID datasets}} & \multicolumn{2}{c|}{\textbf{\begin{tabular}[c]{@{}c@{}}FF-ANN-ID datasets \\ (Gaussian)\end{tabular}}} & \multicolumn{2}{c|}{\textbf{\begin{tabular}[c]{@{}c@{}}FF-ANN-ID datasets \\ (Uniform)\end{tabular}}} & \multicolumn{2}{c|}{\textbf{\begin{tabular}[c]{@{}c@{}}ALE\\ datasets\end{tabular}}} \\ \cline{2-11} 
 & \textbf{SHS} & \textbf{Primary} & \textbf{SHS} & \textbf{Primary} & \textbf{SHS} & \textbf{Primary} & \textbf{SHS} & \textbf{Primary} & \textbf{SHS} & \textbf{Primary} \\ \hline
\textbf{R} & 0.99 & 1 & 1 & 0.98 & 0.99 & 0.78 & 0.82 & 0.79 & 0.71 & 0.82 \\ \hline
\textbf{RMSE} & 3.36 & 0.01 & 0.01 & 6.47 & 5.97 & 41.45 & 27.40 & 48.36 & 0.14 & 0.15 \\ \hline
\textbf{Bias} & 0.99 & -0.01 & 0 & 12.35 & 1.07 & 3.02 & -5.92 & 9.55 & 0.06 & 0.05 \\ \hline
\end{tabular}}
\end{table}

\subsection{SHS for minimum spanning tree (MST) task}
\begin{figure}[ht!]
    \centering
    \includegraphics[width=\textwidth]{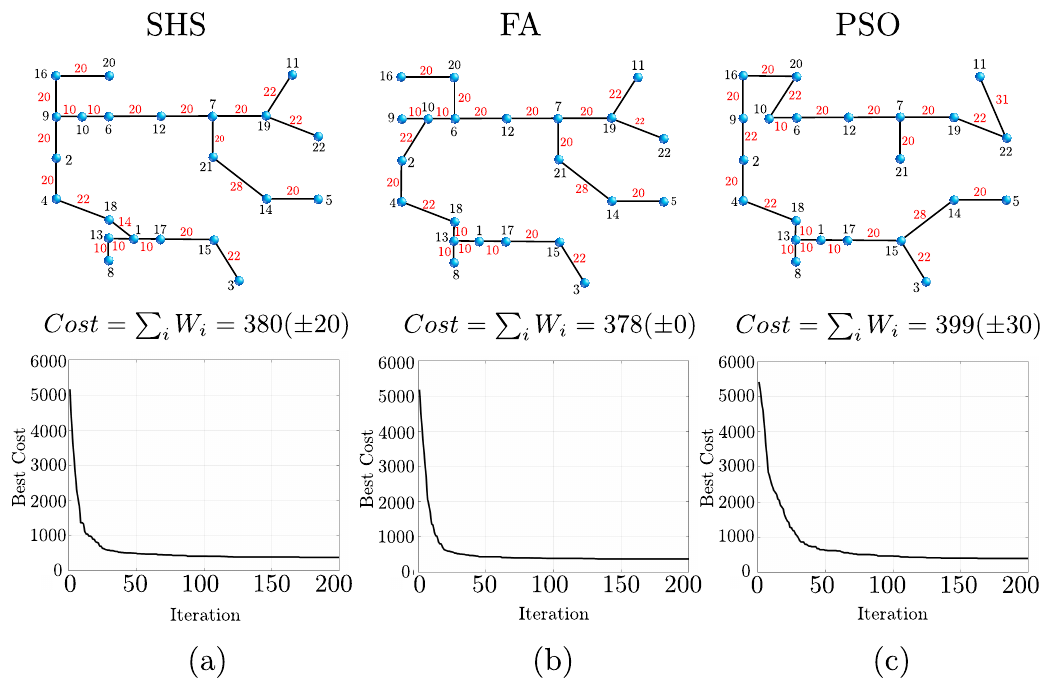}
    \caption{Minimum spanning tree task using (a) SHS, (b) FA, and PSO. The filled blue colored circles in the upper panel represent nodes. The number indicated in red color between two nodes represents the weight of the corresponding branch.}
    \label{fig:MST}
\end{figure}

Minimum Spanning Tree (MST) is a well-known problem in graph theory that involves finding a connected path between multiple nodes with the lowest edges’ weight without any cycle among nodes \citep{graham1985history}. Finding a minimum spanning tree could be achieved using conventional algorithms such as Prim’s or bio-inspired algorithms. Here, we used the SHS algorithm to solve the problem of MST. We considered an arbitrary network graph comprising 22 nodes (Figure \ref{fig:MST}). The coordinates of the nodes are given in Table \ref{tab:MST}. The line connecting two nodes is called a branch, and each branch has been assigned a weight [$W(i,j)$] according to Equation \ref{eq:MST_weight}:
\begin{equation}
    W(i,j) = [\sqrt{(X(i)-X(j))^2+(Y(i)-Y(j))^2}].
    \label{eq:MST_weight}
\end{equation}
We iterated SHS, Firefly Algorithm (FA), and PSO algorithms to find the path with minimum weights. We found that the SHS and FA solve the problem of MST with lower cost values than PSO. The SHS solves it with a cost of 380$\pm$20, and FA with a cost of 278$\pm$0. The PSO shows a poor performance with a cost value of 399$\pm$30. We also test the performance of Harmony Search (HS) and Differential Evolution (DE) algorithms to solve the problem of MST. We found that these algorithms fail to converge, resulting in a mesh-shaped graph. 
\begin{table}[ht!]
\caption{Co-ordinates of all the twenty-two nodes.}
\label{tab:MST}
\resizebox{\textwidth}{!}{ 
\begin{tabular}{ccccccccccccccccccccccc}
\hline
\textbf{Nodes} & \textbf{1} & \textbf{2} & \textbf{3} & \textbf{4} & \textbf{5} & \textbf{6} & \textbf{7} & \textbf{8} & \textbf{9} & \textbf{10} & \textbf{11} & \textbf{12} & \textbf{13} & \textbf{14} & \textbf{15} & \textbf{16} & \textbf{17} & \textbf{18} & \textbf{19} & \textbf{20} & \textbf{21} & \textbf{22} \\ \hline
\textbf{X} & 30 & 0 & 70 & 0 & 100 & 20 & 60 & 20 & 0 & 10 & 90 & 40 & 20 & 80 & 60 & 0 & 40 & 20 & 80 & 20 & 60 & 100 \\ \hline
\textbf{Y} & 20 & 60 & 0 & 40 & 40 & 80 & 80 & 10 & 80 & 80 & 100 & 80 & 20 & 40 & 20 & 100 & 20 & 30 & 80 & 100 & 60 & 70 \\ \hline
\end{tabular}}
\end{table}

\subsection{SHS for hub location-allocation problem (HLP) task}
\begin{figure}[ht!]
    \centering
    \includegraphics[width=\textwidth]{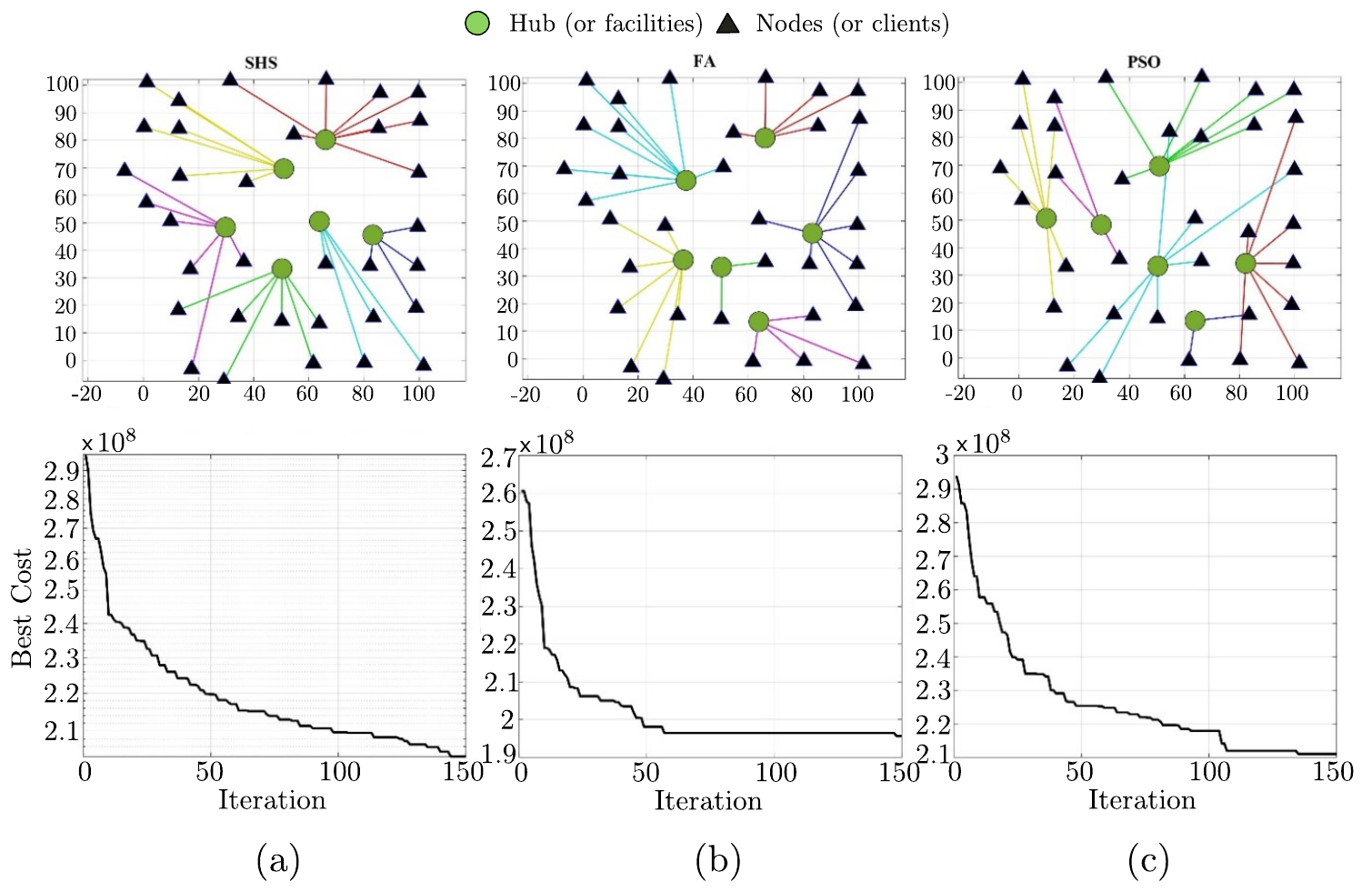}
    \caption{Illustration of Hub Location-Allocation Problem (HLP) by using (a) SHS, (b) FA, and (c) PSO. The circle marked in green represents the facilities (or hub). The triangle marked in black indicates clients (or nodes).}
    \label{fig:hub}
\end{figure}

One of the thriving research areas in classical facility location problems is the Hub Location-Allocation Problem (HLP). The overall objective of the hub-allocation problem is to allocate a hub to demand nodes. The problem simply could be presented in a facilities-clients (i.e., hub-nodes) relation. Considering multiple clients (i.e., nodes) in a 2-D view and these clients should get services from their facilities (or hub); for example, Internet Service Provider (ISP) facilities. The solution is to find the shortest path between clients and facilities in which all clients get their services with the least number of facilities (or hub). It is an NP-Hard problem, meaning the traditional methods cannot solve them in large sizes. We employ the SHS algorithm to evaluate its potential to solve the HLP problem. We considered 40 scattered nodes (or clients) and applied SHS to decide how to allocate facilities (or hubs) to each client (Figure \ref{fig:hub}). We found that the SHS algorithm successfully allocates facilities to each client lying nearest it while maintaining load balance at all the facilities (Figure \ref{fig:hub}a). We also compared the performance of SHS with FA and PSO algorithms. We found that both FA and PSO can also allocate facilities to all the nodes; however, both these algorithms fail to maintain load balance (Figures \ref{fig:hub}b and c). Some facilities are flooded with clients, whereas some have very few clients.

\subsection{SHS for Parallel Machine Scheduling (PMS)}
\begin{table}[ht!]

\caption{Processing time of each machine for all the tasks.}
\label{tab:pms}
\resizebox{\textwidth}{!}{
\begin{tabular}{ccccccccccccccccccccc}
\hline
\textbf{Machines} & \textbf{T1} & \textbf{T2} & \textbf{T3} & \textbf{T4} & \textbf{T5} & \textbf{T6} & \textbf{T7} & \textbf{T8} & \textbf{T9} & \textbf{T10} & \textbf{T11} & \textbf{T12} & \textbf{T13} & \textbf{T14} & \textbf{T15} & \textbf{T16} & \textbf{T17} & \textbf{T18} & \textbf{T19} & \textbf{T20} \\ \hline
\textbf{1} & 42 & 13 & 43 & 31 & 14 & 42 & 23 & 13 & 48 & 14 & 10 & 18 & 18 & 29 & 13 & 10 & 42 & 34 & 22 & 27 \\ \hline
\textbf{2} & 38 & 13 & 20 & 20 & 25 & 38 & 14 & 29 & 12 & 38 & 26 & 21 & 27 & 48 & 20 & 14 & 14 & 15 & 30 & 17 \\ \hline
\end{tabular}}
\end{table}

\begin{figure}[ht!]
    \centering
    \includegraphics[width=0.95\textwidth]{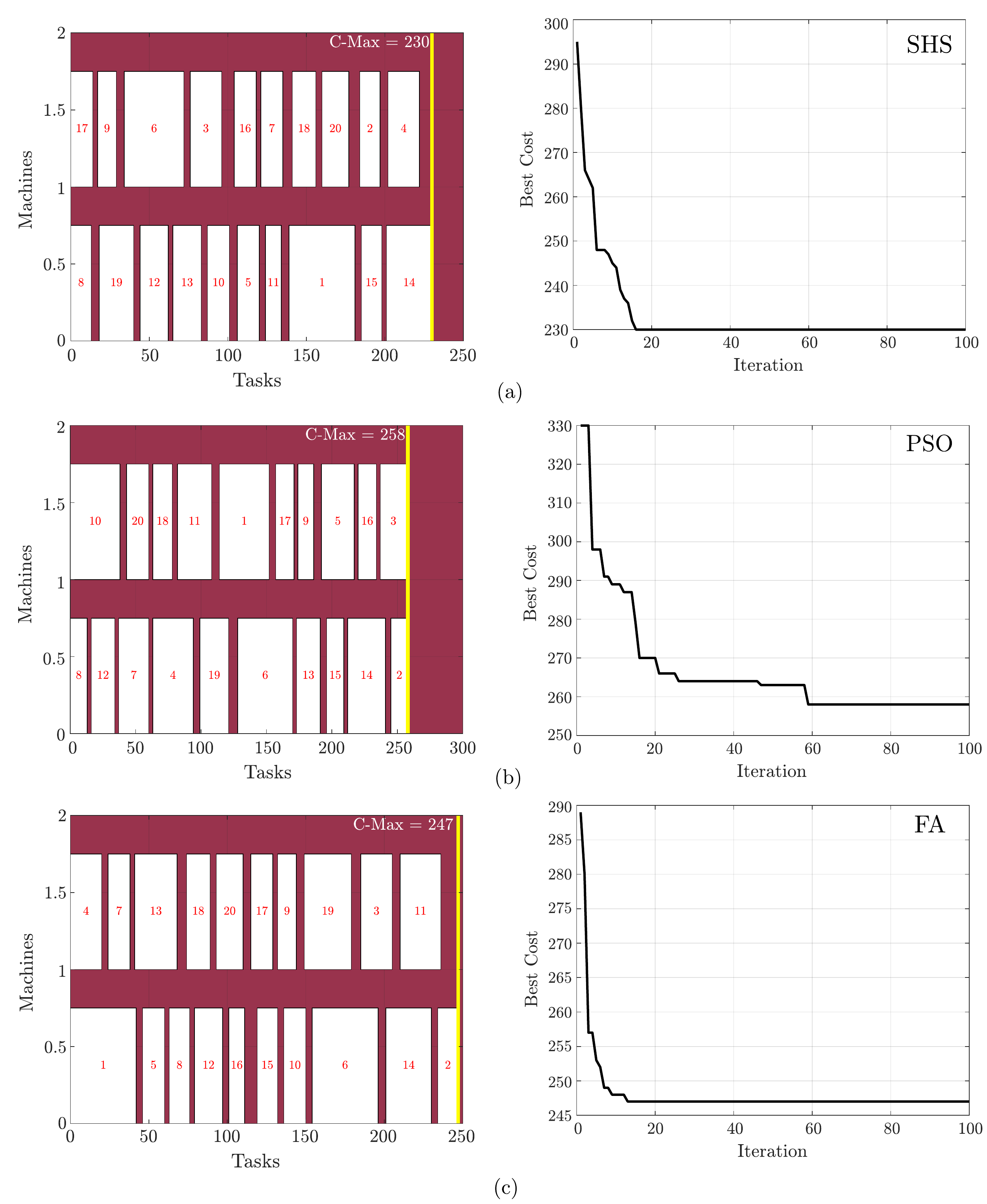}
    \caption{Illustration of the Parallel Machine Scheduling (PMS) of 20 tasks with two machines by using (a) SHS, (b) PSO, and (c) FA. The space between the white boxes in the upper panel represents the setup time.}
    \label{fig:pms}
\end{figure}

Parallel Machine Scheduling (PMS) stands out as a widely recognised and extensively studied issue within the realms of time management and operational research, often tackled through the utilisation of optimisation algorithms. Within this problem domain, multiple machines undertake various tasks, each possessing distinct lengths or levels of complexity. It's crucial to highlight that a single machine must exclusively execute each task, and no task should be duplicated. Moreover, the commencement time for each task equates to the sum of the preceding task's completion time and the setup time. Correspondingly, the culmination time coincides with the addition of the task's commencement time and its processing duration. To evaluate the potential of the SHS algorithm to solve the problem of PMS, we generated an experimental scenario with two machines and 20 tasks (Table \ref{tab:pms}). The processing time for all the tasks is randomly allocated (between 10 and 50 time units). Also, a setup time for each machine is randomly selected between 3 and 9 times. The objective of this problem is to minimise the maximum machine completion time (i.e., C-Max), which is the cost function. C-Max is the sum of the total time employed for machines to finish the complete tasks. Finally, we employed SHS, PSO, and FA for 100 iterations to optimise the problem and reported the corresponding C-Max in (Figure \ref{fig:pms}). We found that SHS finished all tasks with a C-Max of 230, outperforming the other two techniques (FA and PSO achieved a C-Max of 247 and 258, respectively).  

\subsection{SHS for Economic Dispatching (ED) of Electric Power System (EPS)}
Economic Dispatching (ED) of electric power systems is one of the important and critically studied subjects in electrical engineering and operational research \citep{yuan2019study}. Within the realm of ED, it becomes imperative to fulfill the Power Demand (PD) requirement by efficiently coordinating multiple power plants or generators, all while minimising associated costs. It's important to note that individual power generators generate varying amounts of power based on their specific technology and size, collectively contributing to what is referred to as the Power Total (PT). Usually, a loss factor is associated with the ED process, which is also considered in the calculation of the power demand. 

To evaluate the performance of SHS for Economic Dispatching (ED) of an Electric Power System (EPS), we considered an experiment scenario that consists of three power generators (i.e., P1, P2, and P3) having a minimum and maximum power generations cap of Min= [100 50 80] and Max= [500 200 300] for P1, P2, and P3, respectively. The Power Demand (PD) is 900 megawatts, which must be satisfied by the sum of the generated powers by all three power generators with low error. We applied SHS, PSO, and FA to optimise the power from each generator and the power loss to full fill the power demand. We found that the SHS algorithm fulfills the power demand with the least error (Table \ref{tab:ED_EPS}). 

\begin{table}[ht!]
\caption{Performance of SHS, PSO, and FA for Economic Dispatching (ED) of Electric Power System (EPS). The error is calculated by subtracting the power loss and the demand power from the total generated power. The lowest the error, the better the performance.}
\label{tab:ED_EPS}
\centering
\resizebox{0.4\textwidth}{!}{ 
\begin{tabular}{cccc}
\hline
\textbf{PD = 900 MW} & \textbf{SHS} & \textbf{PSO} & \textbf{FA} \\ \hline
\textbf{P1 (MW)} & 491.3283 & 485.7013 & 489.2253 \\ \hline
\textbf{P2 (MW)} & 172.6078 & 179.6767 & 174.4550 \\ \hline
\textbf{P3 (MW)} & 264.5088 & 263.0032 & 264.7629 \\ \hline
\textbf{PT (MW)} & 928.4449 & 928.3812 & 928.4431 \\ \hline
\textbf{Power Loss (PL)} & 28.4446 & 28.3788 & 28.4319 \\ \hline
\textbf{Error=PT-PL-PD} & 0.0002 & 0.0024 & 0.0112 \\ \hline
\textbf{Cost} & 10676 & 10673 & 10675 \\ \hline
\end{tabular}}
\end{table}

\section{Conclusion}
\label{sec:conclusion}
We developed a novel meta-heuristic algorithm, the Scorpion Hunting Strategy (SHS). The SHS is inspired by the hunting behaviour of scorpions. We performed qualitative, quantitative, and statistical analyses to examine the impact of scorpion hunting behaviours on various stages of the proposed SHS algorithm, including the search history, the overall average fitness, and the convergence curve. To test the local exploitation accuracy and global exploration capability of algorithms, ten well-known conventional functions were chosen. The local optimum avoidance ability and the efficiency of SHS in more complex test functions were tested using CEC2020 benchmark functions. Experimental findings show that the proposed SHS algorithm greatly outperforms twelve cutting-edge comparison algorithms on several test benchmark functions. The SHS has high convergence speed and effective local optimum avoidance. The outcomes of the Friedman rank test and Wilcoxon rank sum test demonstrate SHO's dominance for the majority of test functions.
Overall, SHS holds the first rank in 13 out of 20 test functions (65\%), which is much higher than 12 other algorithms. HS and BA achieve the first rank in two benchmark functions (10\% each), and GA, PSO, and FA achieve the first rank in one benchmark function (5\% each). Finally, the applicability of the SHS algorithm in six different real-world applications with low computation complexity verifies its broad range ability.

As part of future work, a binary version of the SHS algorithm can be developed to cover the broad domain of optimisation, including discrete and multi-objective optimisation. The regression capability of the SHS can be further investigated by coupling it with different machine learning setups for exploring its applicability in a broader range of fields. Furthermore, graph theory-related applications, such as finding minimum-spanning trees, can be explored in Geographic Information Science (GIS) and other applications in Earth and Environmental Sciences (EES) to study some intriguing possibilities. For instance, it can be used to study network connectivity, which is a critical issue in geomorphology, hydrology, and ecology. The future work may be extended to address a diverse array of challenges in complex systems. This algorithm holds the potential to enhance modeling accuracy and parameter estimation through its integration with advanced modeling techniques while also providing a robust framework for solving inverse problems \citep{kalayci2020mutual,ccevik2017voxel,taylan2021new,ozougur2020ensemble,savku2017optimal,baltas2022optimal}. Additionally, the algorithm's applicability to the game theory under uncertainties will revolutionise decision-making in adversarial scenarios with incomplete information \citep{ergun2023game,usta2018cooperative,palanci2013forest}. By venturing into multi-objective optimisation and uncertainty quantification, the algorithm will provide a comprehensive solution suite for robust decision-making in complex systems, ultimately reshaping the landscape of meta-heuristic optimisation. Hence, SHS has immense potential to improve real-world complex optimisation solutions.

\section*{CRediT author statement}
\noindent\textbf{Abhilash Singh:} Conceptualisation, Methodology, Software, Data Curation, Validation, Writing- Original draft preparation, Visualization, Investigation, Writing- Reviewing and Editing.
\textbf{Seyed Muhammad Hossein Mousavi:} Conceptualisation, Methodology, Software, Data Curation, Visualization, Writing- Original draft preparation, Visualization, Writing- Reviewing and Editing.
\textbf{Kumar Gaurav:}  Methodology, Data Curation, Visualization, Investigation, Writing- Reviewing and Editing, Supervision, Project Administration.

\section*{Code Availability}
\noindent The datasets used for this study can be made available from the corresponding author upon reasonable request. The code developed for this study will be shared on website:
\url{https://abhilashsingh.net/codes.html} upon acceptance of this article. 

\section*{Acknowledgments}
\noindent The authors would like to acknowledge IISER Bhopal for providing institutional support. They would like to thank to the editor and all the three anonymous reviewers for providing helpful comments and suggestions.
\newpage
\appendix
\setcounter{table}{0}
\setcounter{figure}{0}
\section{}
\begin{table}[ht!]
\caption{Comparison of the best cost value after 100 iterations.}
\resizebox{\textwidth}{!}{  
\begin{tabular}{ccccccccccccccc}
\hline
\rowcolor[HTML]{EFEFEF} 
\textbf{Function} & \textbf{Value} & \textbf{GA} & \textbf{PSO} & \textbf{FA} & \textbf{BBO} & \textbf{ABC} & \textbf{SHS} & \textbf{TLBO} & \textbf{ACO} & \textbf{HS} & \textbf{SA} & \textbf{DE} & \textbf{BA} & \textbf{IWO} \\ \hline
 & Best & 1.4162 & 1.0222 & 0.3973 & 1.3916 & 1.0235 & \textbf{0.2145} & 3.7784 & 4.5754 & 3.4852 & 3.3308 & 4.0139 & 3.0445 & 2.581 \\ \cline{2-15} 
 & Median & 1.4414 & 1.0418 & 0.4016 & 1.3954 & 1.0265 & 0.2155 & 3.8419 & 4.5986 & 3.542 & 3.347 & 4.0884 & 3.0658 & 2.5912 \\ \cline{2-15} 
 & Worst & 1.6087 & 1.1561 & 0.4102 & 1.4246 & 1.1254 & 0.2556 & 4.0356 & 5.1109 & 3.6412 & 3.8312 & 4.2186 & 3.3526 & 2.9414 \\ \cline{2-15} 
\multirow{-4}{*}{\textbf{Ackley}} & Avg & 1.4888 & 1.0734 & 0.403 & 1.4038 & 1.0585 & \textbf{0.2285} & 3.8853 & 4.7616 & 3.5561 & 3.503 & 4.107 & 3.1543 & 2.7046 \\ \hline
\rowcolor[HTML]{EFEFEF} 
 &  &  &  &  &  &  &  &  &  &  &  &  &  &  \\ \hline
 & Best & 24.15 & 41.2025 & 27.967 & 25.1852 & 57.7271 & \textbf{12.8248} & 55.6236 & 57.0112 & 31.5092 & 58.8846 & 55.6199 & 64.685 & 165.2024 \\ \cline{2-15} 
 & Median & 24.1866 & 41.2469 & 28.2639 & 18.4686 & 58.8054 & 12.8581 & 56.2564 & 57.5464 & 31.5167 & 59.2816 & 55.8003 & 65.7125 & 166.2307 \\ \cline{2-15} 
 & Worst & 25.3627 & 46.2789 & 30.6143 & 19.4642 & 67.3193 & 14.326 & 61.7391 & 67.4692 & 33.3105 & 67.8021 & 64.0044 & 69.6068 & 183.9635 \\ \cline{2-15} 
\multirow{-4}{*}{\textbf{Rastrigin}} & Avg & 24.5664 & 42.9094 & 28.9484 & 18.706 & 61.2839 & \textbf{13.3363} & 57.873 & 60.6756 & 32.1121 & 61.9894 & 58.4748 & 66.6681 & 171.7989 \\ \hline
\rowcolor[HTML]{EFEFEF} 
 &  &  &  &  &  &  &  &  &  &  &  &  &  &  \\ \hline
 & Best & 162.2222 & 2.0715 & \textbf{0.0625} & 1.9429 & 251.046 & 1.8013 & 6.7425 & 350.1948 & 181.7691 & 12.138 & 2.6712 & 188.8117 & 30.1434 \\ \cline{2-15} 
 & Median & 163.9688 & 2.1127 & 0.0625 & 1.9601 & 251.5815 & 1.8359 & 6.7431 & 355.6221 & 184.7402 & 12.3488 & 2.6756 & 190.3213 & 30.3 \\ \cline{2-15} 
 & Worst & 179.3702 & 2.1401 & 0.07 & 2.0451 & 283.8868 & 2.0495 & 7.7513 & 381.7502 & 184.8163 & 12.6938 & 3.159 & 194.5658 & 35.1219 \\ \cline{2-15} 
\multirow{-4}{*}{\textbf{Zakharov}} & Avg & 168.5204 & 2.1081 & \textbf{0.065} & 1.9827 & 262.1714 & 1.8955 & 7.0789 & 362.5223 & 183.7752 & 12.3935 & 2.8353 & 191.2329 & 31.855 \\ \hline
\rowcolor[HTML]{EFEFEF} 
\multicolumn{15}{c}{\cellcolor[HTML]{EFEFEF}} \\ \hline
 & Best & 8.473 & \textbf{1.92E-09} & 1.00E-05 & 2.00E-05 & 0.0018 & 2.00E-05 & 0.0009 & 0.0133 & 6.01E-07 & 0.0005 & 0.0024 & 0.1575 & 0.0027 \\ \cline{2-15} 
 & Median & 8.5325 & 1.93E-09 & 1.46E-05 & 1.51E-05 & 0.0018 & 2.07E-05 & 0.0009 & 0.0133 & 6.06E-07 & 0.0005 & 0.0025 & 0.1604 & 0.0027 \\ \cline{2-15} 
 & Worst & 9.8288 & 2.08E-09 & 1.71E-05 & 1.56E-05 & 0.0018 & 2.13E-05 & 0.0009 & 0.0156 & 6.70E-07 & 0.0005 & 0.0025 & 0.1843 & 0.003 \\ \cline{2-15} 
\multirow{-4}{*}{\textbf{Booth}} & Avg & 8.9448 & \textbf{1.98E-09} & 1.54E-05 & 1.52E-05 & 0.0018 & 2.09E-05 & 0.0009 & 0.0141 & 6.25E-07 & 0.0005 & 0.0025 & 0.1674 & 0.0028 \\ \hline
\rowcolor[HTML]{EFEFEF} 
\multicolumn{15}{c}{\cellcolor[HTML]{EFEFEF}} \\ \hline
 & Best & 21.7321 & 0.0498 & 0.2068 & 0.2491 & 75.8654 & \textbf{0.0269} & 3.4589 & 33.8538 & 0.6408 & 3.3664 & 3.807 & 44.0604 & 77.496 \\ \cline{2-15} 
 & Median & 21.982 & 0.0498 & 0.2077 & 0.2508 & 77.1113 & 0.0269 & 3.4619 & 33.9682 & 0.649 & 3.4156 & 3.8563 & 44.4577 & 78.3437 \\ \cline{2-15} 
 & Worst & 23.8585 & 0.0531 & 0.244 & 0.2674 & 77.5526 & 0.0311 & 3.7285 & 35.4902 & 0.6925 & 3.4313 & 3.9075 & 52.3618 & 92.3152 \\ \cline{2-15} 
\multirow{-4}{*}{\textbf{DeJong}} & Avg & 22.5242 & 0.0509 & 0.2195 & 0.2557 & 76.8431 & \textbf{0.0283} & 3.5497 & 34.4374 & 0.6608 & 3.4044 & 3.8569 & 46.9599 & 82.7183 \\ \hline
\rowcolor[HTML]{EFEFEF} 
\multicolumn{15}{c}{\cellcolor[HTML]{EFEFEF}} \\ \hline
 & Best & 1.7539 & 6.00E-05 & 1.00E-05 & 0.6431 & 0.0001 & \textbf{4.88E-07} & 4.00E-05 & 0.0095 & 0.0635 & 0.0005 & 0.0013 & 0.2833 & 0.203 \\ \cline{2-15} 
 & Median & 1.7774 & 6.43E-05 & 1.19E-05 & 0.653 & 0.0001 & 4.99E-07 & 4.30E-05 & 0.0096 & 0.065 & 0.0005 & 0.0013 & 0.2913 & 0.2083 \\ \cline{2-15} 
 & Worst & 1.9098 & 7.80E-05 & 1.24E-05 & 0.7757 & 0.0001 & 5.41E-07 & 5.01E-05 & 0.0117 & 0.065 & 0.0006 & 0.0016 & 0.3247 & 0.2295 \\ \cline{2-15} 
\multirow{-4}{*}{\textbf{Beale}} & Avg & 1.8137 & 6.87E-05 & 1.20E-05 & 0.6906 & 0.0001 & \textbf{5.09E-07} & 4.51E-05 & 0.0103 & 0.0645 & 0.0005 & 0.0014 & 0.2997 & 0.2136 \\ \hline
\rowcolor[HTML]{EFEFEF} 
\multicolumn{15}{c}{\cellcolor[HTML]{EFEFEF}} \\ \hline
 & Best & 240.6615 & 0.743 & 3.6888 & 0.8916 & 495.9886 & \textbf{0.1679} & 22.3489 & 199.0454 & 35.8076 & 28.9715 & 35.6503 & 412.3279 & 213.6543 \\ \cline{2-15} 
 & Median & 244.405 & 0.7773 & 3.7681 & 0.8998 & 518.4291 & 0.1761 & 22.8393 & 200.1512 & 36.2696 & 29.5635 & 36.7107 & 417.7337 & 220.0942 \\ \cline{2-15} 
 & Worst & 306.8622 & 0.9742 & 4.7715 & 0.9842 & 570.696 & 0.2074 & 24.9245 & 283.0732 & 39.2946 & 32.2441 & 38.6932 & 459.264 & 260.1987 \\ \cline{2-15} 
\multirow{-4}{*}{\textbf{Powell}} & Avg & 263.9762 & 0.8315 & 4.0761 & 0.9252 & 528.3712 & \textbf{0.1838} & 23.3709 & 227.4232 & 37.1239 & 30.2597 & 37.018 & 429.7752 & 231.3157 \\ \hline
\rowcolor[HTML]{EFEFEF} 
\multicolumn{15}{c}{\cellcolor[HTML]{EFEFEF}} \\ \hline
 & Best & -4.3564 & -4.1933 & -3.1876 & -6.6347 & -2.6504 & -6.3448 & -3.912 & -2.7495 & \textbf{-6.9323} & -4.1937 & -3.8373 & -3.2789 & -5.9814 \\ \cline{2-15} 
 & Median & -4.202 & -4.1363 & -3.114 & -6.4026 & -2.6103 & -6.2342 & -3.7671 & -2.7057 & -6.5893 & -3.9464 & -3.6691 & -3.2112 & -5.7717 \\ \cline{2-15} 
 & Worst & -2.6072 & -3.7141 & -2.482 & -4.8006 & -1.0893 & -5.7494 & -1.9038 & -1.6858 & -5.3672 & -3.7132 & -2.8512 & -2.3305 & -5.5485 \\ \cline{2-15} 
\multirow{-4}{*}{\textbf{Michalewicz}} & Avg & -3.7219 & -4.0146 & -2.9278 & -5.9459 & -2.1167 & -6.1095 & -3.1943 & -2.3803 & \textbf{-6.2962} & -3.9511 & -3.4525 & -2.9402 & -5.7671 \\ \hline
\rowcolor[HTML]{EFEFEF} 
\multicolumn{15}{c}{\cellcolor[HTML]{EFEFEF}} \\ \hline
 & Best & -20.7917 & -19.9923 & -31.0754 & -21.0719 & -22.2571 & \textbf{-31.4117} & -2.3112 & -8.8059 & -20.1338 & -7.6774 & -3.4488 & -10.3669 & -16.5684 \\ \cline{2-15} 
 & Median & -20.7162 & -19.5133 & -30.093 & -20.5449 & -21.3799 & -30.2281 & -2.2706 & -8.5778 & -20.1151 & -7.2241 & -3.4141 & -10.3008 & -16.1982 \\ \cline{2-15} 
 & Worst & -14.3222 & -8.3211 & -18.9748 & -10.9532 & -16.1969 & -23.2623 & -1.1667 & -8.3648 & -18.525 & -6.8787 & -2.6398 & -5.1956 & -8.5821 \\ \cline{2-15} 
\multirow{-4}{*}{\textbf{Trid}} & Avg & -18.61 & -15.9422 & -26.7144 & -17.5233 & -19.9446 & \textbf{-28.3007} & -1.9161 & -8.5828 & -19.5913 & -7.26 & -3.1675 & -8.6211 & -13.7829 \\ \hline
\rowcolor[HTML]{EFEFEF} 
\multicolumn{15}{c}{\cellcolor[HTML]{EFEFEF}} \\ \hline
 & Best & 0.2144 & 0.3173 & 0.9232 & 0.1972 & 1.2847 & \textbf{0.0063} & 1.0123 & 0.4647 & 0.0375 & 1.0804 & 0.8354 & 0.8898 & 14.0124 \\ \cline{2-15} 
 & Median & 0.2208 & 0.3247 & 0.9553 & 0.2041 & 1.3257 & 0.016 & 1.0158 & 0.4721 & 0.0384 & 1.1157 & 0.8524 & 0.9263 & 14.5156 \\ \cline{2-15} 
 & Worst & 0.2259 & 0.461 & 1.3292 & 0.2778 & 1.4522 & 0.0182 & 1.0237 & 0.5635 & 0.0433 & 1.1676 & 0.91 & 1.0779 & 14.6725 \\ \cline{2-15} 
\multirow{-4}{*}{\textbf{Levy}} & Avg & 0.2204 & 0.3676 & 1.0692 & 0.2263 & 1.3542 & \textbf{0.0135} & 1.0173 & 0.5001 & 0.0397 & 1.1212 & 0.8659 & 0.9647 & 14.4001 \\ \hline
\end{tabular}
}
\label{tab:100iterations}
\end{table}

\begin{table}[ht!]
\caption{Comparison of the best cost value after 1000 iterations.}
\resizebox{\textwidth}{!}{  
\begin{tabular}{ccccccccccccccc}
\hline
\rowcolor[HTML]{EFEFEF} 
\textbf{Function} & \textbf{Value} & \textbf{GA} & \textbf{PSO} & \textbf{FA} & \textbf{BBO} & \textbf{ABC} & \textbf{SHS} & \textbf{TLBO} & \textbf{ACO} & \textbf{HS} & \textbf{SA} & \textbf{DE} & \textbf{BA} & \textbf{IWO} \\ \hline
 & Best & 0.0007 & 9.28E-05 & 6.22E-10 & 0.0011 & 0.0052 & 4.44E-15 & 2.80E-05 & 1.17E-06 & 0.0073 & 2.39E-08 & \textbf{2.22E-15} & 1.43E-05 & 0.001 \\ \cline{2-15} 
 & Median & 0.0007 & 9.68E-05 & 6.38E-10 & 0.0111 & 0.0054 & 4.65E-15 & 2.84E-05 & 1.20E-06 & 0.0074 & 2.48E-08 & 2.29E-15 & 1.44E-05 & 0.001 \\ \cline{2-15} 
 & Worst & 0.001 & 0.0001 & 7.23E-10 & 0.0221 & 0.0068 & 6.16E-15 & 3.40E-05 & 1.22E-06 & 0.0083 & 2.58E-08 & 2.53E-15 & 1.75E-05 & 0.0012 \\ \cline{2-15} 
\multirow{-4}{*}{\textbf{Ackley}} & Avg & 0.0008 & 0.0001 & 6.61E-10 & 0.0114 & 0.0058 & 5.08E-15 & 3.01E-05 & 1.20E-06 & 0.0077 & 2.48E-08 & \textbf{2.34E-15} & 1.54E-05 & 0.0011 \\ \hline
\rowcolor[HTML]{EFEFEF} 
 &  &  &  &  &  &  &  &  &  &  &  &  &  &  \\ \hline
 & Best & 1.0341 & 0.9949 & 7.9596 & 3.5831 & 37.0984 & \textbf{0.5337} & 6.9647 & 33.0328 & 1.2478 & 8.9546 & 2.9851 & 32.3415 & 4.975 \\ \cline{2-15} 
 & Median & 1.0675 & 1.0485 & 8.1393 & 3.78 & 38.3225 & 1.5351 & 7.0504 & 34.7017 & 1.2892 & 9.0503 & 3.0563 & 33.7166 & 5.0418 \\ \cline{2-15} 
 & Worst & 1.1963 & 1.46 & 12.6644 & 4.4471 & 57.3677 & 1.9861 & 6.9974 & 46.9568 & 1.5857 & 14.6962 & 2.9874 & 42.8109 & 6.4528 \\ \cline{2-15} 
\multirow{-4}{*}{\textbf{Rastrigin}} & Avg & 1.0993 & \textbf{1.1678} & 9.5878 & 3.9367 & 44.2629 & 1.3516 & 7.0042 & 38.2304 & 1.3742 & 10.9004 & 3.0096 & 36.2897 & 5.4899 \\ \hline
\rowcolor[HTML]{EFEFEF} 
 &  &  &  &  &  &  &  &  &  &  &  &  &  &  \\ \hline
 & Best & 4.4082 & 9.14E-17 & 7.69E-19 & 0.2193 & 1.4869 & 5.77E-10 & \textbf{9.41E-21} & 112.7676 & 3.9019 & 9.80E-16 & 2.38E-06 & 10.1865 & 2.72E-06 \\ \cline{2-15} 
 & Median & 4.608 & 9.58E-17 & 8.03E-19 & 0.2338 & 1.5086 & 6.06E-10 & 9.56E-21 & 113.7101 & 4.0677 & 1.01E-15 & 2.45E-06 & 10.6585 & 2.86E-06 \\ \cline{2-15} 
 & Worst & 6.6868 & 1.07E-16 & 1.26E-18 & 0.2606 & 2.2836 & 6.54E-10 & 1.13E-20 & 119.9598 & 5.4757 & 1.45E-15 & 3.29E-06 & 13.2222 & 3.94E-06 \\ \cline{2-15} 
\multirow{-4}{*}{\textbf{Zakharov}} & Avg & 5.2343 & 9.81E-17 & 9.45E-19 & 0.2379 & 1.7597 & 6.13E-10 & \textbf{1.00E-20} & 115.4792 & 4.4817 & 1.14E-15 & 2.70E-06 & 11.3557 & 3.17E-06 \\ \hline
\rowcolor[HTML]{EFEFEF} 
\multicolumn{15}{c}{\cellcolor[HTML]{EFEFEF}} \\ \hline
 & Best & 0.0001 & \textbf{4.43E-28} & 3.63E-14 & 2.19E-07 & 1.22E-16 & 9.98E-23 & 0.0022 & 3.03E-19 & 3.79E-08 & 8.91E-20 & 6.47E-05 & 4.84E-09 & 1.14E-10 \\ \cline{2-15} 
 & Median & 0.0021 & 4.63E-28 & 3.75E-14 & 2.28E-07 & 1.26E-16 & 1.04E-22 & 0.0023 & 3.18E-19 & 3.93E-08 & 9.52E-20 & 6.57E-05 & 4.88E-09 & 1.15E-10 \\ \cline{2-15} 
 & Worst & 0.0031 & 6.48E-28 & 4.68E-14 & 3.48E-07 & 1.93E-16 & 1.17E-22 & 0.0031 & 4.26E-19 & 5.23E-08 & 1.43E-19 & 7.68E-05 & 5.92E-09 & 1.24E-10 \\ \cline{2-15} 
\multirow{-4}{*}{\textbf{Booth}} & Avg & 0.0017 & \textbf{5.18E-28} & 4.02E-14 & 2.65E-07 & 1.47E-16 & 1.07E-22 & 0.0025 & 3.49E-19 & 4.32E-08 & 1.09E-19 & 6.91E-05 & 5.22E-09 & 1.18E-10 \\ \hline
\rowcolor[HTML]{EFEFEF} 
\multicolumn{15}{c}{\cellcolor[HTML]{EFEFEF}} \\ \hline
 & Best & 5.55E-11 & 0.0015 & 1.46E-10 & 0.0114 & 1.30E-05 & \textbf{3.17E-19} & 1.46E-09 & 2.27E-07 & 3.65E-06 & 2.53E-16 & 1.53E-12 & 1.62E-09 & 1.29E-06 \\ \cline{2-15} 
 & Median & 5.75E-11 & 0.0016 & 1.54E-10 & 0.0118 & 1.34E-05 & 3.31E-19 & 1.54E-09 & 2.35E-07 & 3.73E-06 & 2.56E-16 & 1.59E-12 & 1.65E-09 & 1.29E-06 \\ \cline{2-15} 
 & Worst & 5.79E-11 & 0.022 & 1.91E-10 & 0.0143 & 2.01E-05 & 4.56E-19 & 2.26E-09 & 3.75E-07 & 6.13E-06 & 2.88E-16 & 1.67E-12 & 2.42E-09 & 1.37E-06 \\ \cline{2-15} 
\multirow{-4}{*}{\textbf{DeJong}} & Avg & 5.70E-11 & 0.0083 & 1.64E-10 & 0.0125 & 1.55E-05 & \textbf{3.68E-19} & 1.76E-09 & 2.79E-07 & 4.50E-06 & 2.66E-16 & 1.60E-12 & 1.90E-09 & 1.31E-06 \\ \hline
\rowcolor[HTML]{EFEFEF} 
\multicolumn{15}{c}{\cellcolor[HTML]{EFEFEF}} \\ \hline
 & Best & 0.0012 & 1.11E-09 & 1.26E-14 & 3.25E-07 & 4.20E-12 & \textbf{2.77E-32} & 3.90E-20 & 1.94E-16 & 6.42E-08 & 6.72E-21 & 2.57E-23 & 9.87E-05 & 6.36E-10 \\ \cline{2-15} 
 & Median & 0.0013 & 1.12E-09 & 1.34E-14 & 3.35E-07 & 4.47E-12 & 2.84E-32 & 4.12E-20 & 1.99E-16 & 6.78E-08 & 7.07E-21 & 2.63E-23 & 0.0001 & 6.72E-10 \\ \cline{2-15} 
 & Worst & 0.0019 & 1.29E-09 & 1.66E-14 & 4.82E-07 & 5.26E-12 & 4.19E-32 & 4.99E-20 & 2.86E-16 & 9.58E-08 & 8.80E-21 & 2.60E-23 & 0.0001 & 8.26E-10 \\ \cline{2-15} 
\multirow{-4}{*}{\textbf{Beale}} & Avg & 0.0015 & 1.17E-09 & 1.42E-14 & 3.81E-07 & 4.65E-12 & \textbf{3.27E-32} & 4.34E-20 & 2.26E-16 & 7.59E-08 & 7.53E-21 & 2.60E-23 & 0.0001 & 7.11E-10 \\ \hline
\rowcolor[HTML]{EFEFEF} 
\multicolumn{15}{c}{\cellcolor[HTML]{EFEFEF}} \\ \hline
 & Best & 0.0505 & 0.0015 & 1.52E-09 & 1.0383 & 0.0461 & \textbf{2.90E-18} & 2.76E-15 & 5.87E-07 & 2.1321 & 1.93E-15 & 3.54E-08 & 0.0276 & 1.25E-05 \\ \cline{2-15} 
 & Median & 0.0512 & 0.0015 & 1.61E-09 & 1.0983 & 0.0486 & 2.96E-18 & 2.86E-15 & 5.91E-07 & 2.1488 & 1.94E-15 & 3.70E-08 & 0.0286 & 1.26E-05 \\ \cline{2-15} 
 & Worst & 0.0841 & 0.0018 & 2.24E-09 & 1.3571 & 0.073 & 4.46E-18 & 3.08E-15 & 9.42E-07 & 3.6095 & 2.62E-15 & 5.73E-08 & 0.0391 & 1.38E-05 \\ \cline{2-15} 
\multirow{-4}{*}{\textbf{Powell}} & Avg & 0.062 & 0.0016 & 1.79E-09 & 1.1646 & 0.0559 & \textbf{3.44E-18} & 2.90E-15 & 7.07E-07 & 2.6302 & 2.16E-15 & 4.32E-08 & 0.0318 & 1.30E-05 \\ \hline
\rowcolor[HTML]{EFEFEF} 
\multicolumn{15}{c}{\cellcolor[HTML]{EFEFEF}} \\ \hline
 & Best & -9.8209 & -8.4521 & -9.2361 & -8.5959 & -4.9605 & -9.5645 & -9.0291 & -3.4401 & -9.6645 & -7.3356 & -9.7231 & \textbf{-9.8441} & -8.0505 \\ \cline{2-15} 
 & Median & -9.2278 & -8.1656 & -8.6899 & -8.4699 & -4.7687 & -9.4428 & -9.0089 & -3.2921 & -9.4193 & -7.3102 & -9.3899 & -5.7114 & -7.9812 \\ \cline{2-15} 
 & Worst & -5.2395 & -7.9582 & -5.1891 & -4.6189 & -2.4265 & -8.5007 & -2.8206 & -1.5882 & -5.7309 & -2.5688 & -5.7749 & -4.7271 & -7.3695 \\ \cline{2-15} 
\multirow{-4}{*}{\textbf{Michalewicz}} & Avg & -8.0961 & -8.192 & -7.705 & -7.2283 & -4.0519 & \textbf{-9.1693} & -6.9529 & -2.7735 & -8.2715 & -5.7382 & -8.296 & -6.7609 & -7.8004 \\ \hline
\rowcolor[HTML]{EFEFEF} 
\multicolumn{15}{c}{\cellcolor[HTML]{EFEFEF}} \\ \hline
 & Best & \textbf{-32.0595} & -32.0594 & \textbf{-32.0595} & -31.2598 & -31.7512 & \textbf{-32.0595} & \textbf{-32.0595} & -32.0579 & -32.0594 & \textbf{-32.0595} & \textbf{-32.0595} & \textbf{-32.0595} & \textbf{-32.0595} \\ \cline{2-15} 
 & Median & -30.6922 & -30.6585 & -31.6262 & -30.934 & -30.8543 & -31.0331 & -31.391 & -31.7278 & -30.5399 & -31.928 & -31.8901 & -31.8318 & -31.7926 \\ \cline{2-15} 
 & Worst & -26.7891 & -28.3016 & -27.2103 & -30.1931 & -15.6166 & -24.8303 & -18.2458 & -14.2295 & -19.2583 & -18.3175 & -20.6416 & -28.8809 & -19.483 \\ \cline{2-15} 
\multirow{-4}{*}{\textbf{Trid}} & Avg & -29.8469 & -30.3398 & -30.2987 & -30.7956 & -26.074 & -29.3076 & -27.2321 & -26.0051 & -27.2859 & -27.435 & -28.1971 & \textbf{-30.9241} & -27.7783 \\ \hline
\rowcolor[HTML]{EFEFEF} 
\multicolumn{15}{c}{\cellcolor[HTML]{EFEFEF}} \\ \hline
 & Best & 2.00E-06 & 2.88E-10 & 1.17E-10 & 0.0069 & 0.1163 & 7.36E-11 & 2.82E-16 & 1.31E-08 & 2.48E-06 & \textbf{1.60E-16} & 1.87E-16 & 1.57E-09 & 3.2722 \\ \cline{2-15} 
 & Median & 2.02E-06 & 2.93E-10 & 1.19E-10 & 0.0072 & 0.1211 & 7.39E-11 & 2.85E-16 & 1.33E-08 & 2.49E-06 & 1.65E-16 & 1.91E-16 & 1.58E-09 & 3.3064 \\ \cline{2-15} 
 & Worst & 3.36E-06 & 3.68E-10 & 1.58E-10 & 0.0072 & 0.1328 & 1.12E-10 & 3.57E-16 & 2.25E-08 & 2.72E-06 & 2.88E-16 & 2.69E-16 & 2.45E-09 & 5.8886 \\ \cline{2-15} 
\multirow{-4}{*}{\textbf{Levy}} & Avg & 2.46E-06 & 3.16E-10 & 1.31E-10 & 0.0071 & 0.1234 & 8.67E-11 & 3.08E-16 & 1.63E-08 & 2.57E-06 & \textbf{2.04E-16} & 2.16E-16 & 1.87E-09 & 4.1557 \\ \hline
\end{tabular}}
\label{tab:1000iterations}
\end{table}

\section{}
\begin{table}[ht!]
\caption{Details of the CEC2020 benchmark functions.}
\label{tab:cec2020}
\centering
\resizebox{1\columnwidth}{!}{%
\begin{tabular}{cccc}
\hline
\textbf{Function} & \textbf{Description} & \textbf{Type} & \textbf{Optimum value} \\ \hline
F1 & Shifted and Rotated bent cigar function & Unimodal function & 100 \\ \hline
F2 & Shifted and rotated Schwefel’s function & Basic function & 1100 \\ \hline
F3 & Shifted and rotated Lunacek bi-Rastrigin function & Basic function & 700 \\ \hline
F4 & Expanded Rosenbrock’s plus Griewangk’s function & Basic function & 1900 \\ \hline
F5 & Hybrid function 1 (N=3) & Hybrid function & 1700 \\ \hline
F6 & Hybrid function 2 (N=4) & Hybrid function & 1600 \\ \hline
F7 & Hybrid function 3 (N=5) & Hybrid function & 2100 \\ \hline
F8 & Composition function 1 (N=3) & Composition function & 2200 \\ \hline
F9 & Composition function 2 (N=4) & Composition function & 2400 \\ \hline
F10 & Composition function 3 (N=5) & Composition function & 2500 \\ \hline
\end{tabular}}
\vspace{10ex}
\end{table}
\begin{figure}[ht!]
    \centering
    \includegraphics[width=1\textwidth]{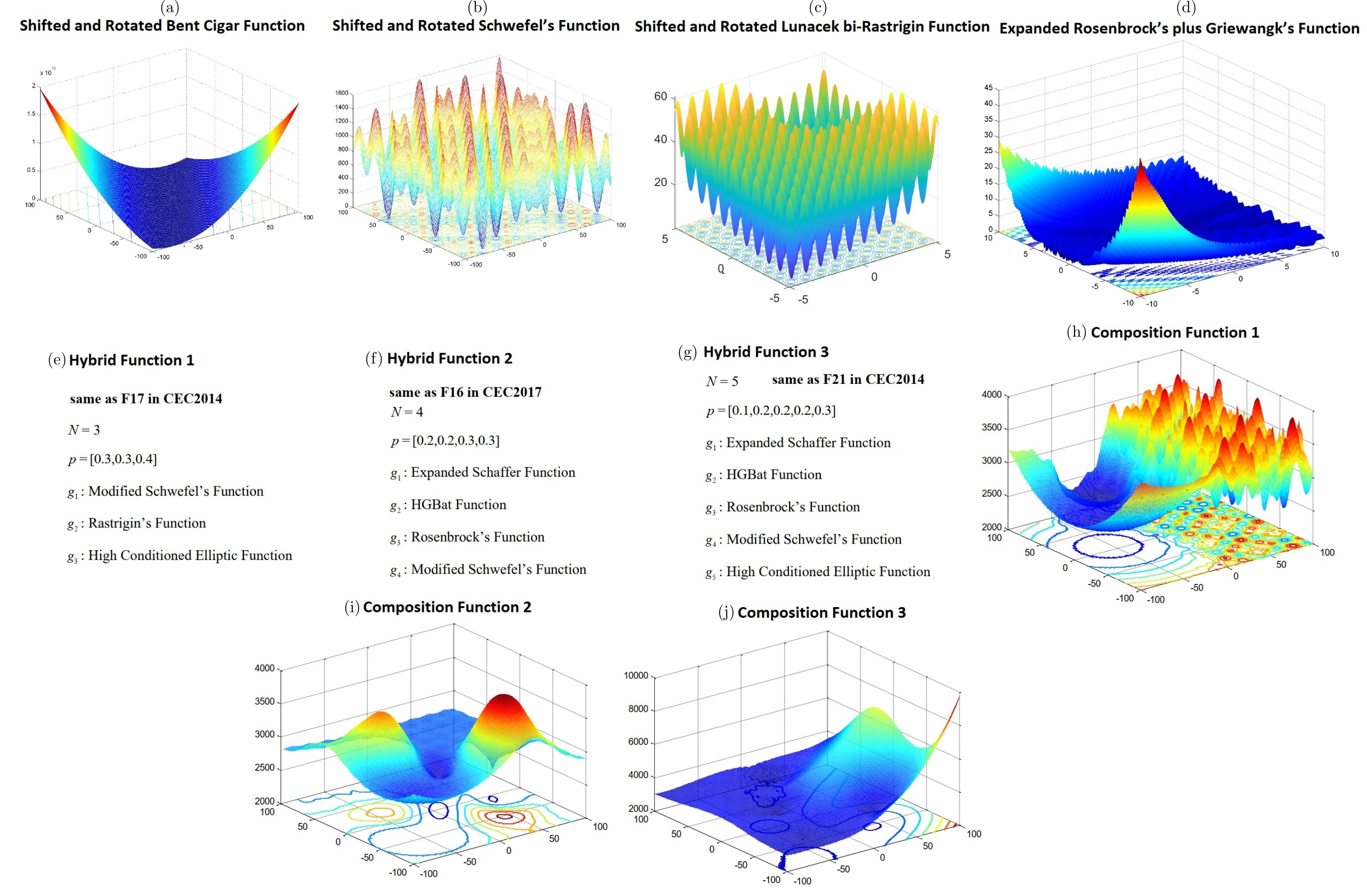}
    \caption{Illustration of the CEC2020 benchmark functions for evaluation in three dimensions.}
    \label{fig:cec2020funcplot}
\end{figure}

\begin{figure}[ht!]
    \centering
    \includegraphics[width=\textwidth]{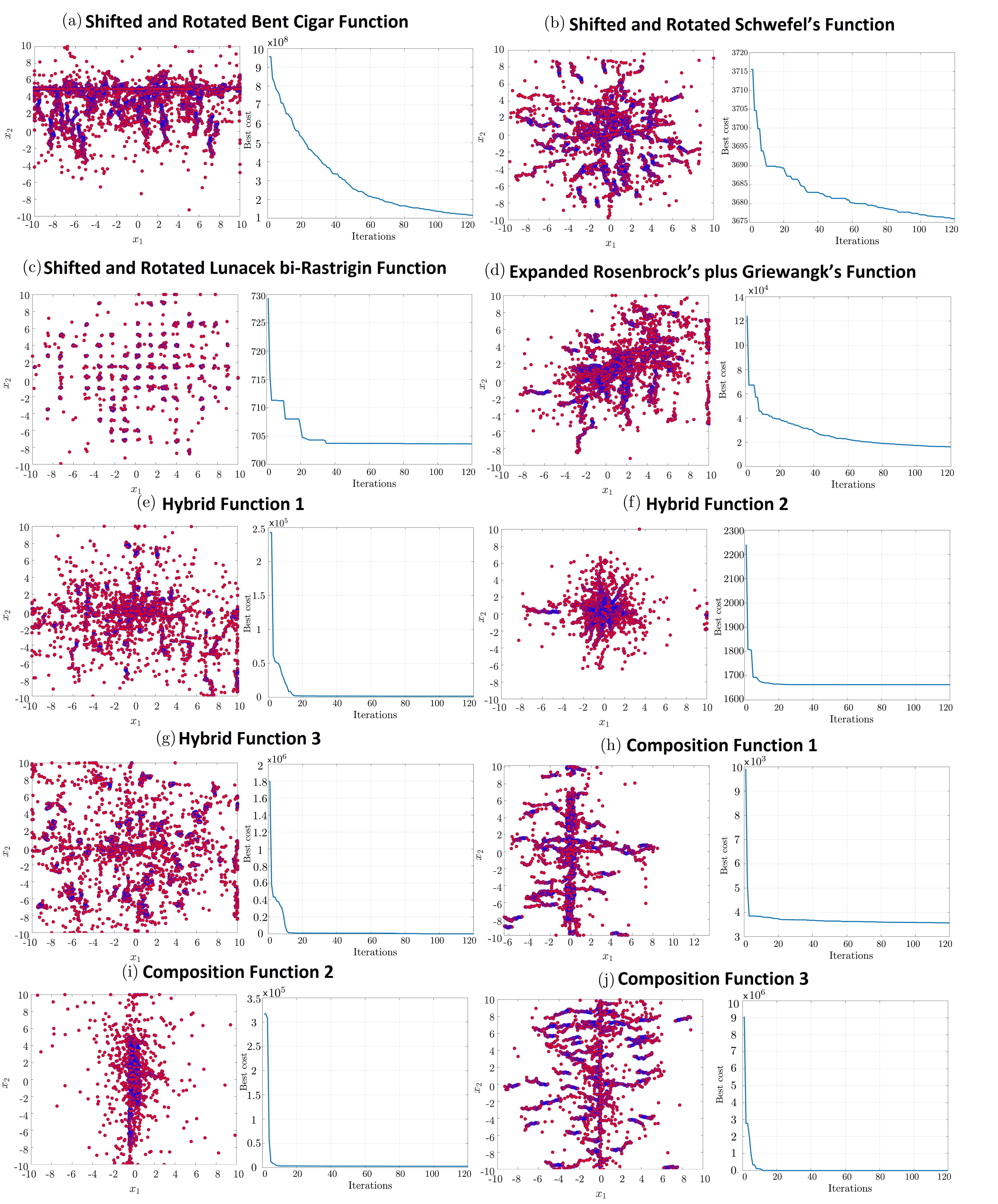}
    \caption{Illustrating the search history (i.e., the position of the scorpions) and the corresponding best cost curve for all the
CEC2020 benchmark functions.}
    \label{fig:cec2020searchplot}
\end{figure}

\begin{table}[]
\caption{Results of all the benchmark algorithms on CEC2020 benchmark functions.}
\label{tab:stat_cec2020}
\resizebox{\columnwidth}{!}{%
\begin{tabular}{ccccccccccccccc}
\hline
\rowcolor[HTML]{EFEFEF} 
\textbf{Functions} & \textbf{Value} & \textbf{GA} & \textbf{PSO} & \textbf{FA} & \textbf{BBO} & \textbf{ABC} & \textbf{SHS} & \textbf{TLBO} & \textbf{ACO} & \textbf{HS} & \textbf{SA} & \textbf{DE} & \textbf{BA} & \textbf{IWO} \\ \hline
 & Best & \textbf{100.00} & 102.05 & \textbf{100.00} & \textbf{100.00} & \textbf{100.00} & \textbf{100.00} & \textbf{100.00} & \textbf{100.00} & 105.25 & \textbf{100.00} & \textbf{100.00} & 94375 & 26109 \\ \cline{2-15} 
 & Median & 100.09 & 105.00 & 100.07 & 100.04 & 100.07 & 100.02 & 100.04 & 100.09 & 109.66 & 100.02 & 100.03 & 95221 & 29917 \\ \cline{2-15} 
 & Worst & 105.07 & 109.11 & 104.91 & 111.63 & 103.00 & 102.01 & 107.00 & 106.31 & 115.87 & 105.88 & 104.44 & 99554 & 31008 \\ \cline{2-15} 
\multirow{-4}{*}{\begin{tabular}[c]{@{}c@{}}CEC2020\\    \\ (F1)\end{tabular}} & Avg & 101.72 & 105.38 & 101.66 & 103.89 & 101.02 & \textbf{100.67} & 102.34 & 102.13 & 110.26 & 101.96 & 101.49 & 96383 & 29011 \\ \hline
\rowcolor[HTML]{EFEFEF} 
 &  &  &  &  &  &  &  &  &  &  &  &  &  &  \\ \hline
 & Best & 1100.004 & 1100.005 & \textbf{1100.000} & 1100.057 & 1126.814 & \textbf{1100.000} & 1100.002 & 1100.057 & 1100.004 & 1100.002 & 1100.002 & \textbf{1100.000} & \textbf{1100.000} \\ \cline{2-15} 
 & Median & 1100.009 & 1100.008 & 1100.004 & 1100.097 & 1126.991 & 1100.003 & 1100.004 & 1100.088 & 1100.011 & 1110.051 & 1100.070 & 1101.771 & 1120.900 \\ \cline{2-15} 
 & Worst & 1122.111 & 1181.400 & 1197.605 & 1109.667 & 1141.805 & 1109.012 & 1122.009 & 1200.001 & 1188.880 & 1152.673 & 1121.600 & 1145.544 & 1149.000 \\ \cline{2-15} 
\multirow{-4}{*}{\begin{tabular}[c]{@{}c@{}}CEC2020\\    \\ (F2)\end{tabular}} & Avg & 1107.374 & 1127.137 & 1132.536 & 1103.273 & 1131.870 & \textbf{1103.005} & 1107.338 & 1133.382 & 1129.631 & 1120.908 & 1107.224 & 1115.771 & 1123.300 \\ \hline
\rowcolor[HTML]{EFEFEF} 
 &  &  &  &  &  &  &  &  &  &  &  &  &  &  \\ \hline
 & Best & 705.178 & 760.406 & \textbf{700.000} & 736.980 & \textbf{700.000} & \textbf{700.000} & 705.43 & 700.958 & 705.567 & 707.428 & 706.768 & 700.181 & 761.147 \\ \cline{2-15} 
 & Median & 709.140 & 766.978 & 700.112 & 745.901 & 701.028 & 700.092 & 708.88 & 700.999 & 706.411 & 709.299 & 720.007 & 711.728 & 766.440 \\ \cline{2-15} 
 & Worst & 723.099 & 794.402 & 745.001 & 799.500 & 708.664 & 705.998 & 733.312 & 706.444 & 711.000 & 710.906 & 775.400 & 734.452 & 781.256 \\ \cline{2-15} 
\multirow{-4}{*}{\begin{tabular}[c]{@{}c@{}}CEC2020\\    \\ (F3)\end{tabular}} & Avg & 712.472 & 773.928 & 715.037 & 760.793 & 703.230 & \textbf{702.030} & 715.874 & 702.800 & 707.659 & 709.211 & 734.058 & 715.453 & 769.614 \\ \hline
\rowcolor[HTML]{EFEFEF} 
 &  &  &  &  &  &  &  &  &  &  &  &  &  &  \\ \hline
 & Best & 1938.48 & 1900.15 & 1904.16 & \textbf{1900.00} & \textbf{1900.00} & \textbf{1900.00} & \textbf{1900.00} & 1932.65 & 1908.84 & 1900.00 & 1900.00 & 1901.46 & 1999.01 \\ \cline{2-15} 
 & Median & 1952.41 & 1905.36 & 1998.30 & 1909.08 & 1911.07 & 1900.05 & 1900.88 & 1944.22 & 1922.74 & 1920.31 & 1908.82 & 1906.32 & 2005.60 \\ \cline{2-15} 
 & Worst & 1966.87 & 1922.87 & 2009.77 & 1952.67 & 1936.34 & 1911.87 & 1939.60 & 1968.00 & 1985.90 & 1990.32 & 1981.00 & 1965.32 & 2054.30 \\ \cline{2-15} 
\multirow{-4}{*}{\begin{tabular}[c]{@{}c@{}}CEC2020\\    \\ (F4)\end{tabular}} & Avg & 1952.58 & 1909.46 & 1970.74 & 1920.58 & 1915.80 & \textbf{1903.97} & 1913.49 & 1948.29 & 1939.16 & 1936.87 & 1929.94 & 1924.36 & 2019.63 \\ \hline
\rowcolor[HTML]{EFEFEF} 
 &  &  &  &  &  &  &  &  &  &  &  &  &  &  \\ \hline
 & Best & 5847.98 & 4557.80 & 3485.22 & 3002.60 & 258024.58 & 3533.55 & 4053.34 & 51094.29 & \textbf{2824.95} & 5719.99 & 5630.06 & 3190.27 & 25098.93 \\ \cline{2-15} 
 & Median & 5933.60 & 4721.00 & 3785.54 & 3049.66 & 263562.10 & 3600.27 & 4099.98 & 53952.04 & 2947.64 & 5936.91 & 5945.00 & 3291.87 & 27999.04 \\ \cline{2-15} 
 & Worst & 6044.52 & 5036.45 & 3952.00 & 3614.29 & 289561.05 & 3610.59 & 4152.08 & 54259.00 & 3209.84 & 6039.54 & 6354.08 & 3321.12 & 29542.73 \\ \cline{2-15} 
\multirow{-4}{*}{\begin{tabular}[c]{@{}c@{}}CEC2020\\    \\ (F5)\end{tabular}} & Avg & 5942.03 & 4771.75 & 3740.92 & 3222.18 & 270382.57 & 3581.47 & 4101.80 & 53101.77 & \textbf{2994.14} & 5898.81 & 5976.38 & 3267.75 & 27546.90 \\ \hline
\rowcolor[HTML]{EFEFEF} 
 &  &  &  &  &  &  &  &  &  &  &  &  &  &  \\ \hline
 & Best & 2593.77 & 2586.03 & 3450.29 & 2587.75 & 3347.59 & 2584.59 & 2584.67 & 3174.86 & 2584.66 & 2584.67 & 2584.67 & \textbf{1969.44} & 2584.59 \\ \cline{2-15} 
 & Median & 2735.25 & 2611.20 & 3886.06 & 2599.35 & 3645.98 & 2601.36 & 2698.54 & 3197.25 & 2864.90 & 2699.00 & 2700.01 & 2005.63 & 2674.35 \\ \cline{2-15} 
 & Worst & 3003.52 & 3011.85 & 4152.98 & 2964.00 & 2699.74 & 2719.87 & 2864.36 & 3400.87 & 2974.00 & 2780.08 & 2759.10 & 2059.02 & 2687.00 \\ \cline{2-15} 
\multirow{-4}{*}{\begin{tabular}[c]{@{}c@{}}CEC2020\\    \\ (F6)\end{tabular}} & Avg & 2777.51 & 2736.36 & 3829.77 & 2717.03 & 3231.10 & 2635.27 & 2715.85 & 3257.66 & 2807.85 & 2687.91 & 2681.26 & \textbf{2011.36} & 2648.64 \\ \hline
\rowcolor[HTML]{EFEFEF} 
 &  &  &  &  &  &  &  &  &  &  &  &  &  &  \\ \hline
 & Best & 5845.64 & 8813.24 & 5966.61 & 5969.59 & 22618.07 & 3326.35 & 6252.71 & 3489.56 & \textbf{3126.72} & 6022.11 & 7339.57 & 3290.37 & 10615.35 \\ \cline{2-15} 
 & Median & 5897.21 & 8968.95 & 6124.84 & 6333.25 & 25354.88 & 3418.99 & 6451.00 & 3490.25 & 3269.74 & 6125.00 & 7386.47 & 3361.09 & 11256.68 \\ \cline{2-15} 
 & Worst & 6000.87 & 9005.67 & 6247.85 & 6452.14 & 2597.75 & 3500.06 & 6600.74 & 3947.28 & 3458.60 & 6203.03 & 7529.42 & 3669.96 & 15666.85 \\ \cline{2-15} 
\multirow{-4}{*}{\begin{tabular}[c]{@{}c@{}}CEC2020\\    \\ (F7)\end{tabular}} & Avg & 5914.57 & 8929.28 & 6113.10 & 6251.66 & 16856.90 & 3415.13 & 6434.81 & 3642.36 & \textbf{3285.02} & 6116.71 & 7418.48 & 3440.47 & 12512.96 \\ \hline
\rowcolor[HTML]{EFEFEF} 
 &  &  &  &  &  &  &  &  &  &  &  &  &  &  \\ \hline
 & Best & 2207.23 & 2592.47 & 2591.20 & 2593.54 & 2302.78 & \textbf{2201.06} & \textbf{2201.06} & 2591.20 & 2201.12 & 2201.07 & 2201.08 & 2219.51 & 2591.20 \\ \cline{2-15} 
 & Median & 2498.60 & 2987.11 & 2864.36 & 2614.14 & 2869.85 & 2499.67 & 2455.91 & 2697.36 & 2299.00 & 2377.50 & 2400.00 & 2278.83 & 2609.91 \\ \cline{2-15} 
 & Worst & 3005.36 & 3004.80 & 3269.96 & 2997.76 & 3125.21 & 2529.37 & 2533.77 & 2886.49 & 2566.78 & 2902.00 & 3008.81 & 2464.19 & 2831.46 \\ \cline{2-15} 
\multirow{-4}{*}{\begin{tabular}[c]{@{}c@{}}CEC2020\\    \\ (F8)\end{tabular}} & Avg & 2570.39 & 2861.46 & 2908.50 & 2735.14 & 2765.94 & 2410.03 & 2396.91 & 2725.01 & 2355.63 & 2493.52 & 2536.63 & \textbf{2320.84} & 2677.52 \\ \hline
\rowcolor[HTML]{EFEFEF} 
 &  &  &  &  &  &  &  &  &  &  &  &  &  &  \\ \hline
 & Best & 9682.32 & 5628.91 & \textbf{2449.17} & 4167.93 & 46601.01 & 2969.59 & 2612.46 & 2895.66 & 2726.30 & 7735.53 & 2693.20 & 2581.68 & 9759.78 \\ \cline{2-15} 
 & Median & 9724.92 & 5671.02 & 2547.58 & 4325.87 & 49785.00 & 3265.54 & 2658.41 & 2985.09 & 2711.65 & 7952.19 & 2734.91 & 2766.55 & 10025.44 \\ \cline{2-15} 
 & Worst & 10584.21 & 6004.67 & 2841.19 & 4452.91 & 60875.32 & 3508.98 & 2794.18 & 3000.02 & 3154.90 & 8125.25 & 2913.37 & 2784.99 & 12449.22 \\ \cline{2-15} 
\multirow{-4}{*}{\begin{tabular}[c]{@{}c@{}}CEC2020\\    \\ (F9)\end{tabular}} & Avg & 9997.15 & 5768.20 & \textbf{2612.64} & 4315.57 & 52420.44 & 3248.03 & 2688.35 & 2960.25 & 2864.28 & 7937.65 & 2780.49 & 2711.07 & 10744.81 \\ \hline
\rowcolor[HTML]{EFEFEF} 
 &  &  &  &  &  &  &  &  &  &  &  &  &  &  \\ \hline
 & Best & 6137.49 & 2612.74 & 2737.46 & 2509.69 & 2638.63 & \textbf{2502.56} & 2695.02 & 2779.54 & 2502.63 & 2886.54 & 2691.17 & 2506.48 & 2933.48 \\ \cline{2-15} 
 & Median & 6451.83 & 2944.32 & 2985.00 & 2849.64 & 2900.48 & 2584.32 & 2795.39 & 2985.65 & 2832.26 & 2978.44 & 2982.91 & 2800.11 & 3002.50 \\ \cline{2-15} 
 & Worst & 6630.09 & 3150.47 & 3200.97 & 3056.19 & 2987.36 & 2591.03 & 3120.90 & 3365.00 & 2964.95 & 3291.76 & 3252.61 & 2998.64 & 3581.77 \\ \cline{2-15} 
\multirow{-4}{*}{\begin{tabular}[c]{@{}c@{}}CEC2020\\    \\ (F10)\end{tabular}} & Avg & 6406.47 & 2902.51 & 2974.47 & 2805.17 & 2842.15 & \textbf{2559.30} & 2870.43 & 3043.39 & 2766.61 & 3052.24 & 2975.56 & 2768.41 & 3172.58 \\ \hline
\end{tabular}}
\end{table}

\begin{figure}[t!]
    \centering
    \includegraphics[width=\textwidth]{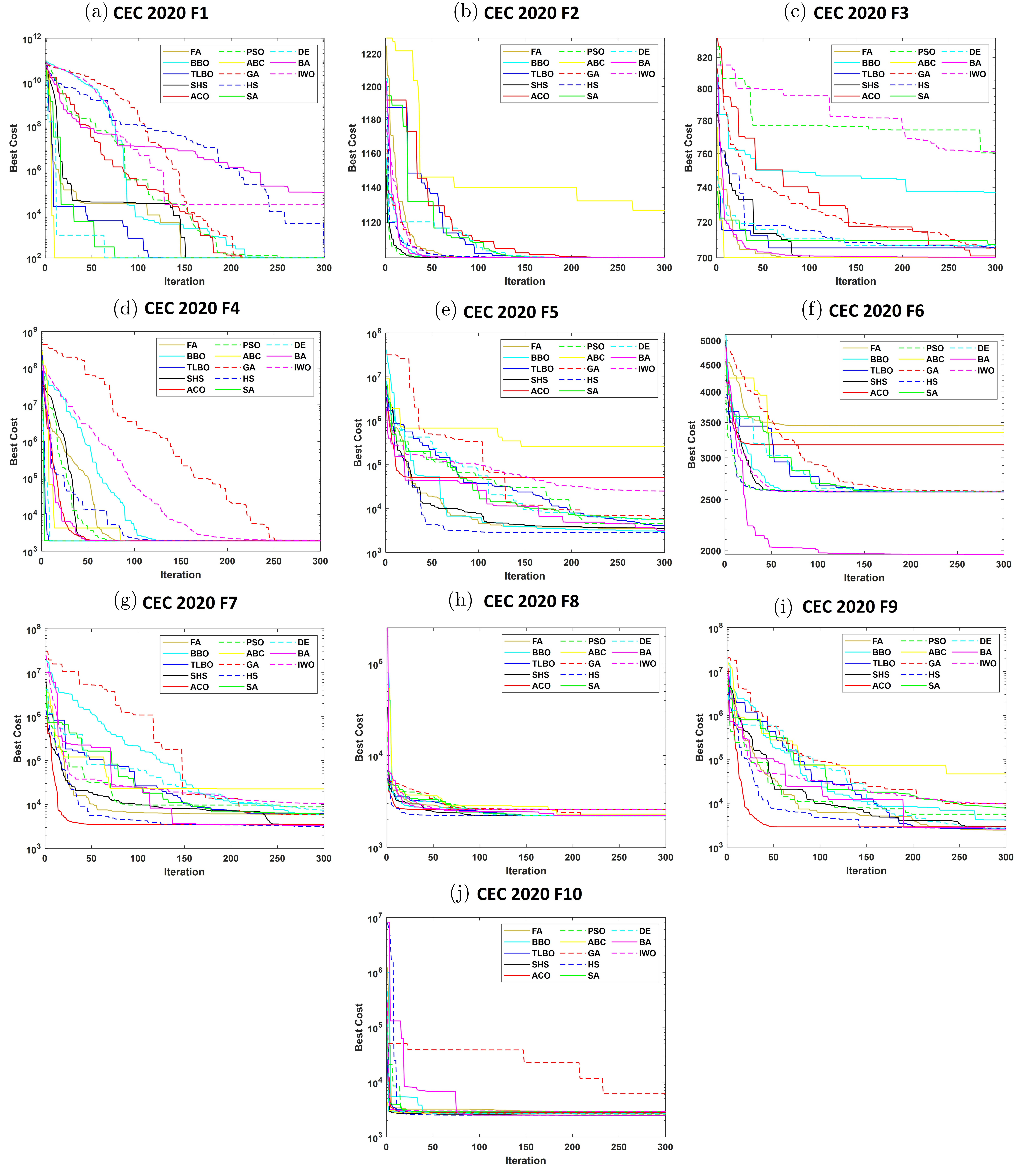}
    \caption{Comparison of SHS algorithm with benchmark algorithms on CEC2020 functions.}
    \label{fig:cec2020convergence}
\end{figure}
\newpage
\bibliography{Manuscript}

\end{document}